\documentclass[twoside,11pt]{article}
\usepackage{jair,rawfonts}
\usepackage[apaciteclassic,nodoi]{apacite}

\usepackage{microtype}
\usepackage{graphicx}
\usepackage{subfigure}
\usepackage{booktabs} % for professional tables
\usepackage[implicit=false]{hyperref}

\usepackage{array}
\usepackage{wrapfig}
\usepackage{mathrsfs}
\usepackage{tabu}
\usepackage{mathtools}
\usepackage{amsmath}
\usepackage{amsthm}
\usepackage{bm}
\usepackage{amssymb}
\usepackage{cancel}
\usepackage{bbm,dsfont}
\usepackage{soul}
\usepackage{graphicx}
\usepackage{tcolorbox}
\tcbuselibrary{theorems}
\mathtoolsset{showonlyrefs}

\usepackage{multicol, blindtext}
\usepackage[algo2e,linesnumbered,ruled,lined]{algorithm2e}

\usepackage[utf8]{inputenc}     
\usepackage[english]{babel}    
\usepackage{tikz}

\usepackage{graphicx}
\newcolumntype{L}{>{$}l<{$}}
\newcommand{\dg}{\textsl g}
\newcommand{\bd}{\boldsymbol{\cdot}}
\newcommand{\bs}{\boldsymbol{\sigma}}
\newcommand{\bp}{\boldsymbol{\psi}}

\DeclareMathOperator*{\argmax}{argmax}
\DeclareMathOperator*{\argmin}{argmin}
\DeclareMathOperator*{\expec}{\mathop{\mathbb{E}}}
\newcommand\mydots{\hbox to 1em{.\hss.\hss.\hss}}

\newtcbtheorem[number within=section]{mydef}{Definition}%
{colback=blue!5,colframe=blue!35!black,fonttitle=\bfseries}{th}

\newtcbtheorem[number within=section]{mytheo}{Theorem}%
{colback=green!5,colframe=green!35!black,fonttitle=\bfseries}{th}

\newtheorem{theorem}{Theorem}[section]
\newtheorem{definition}{Definition}[section]

\newtheorem{problem}{Problem}[section]

\usepackage{xcolor}

\newcommand{\mc}{\mathcal}
\DeclarePairedDelimiterX{\infdivx}[2]{(}{)}{%
  #1\;\delimsize\|\;#2%
}

\newcommand{\xoverbrace}[2][\vphantom{\dfrac{A}{A}}]{\overbrace{#1#2}}
\newcommand{\xunderbrace}[2][\vphantom{\dfrac{A}{A}}]{\underbrace{#1#2}}

\newcommand{\tom}[1]{\textcolor{black}{  #1}}
\ShortHeadings{Goal Kernel Planning: Non-Markovian Policies for Logical Tasks}
{Ringstrom, Hasanbeig, \& Abate}
\firstpageno{1}

\begin{document}

% \title{Linearly Solvable Goal Kernel Planning: Non-Markovian Policies for Logical Tasks with Zero-Shot Transfer}

\title{Goal Kernel Planning: Linearly-Solvable Non-Markovian Policies for Logical Tasks with Goal-Conditioned Options}

% \title{Goal Kernel Planning: Linearly Solvable Non-Markovian Policies for Logical Hierarchical Tasks}

\author{\name Thomas J. Ringstrom \email rings034@gmail.com \\
\addr Department of Computer Science, University of Minnesota,\\
Minneapolis Minnesota, USA, 55455
\AND
       \name Hosein Hasanbeig \email hosein.hasanbeig@cs.ox.ac.uk \\
       \addr Department of Computer Science, University of Oxford,\\
       Oxford, United Kingdom, OX1 3QD
       \AND
       \name Alessandro Abate \email alessandro.abate@cs.ox.ac.uk \\
       \addr Department of Computer Science, University of Oxford,\\
       Oxford, United Kingdom, OX1 3QD}
% For research notes, remove the comment character in the line below.
% \researchnote

\maketitle

\begin{abstract}
In the domain of hierarchical planning, compositionality, abstraction, and task transfer are crucial for designing algorithms that can efficiently solve a variety of problems with maximal representational reuse. Many real-world problems require non-Markovian policies to handle complex structured tasks with logical conditions, often leading to prohibitively large state representations; \tom{this requires efficient methods for breaking these problems down and reusing structure between tasks.  To this end, we introduce a compositional framework called Linearly-Solvable Goal Kernel Dynamic Programming (LS-GKDP) to address the complexity of solving non-Markovian Boolean sub-goal tasks with ordering constraints. LS-GKDP combines the Linearly-Solvable Markov Decision Process (LMDP) formalism with the Options Framework of Reinforcement Learning. LMDPs can be efficiently solved as a principal eigenvector problem, and \textit{options} are policies with termination conditions used as temporally extended actions; with LS-GKDP we expand LMDPs to control over options for logical tasks. This involves decomposing a high-dimensional problem down into a set of goal-condition options for each goal and constructing a \textit{goal kernel}, which is an abstract transition kernel that jumps from an option's initial-states to its termination-states along with an update of the higher-level task-state. We show how an LMDP with a goal kernel enables the efficient optimization of meta-policies in a lower-dimensional subspace defined by the task grounding.} Options can also be remapped to new problems within a super-exponential space of tasks without significant recomputation, and we identify cases where the solution is invariant to the task grounding, permitting zero-shot task transfer.
\end{abstract}

\section{Introduction}
% \textcolor{red}{This is a draft, do not distribute...}\\

Many problems, both naturally occurring and in artificial scenarios such as video games, involve tasks which are only satisfied contingent on a history of logical events, whether it be collecting multiple keys to open a master chest or retrieving objects to forge into new item. Often, these tasks have precedence constraints on events; for example, an agent may need to obtain wood and water, but needs to retrieve an axe at some point in time before chopping wood. When approached with model-free learning, these problems can be thought of as having sparse rewards that are conditioned on a history of logical \textit{events}, that is, they are \textit{non-Markovian}. These event conditions that can be represented as a high-level automata, where a rewarding accepting state represents task completion \cite{gaon2020reinforcement}. In reinforcement learning (RL), a recent popular method called Reward Machines  tackle this problem ~\cite{icarte2018using,xu2020joint,rens2020learning,toro2020reward}. Reward machines have high-level automaton that records an agent's history to condition reward payouts. However, there still remains the open question of how to quickly synthesize optimal policies for long-range non-Markovian tasks for many problems while reusing previous solutions.

In this paper, we develop a complementary model-based algorithm called Linearly-Solvable Goal Kernel Dynamic Programming (LS-GKDP) which can solve the same kinds of non-Markovian tasks, where a goal kernel is an abstract transition operator that maps an agent from goal to goal. Here we seek to understand long horizon planning with dynamic programming (DP)~\cite{bertsekas2012dynamic}, where Boolean logic conditions on an agent's history are incorporated into a Markovian state representation as a Boolean vector. DP is a recursive and exact algorithm for computing optimal policies that provides analytic insight into the problem of sequential decision making, often to the benefit of approximation or learning frameworks like RL. For instance, we understand the convergence properties of RL algorithms in relation to the optimal value function produced by DP~\cite{watkins1992q,jaakkola1994convergence,tsitsiklis1994asynchronous}, as they share the same underlying formalism, i.e.; Markov Decision Processes (MDP)~\cite{puterman2014markov}.
Likewise, we argue that DP-based methods can help reveal mechanisms for transfer and generalization for \textit{sequential goal-problems}. 

A number of other model-based approaches have been proposed that have more general non-Markovian task formulations based on past linear temporal logic \cite{bacchus1996rewarding} \cite{bacchus1997structured} and forward linear temporal logic \cite{thiebaux2006decision}, however, these formalisms do not have natural decompositions over space of possible tasks. Similarly, in hierarchical RL, the MAX-Q framework \cite{dietterich2000hierarchical} also decomposes hierarchical problems into sub-problems but the learning algorithm converges to a local ``recursive-optimality". However for long-horizon ``sparse-reward" problems there is not always a guarantee that optimal solutions for local sub-goals can be used to find globally optimal solutions for sparse objectives. While the mentioned approaches do not exploit optimal sub-problems, our DP algorithm has a simpler non-Markovian task semantics which can help break down the problem into optimal components that can be used to solve the global problem for many different task instances. 

\tom{Our framework, LS-GKDP, combines two major ideas: the Linearly-solvable Markov Decision Process (LMDP) \cite{todorov2009efficient} and the \textit{Options Framework} in Reinforcement Learning \cite{sutton1999between}. Briefly, the linearly-solvable framework has an entropy-regularized objective function which allows one to solve optimal control problems as a linear system of equations, where the value function can be computed as a principal eigenvector problem, which helps simply planning algorithms. Additionally, an \textit{option}, $o=(\pi,\beta)$, in RL is a control policy $\pi(a|x)$ paired with a termination function $\beta(x)\rightarrow [0,1]$ that determines if the policy terminates at a state with a given probability. Options can be sequenced one after another with a meta-policy that selects options after each termination event \cite{fox2017multi,barreto2019option}, and can be used to plan over long horizons while also promoting planning representation reuse across problems. An \textit{option transition kernel} $J(x_f|x_i,o)$ is a stochastic \textit{macro operator}, meaning it is a conditional probability distribution which maps the state $x_i$ that an option-policy is initiated from to a final termination state $x_f$ under and option $o$. Thus, if one has a transition kernel for an option, one can compose the distributions such that the terminal states of one option are the initial states of the next option, allowing a controller to plan with long-ranged jumps. Silver and Ciosek \citeyear{silver2012compositional, ciosek2015value} have developed Bellman equations for ``jumpy" planning with a framework called \textit{Option Models}, where new option transition kernels can be created from simpler components. However, their work uses infinite horizon discounted reward-maximization and currently there are no algorithms of this kind for hierarchical or non-Markovian tasks, which is what we pursue in this paper.}

When including a task space in the state-space definition, we are required to work in a Cartesian product of the low-level and task-level state-spaces, which becomes intractable with increases in size to either space. Therefore, with LS-GKDP we solve an abstract problem in a lower-dimensional state-space using \textit{goal kernels}. A goal kernel is an option transition kernel extended over the high-dimensional Cartesian product-space of state-variables. This abstract transition kernel which is derived from an ensemble of possible options maps the agent from any initial state to a single terminal (goal) state with the probability it reaches it under a goal-conditioned option, while also updating higher-level task-space state-transition of completing the goal. Thus, our options will be point options \cite{jinnai2019finding}, which terminate at a single goal (or at states where achieving the goal is infeasible). This allows us to ignore the non-terminal states that do not participate in the task-state transitions and only optimize over the set of terminal states of options. Thus, our work contributes a scalable closed-form cost-minimizing DP solution to state-reachability problems with long-horizon non-Markovian terminal rewards at a task-completion state. 

LS-GKDP is a framework that uses specific kinds of objects and functions that facilitates key properties. We show that ordered sub-goal tasks can be formulated in a binary-vector task space that records sub-goal progress (Section~\ref{section:BOGtask}). Dynamics through the task-space can be induced by high-level goal variables induced from low-level state-actions through an \textit{affordance function} which links low-level state-actions to their effect on the task space (Section~\ref{Section:Grounding}). We demonstrate how to construct an ensemble of reusable goal-conditioned shortest-path options on a base state-space in order to drive the task-space dynamics. Crucially, from these options we can build new abstract transition kernels called \textit{goal kernels} that represent initial-to-final transitions for low-level and task states for each policy (Section \ref{section:goalop}). The goal kernel plays a central role in an efficient task-policy optimization, called the Task-LMDP (TLMDP), which stitches together low-level goal-condition options to solve the task (Section~\ref{section:TLMDP}). The key property of an option ensemble is \textit{remappability}: we can remap options in an ensemble to new high-level tasks with different affordance functions, allowing us to compute meta-policies within a super-exponential space of problems without recomputing low-level representations, drastically decreasing the time it takes to solve new tasks. Furthermore, we show that it is sufficient to compute the task-solution restricted to the \textit{affordance subspace} of state-actions defined by the affordance function, which has significantly better scaling properties under the increase of state-space size and sub-goal number than computing a solution with DP in the full Cartesian product state-space formed by the task- and base-space. A powerful consequence of this is that solutions to the TLMDP in the affordance subspace can have an important invariance property, which allows the solution to be remapped to different options with perfect zero-shot transfer.  

We will now illustrate the key components of our approach that we have just mentioned with a simple motivating problem.

\subsection{High-level Overview}

\begin{figure}[h!]
\centerline{\includegraphics[scale=0.55]{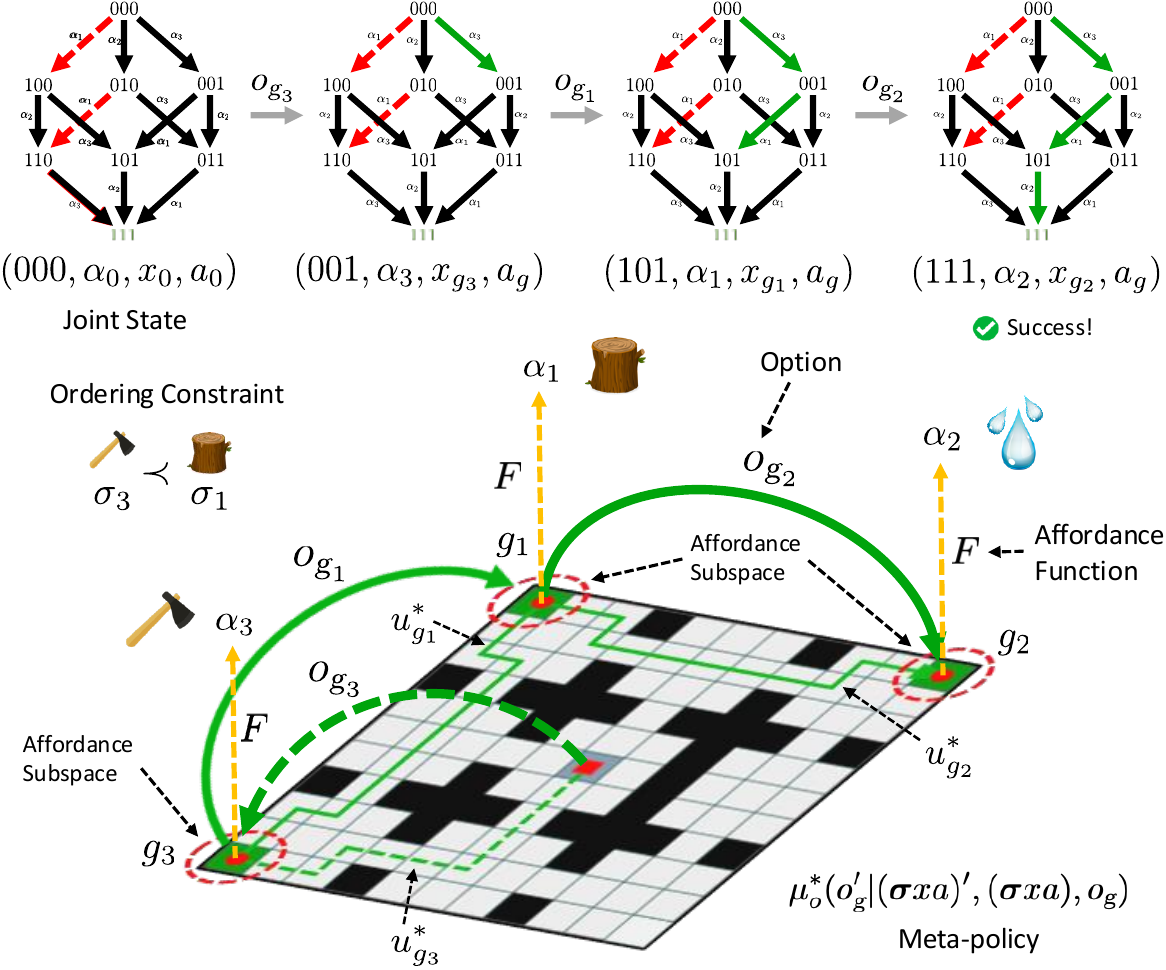}}
\caption{A running example of a non-Markovian problem: An agent needs to obtain an axe, wood, and water, but it needs to retrieve the axe in order to obtain the wood. Thus there are a number of ways of solving the problem, subject to these logical and precedence constraints.  The goal is to, in a cost-efficient manner, arrive at the satisfying task state $111$ in the shown binary cube, where transitions on the cube are induced by actions on the low-level space generated by the policy $u_{\dg}$ of a goal-conditioned option $o_{\dg}$. Red arrows denote transitions that violate precedence constraints. Green arrows show state-option to state-option jumps under the goal kernel. We show how to optimize the meta-policy $\mu^*$ in a small sub-space of the full product-space called the \textit{affordance subspace}.}
\label{fig:running_example}
\end{figure}

\tom{In Fig. \ref{fig:running_example}, we provide a high-level overview of our approach with a motivating example.  An agent has to perform a 3-goal task where it must retrieve an axe, chop wood, and obtain water. The axe must be retrieved before chopping the wood (indicated by an ordering constraint $\prec$) but the water can be obtained at any time. Therefore, the agent would like to complete this task with the minimum cost (or number of steps). There are two state-spaces: the low-level state-space $\mc X$ and the task space $\Sigma$, where a task-state (e.g.) $\bs = (0,1,1)$ is a binary vector which records task progress. Each of these spaces has a transition kernel $P_x(x'|x,a)$ and $P_{\boldsymbol{\sigma}}(\bs'|\bs,\alpha)$ respectively, where $a$ and $\alpha$ are actions and $\alpha_{\dg} \in \{\alpha_\epsilon,\alpha_1,\alpha_2,\alpha_3\}$ flips the corresponding $\dg^{th}$ bit of a binary vector (and $\alpha_0$ does not flip a bit) where each bit corresponds to a goal. The ordering constraint $\sigma_3 \prec \sigma_1$ expresses a constraint on the acceptable dynamics in $P_\sigma$ denoted by the red-dashes, which means that goal $3$ must be completed in order to complete goal $1$. This ordering constraint can express task rules such as \textit{get the axe before you chop wood}.}

\tom{We can link the dynamics of the low-level state-space to the high-level state-space with the affordance function $F:\mc X \times \mc A \times \mc A_{\sigma}\rightarrow[0,1]$. This function $F(\alpha|x,a)$ is a probability distribution that sets the high-level actions from a given low-level state-action, where completing a goal $(x,a)_{\dg}$ at a green square induces $\alpha_{\dg}$ to flip the $\dg^{th}$ bit of the task state indicated by a green line in task space, shown in Fig. \ref{fig:running_example}). Then, the full product-space kernel for the example in Fig. \ref{fig:running_example} can be defined as:}
\begin{align}
    P_{\sigma x}(\bs',x'|\bs,x,a) = \sum_{\alpha}P_{\boldsymbol{\sigma}}(\bs'|\bs,\alpha)F(\alpha|x,a)P_x(x'|x,a).\label{eq:product}
\end{align}
\tom{However, the problem that arises is that the effective Cartesian product state-space $\Sigma \times \mc X$ grows combinatorially in a way that quickly becomes intractable for dynamic programming as the size of $\mc X$ or $\Sigma$ gets bigger, making it impractical to compute a control policy $\pi(\bs,x)\rightarrow a$. In this paper, we show how for each possible high-level action $\alpha_{\dg}$ from the domain of the function $F(\alpha_{\dg}|x,a)$, we can compute a goal-conditioned option $o_{\dg}$, constituting an option set $\mc O = \{o_{\dg_1},...,o_{\dg_n}\}$ that each have corresponding goal (terminal) states $x_g$ in the set $\mc X_g \subset \mc X$ (the green goal states). Then, using this basis of solutions, we can construct a \textit{goal kernel} $G$, by composing the high-level transition kernel $P_{\bs}$ with an option transition kernel $J$ linked by an affordance function $F$,}
\begin{align*}
    G(\bs_f,x_f,a_f|\bs,x,a,o_{\dg}) = \sum_{\alpha_f}P_{\boldsymbol{\sigma}}(\bs_f|\bs,\alpha_f)F(\alpha_f|x_f,a_f)J(x_f,a_f|x,a,o_{\dg}),\label{eq:goalkernel}
\end{align*}
\tom{which transitions the agent around the much smaller goal-subset $\mc X_g$ (called the ``affordance sub-space") by with options in the set $\mc O$. Critically, the goal kernel is defined by a option kernel $J$ which maps the agent from the initial state to the final goal-state with the probability that the option can satisfy the goal under the option; each jump is indicated by a solid green arc in Fig. \ref{fig:running_example}, and the solid green line on $\mc X$ is the agent's trajectory of following the policy. Thus, the goal kernel is more tractable for dynamic programming because it is defined over the significantly smaller product-space $\mc X_g \times \Sigma$.}

\tom{The second thing to note is that when we perform dynamic programming using $G$, the cost $q(x,o_{\dg})$ of using a particular option $o_{\dg}$ from state $x$ will be equal to the value function $v_{\dg}(x)$ corresponding to the option $o_\dg$ evaluated at a starting state $x$. Since the value function of a policy expresses the total cost (or path length) of traveling to the goal, our algorithm will use $G$ to optimize a semi-Markov meta-policy $\mu$ which stitches low-level goal-condition options together while summing up the cost of traveling to each goal using its value function. Thus, the meta-policy will generate a chain of options that each induce trajectories on $\mc X$, which solve the task with the minimum cost that respects the task constraints.}

%\comment{ICML Reviewer: What is passive dynamics in the goal kernel contexts?}
% \comment{ICML Reviewer: MDP is a special case of LMDP, so it would be great to explain problems that can be addressed by LMDP but not MDP.}
% \tom{Re: problems that cannot be addressed by MDPs: I said below in the Why LMDP section that LMDPs allow for the specification of "priors" over dynamics (passive dynamics), and this can be imposed on a symbolic space, necessary for formalizing ``habits" and inductive biases for problems in new domains.}

\subsection{Why Linearly Solvable MDPs?}

The algorithm we present is centered around an entropy-regularized dynamic programming paradigm called the Linearly-Solvable Markov Decision Process (LMDP)~\cite{todorov2009efficient}. One might wonder why our approach should use entropy-regularized DP (LMDPs) rather than standard MDP conventions. Indeed, our algorithm is compatible with traditional MDPs, and we outline this variation in the Appendix. However, the main reason we are working within the LMDP framework is because it has a number of unique and useful properties that we anticipate for current and future use. 

Firstly, the LMDP can be solved for efficiently as a sparse linear system or with an iterative method called z-iteration (analogous to power-iteration). These methods have the same computational complexity of one half of one step of policy-iteration \cite{howard1966dynamic}, a well-known algorithm for computing optimal policies.

Secondly, an important property is that it has a linear superposition principle~\cite{todorov2009compositionality} that allows us to optimally mix component policies into a composite policy. This effectively carries out a weighted OR-logic on the terminal state of a given state space and is therefore natural when task states represent Boolean logical objectives in disjunctive normal form, which is an emphasis of this paper. This property is demonstrated in Fig. \ref{eq:LMDP-pol-mixture}. Standard MDPs do not have this compositionality property and cannot decompose solutions with this OR-logic.
% Secondly, the LMDP framework allows one to set the prior over dynamics in the form of the ``passive dynamics." This could be especially useful %\tom{\st{in the behavioral sciences}} 
% for formalizing and modeling ``habits" over symbolic variable dynamics, rather than just low-level state-spaces, especially when considering the transfer of solutions from one environment to another. 

% The specification of a symbolic passive dynamics could be valuable as a way to preserving information from past environments to a new problem which can be immediately attempted under computational resource or time constraints.

Lastly, another key strength of the LMDP is that it has a control-as-inference interpretation \cite{levine2018reinforcement}, and a max-entropy passive dynamics (prior) in the objective entails a controlled Markov chain that can generate all possible trajectories solving the task, however unlikely they may be under the optimal policy. While this is possible in other entropy regularized RL formulations~\cite{haarnoja2017reinforcement}, with the LMDP, there exists a useful formula for the likelihoods of trajectories sampled from the policy. This leads to a simple formulation for the Bayesian inverse problem of inferring which logical formula is driving the agent's dynamics given a partial observation of a hierarchical trajectory through both the high-level symbolic space and the low-level trajectories which drive the symbolic dynamics 
(see \ref{likelihood_1} and \ref{likelihood_2} for discussion).
% (see app.5.10 for discussion).  
This formula can be very useful for follow-up work in the domain of hierarchical inverse control, wherein the task is to identify the latent motivations (in the form of intended states) of agents that are engaged in tasks. This work can help to extend current ideas in this area to reasoning about intentions over structured non-Markovian tasks. 
% Computational theory-of-mind requires a theory 

\subsection{Related Research} 

This work has connections with several areas of research. The saLMDP and TLMDP are new members in the family of Linearly-Solvable Optimal Control problems~\cite{kappen2005linear,todorov2009efficient}, and are inspired by recent hierarchical innovations for controlling over sub-problems~\cite{jonssonhierarchical,saxe2017hierarchy,tirumala2019exploiting}. Research regarding goal-conditioned policy \textit{ensembles} has been pursued in both RL~\cite{nasiriany2019planning}, and in model-based non-stationary control~\cite{ringstrom2019constraint}. Ordered-goal tasks have also been recently proposed in RL for training agents with symbolic and temporal rules
% cautiousRL, lcnfq, lcrl, certified, yuan2019modular
\cite{illanessymbolic,plmdp,icarte2018using,certified,cautiousRL,hasanbeig2020deep,skalse1,skalse2}. Policy-concatenation (stitching) is used in Hierarchical Control~\cite{peng2019mcp,ringstrom2019constraint}, and~\cite{horowitz2014compositional} provide a continuous HJB approach for LTL tasks, but do not formalize a general algorithm for coupled transition kernels. Other policies concatenating schemes are policy sketches (using modular policies and task-specific reward functions)~\cite{andreas2017modular}, and compositional plan vectors~\cite{devin2019plan}. Lastly, a compositional Boolean task algebra was recently introduced demonstrating zero-shot task transfer for non-sequential policies~\cite{tasse2020boolean}. 
% A Deep-RL approach, h-DQN~\cite{kulkarni2016hierarchical} uses options for long-range planning but does not model a structured task.
\tom{Furthermore, there are a number of other methods which have used options in the RL framework to stitch together policies to complete complex tasks}. Skill machines \cite{tasse2022skill}, as well as other frameworks have been used solve logical problems with decomposed policies \cite{chane2021goal} \cite{araki2021logical} \cite{leon2020systematic} \cite{liu2022skill} \cite{jothimurugan2021compositional}. However, these approaches general lack analysis of a full hierarchical Bellman equation. The approach that we put forth in this paper differs in that we are directly solving over a full Bellman equation by ``jumping" from sub-goal to sub-goal with option transition kernels to solve non-Markovian tasks in a subspace sub-goal.

% However, for non-Markovian problems, rewards can be unnatural for stitching together policies because the agent can still incentivized to complete sub-goals for impossible tasks simply because they are rewarded: rewarding sub-goals are not additive to the full value function, whereas paths-lengths and costs are.

\subsection{Contributions} The major contributions of this work are listed as follows:\\
\textbf{1) saLMDP/TLMDP:} We extend the LMDP formalism to operate on state-action pairs (saLMDP, Section~\ref{section:saLMDP}), and use this to define a task-LMDP (TLMDP, Section~\ref{section:TLMDP}) to solve logical tasks with precedence constraints using options. TLMDPs can embed traveling salesman problems (See Fig. \ref{fig:results_fig}.A).\\
\textbf{2) BOG task grounding:} The logical task is specified as a Boolean Ordered Goal (BOG) task in a binary vector state-space, with precedence constraints on actions which flip the bits (Section~\ref{section:BOGtask}). We map the actions of a BOG task to low-level state-actions with an affordance function (Section~\ref{Section:Grounding}). \\
\textbf{3) Scalability:} The TLMDP is \textit{scalable} in that it mitigates the complexity introduced by ``interior states" of the low-level space by solving a compact abstract saLMDP defined on the terminal states of options. This requires creating a option kernel and goal kernel(Sections~\ref{section:goalop}, \ref{Section:JumpFeas}) from the low-level policies to summarize the probability of inducing a high-level action.  We demonstrate the scalability of this approach in (Fig. \ref{fig:results_fig} B,C). \\
\textbf{4) DP solution for Non-Markovian tasks:} The TLMDP solves otherwise non-Markovian task by solving a DP problem with a state-space that represents the full task-space, rendering the problem Markovian.\\
\textbf{5) Disjunctive Normal Form:} The TLMDP solution admits a Disjunctive Normal Form (DNF) policy decomposition for each clause in the Boolean formula DNF (See Fig. \ref{fig:dnf}) \\
\textbf{6) Remappability and Transfer:} We show when the action space of Boolean tasks can be flexibly remapped to different options with zero-shot transfer, which we conceptualize as task transfer. The computational advantage of this is demonstrated in Fig. \ref{fig:results_fig} D,E.

\newpage
\section{Background}
\subsection{The Linearly-Solvable Markov Decision Process}
The Linearly Solvable Markov Decision Process (LMDP)~\cite{todorov2009efficient} is an entropy-regularized alternative to a standard Markov Decision Process (MDP). It is defined as a three-tuple $(\mc X,p,q)$ where $\mc X$ is the finite set of states, $p$ is an uncontrolled ``passive dynamics" transition kernel $p : \mc X \times \mc X \rightarrow [0, 1]$, and $q: \mc X\rightarrow \mathbb{R}$ is a state-cost function. The Linear Bellman Equation is defined as:
\begin{equation}
v^*(x)=\min\limits_{\substack{u}} \Big[q(x) + \mc{D}_{KL}(u(x'|x)||p(x'|x)) + ~\expec\limits_{~~~~\mathclap{{x'\sim u(\cdot|x)}}~~~~}~v^*(x')\Big],
\label{eq:original}
\end{equation}
where $v:\mc X \rightarrow \mathbb{R}$ is the value function, $u:\mc X \times \mc X \rightarrow [0,1]$ is an arbitrary policy, and $\mc{D}_{KL}(a||b) := \expec_{x\sim a}\log(\frac{a(x)}{b(x)})$ is the Kullback–Leibler (KL) divergence \cite{kullback1951information} between two given probability distributions $a$ and $b$. Unlike the standard MDP, the action variable has been replaced with a controlled transition matrix $u(x'|x)$. Therefore, instead of optimizing a state-to-action map $\pi^*(a|x)$, as one would with standard algorithms such as value-iteration~\cite{bellman1957markovian}, we optimize for desirable state-to-state Markov chain dynamics. Action costs in \eqref{eq:original} are conceptualized as deviations away from the passive dynamics, as measured by the KL-divergence. The KL-divergence acts as an information cost that penalizes how much \textit{extra} information it takes to identify transitions emitted from a new controlled distribution $u$ when assuming the transitions were generated with the encoding of a prior distribution $p$. The larger the state cost $q(x)$ is, the more incentivized the agent will be to diverge from $p$ in order to avoid costly trajectories through the state-space.

The solution to \eqref{eq:original} will be a policy $u^*$ which minimizes the expected accumulated cost contributed by both the state-cost function $q$ and the action cost under the future dynamics. The details of the solution derivation for this objective function will not be presented here in full, but is described in Todorov \citeyear{todorov2009efficient}. The methodology used in the original proof will be mirrored in our derivation of the \textit{state-action LMDP} in the subsequent section. 

Solving \eqref{eq:original} requires a transformation of the value function $v(x)$ to a \textit{desirability} function $z(x)$ by negative exponentiation: 
\begin{align*}
    z(x) = \exp(-v(x)). 
\end{align*}
In the original work, it was shown that the optimal controlled dynamics, $u^*$, can be determined by rescaling the passive dynamics by the optimal desirability function and normalizing,
\begin{align}
    u^*(x'|x)=\frac{p(x'|x)z^*(x')}{\mathfrak{N}[z^*](x)},\label{eq:LMDP_pol}
\end{align}
where the normalization function $\mathfrak{N}[z](x) = \sum_{x'}p(x'|x)z(x')=\mathop{\mathbb{E}}_{p(\cdot|x)}z(x')$ is a linear operator of the expected desirability under one-step of the passive dynamics.
The desirability function $z$ can be computed as the solution to the simple linear equation,
\begin{align}
    z^*(x) = \exp(-q(x))\mathfrak{N}[z^*](x), \label{eq:LMDP_linear}
\end{align}
To write this in linear algebraic form, let $Q=\text{diag}(\exp(-\mathbf{q}))$ be a diagonal matrix where $\mathbf{q}$ is the cost vector on $\mc X$. The normalization function $\mathfrak{N}[z^*](x)$ is $P\mathbf{z}$ where $P$ is the Markov matrix of the passive transition kernel $p$ such that $\forall i:\sum_{j}P_{ij}=1$, $P_{ij}=p(x_j|x_i)$, and $\mathbf{z}$ is the desirability function vector on $\mc X$. Thus, the solution to Eq. \eqref{eq:LMDP_linear} is computed for all $\mc X$ as a principal eigenvector of the matrix $QP$:
\begin{align}
    \lambda_1\mathbf{z}=QP\mathbf{z},
\end{align}
where $\mathbf{z}$ is the principal eigenvector corresponding to an eigenvalue $\lambda_1=1$. Thus, computing the policy in Eq. \eqref{eq:LMDP_pol} is effectively reshaping a (typically maximum entropy) prior over dynamics $p$ into a posterior ``desired dynamics," $u^*$.  The LMDP is not conventional in the sense that the world-transition model is implicitly encoded in the non-zero entries of the passive matrix $P$. The agent can only alter the dynamics for these non-zero entries because adding any probability mass to the zero-entries in $P$ would result in an infinite KL-cost. When we develop the state-action LMDP in this paper, we will actually reintroduce symbolic actions with a an explicit transition kernel $p(x'|x,a)$, which is a modeling advantage.

Our primary focus in this paper will be on the case of first-exit control (a generalization of stochastic shortest path problems), where the value function is defined up until the event that the agent hits a specified terminal state in the set $\mc T$ (also called the boundary or goal state). The value function of the terminal state is equal to the cost associated to that state. A first-exit Bellman equation is generally defined as \cite{todorov2009efficient, bertsekas2012dynamic},
\begin{align*}
    v^*(x) = \min_u\left[\ell(x,u)+\sum_{x'}P(x'|x,u)v^*(x')\right],\quad s.t. \quad v(\tilde{x}) = q(\tilde{x}),\quad \forall \tilde{x} \in \mc T,
\end{align*}
where in the LMDP the cost function is $\ell(x,u) = q(x) + \mc D_{KL}(u||p)$, where $u(x'|x)$ is a controllable state transition kernel instead of an action variable, and $\mc T \subset \mc X$ is a set of terminal states. By using the notation $\mc N$ for non-terminal states and $\mc T$ for terminal (goal) states, one can segregate the non-terminal to non-terminal transitions from the non-terminal to terminal transitions. We express these sub-matrices as $P_{\mc N \mc N}$, $P_{\mc N \mc T}$, and $Q_{\mc N}$, and the sub-vectors are expressed as $\mathbf{z}_{\mc N}$, $\mathbf{z}_{\mc T}$, $\mathbf{q}_{\mc N}$ and $\mathbf{q}_{\mc T}$. Using the definition of the desirability function at the terminal states, $\mathbf{z}_{\mc T}=\exp(-\mathbf{q}_{\mc T})$, one can compute $\mathbf{z}$ with an algorithm known as z-iteration:
\begin{align}
    \mathbf{z}_{\mc N} \leftarrow Q_{\mc N}P_{\mc N \mc N}\mathbf{z}_{\mc N}+Q_{\mc N}P_{\mc N \mc T}\mathbf{z}_{\mc T},\label{eq:z-iter}
\end{align}
which is a variant of power-iteration, an algorithm typically used for computing principal eigenvectors. Alternatively, $\mathbf{z}$ can be computed as the solution to the following linear system:
\begin{align*}
    (\text{diag}(\exp(\mathbf{q}_{\mc N}))-P_{\mc N \mc N})\mathbf{z}_{\mc N} = P_{\mc N \mc T}\exp(-\mathbf{q}_{\mc T}),\label{eq:z-linsys}
\end{align*}
% $\tilde{Q}_{\mc N} = diag(\exp(\mathbf{q}_{\mc N}))$
where $\mathbf{z}_{\mc N}$ is unknown, and $\mathbf{q}_{\mc T}$ is known.
% \begin{align*}
% % u^*(x'|x) = p(x'|x)z(x')/\mathfrak{N}[z^*](x)\\
% u^*(x'|x)=\frac{p(x'|x)z(x')}{\mathfrak{N}[z^*](x)}\quad \text{where} \quad \mathbf{z}=QP\mathbf{z}
% \label{eq:largesteig}
% \end{align*}
To obtain the policy, we re-scale the passive dynamics with $z$, and normalize the rescaled dynamics by the average desirability under the passive dynamics, as shown in $\eqref{eq:LMDP_pol}$.

% \subsection{Properties of saLMDP}
% The original LMDP had a number of interesting properties

In the following sections we will show that we can extend the LMDP to state-action to state-action transitions, which will give us a more expressive representational capacity, allowing us to specify an explicit transition model and define explicit costs on specific symbolic actions, which we can use to our advantage for hierarchical non-Markovian planning.

\subsection{The Options Framework}
\tom{As we noted in the introduction, our theory is formalized in the Options framework \cite{sutton1999between}. An option $o=(\pi, \beta)$ is a $\pi$ is a policy and termination function $\beta: \mc X \rightarrow [0,1]$ that returns the probability that a termination event will occur at state $x$. Options can be thought of as instructions for using a policy $\pi$, and are a form of \textit{semi-Markov} planning with meta-policies. This means an agent initializes and follows a policy until a variable time-length termination event, in which a new policy is initialized from a meta-policy $\mu:\mc O \times \mc X \rightarrow [0,1]$ which outputs an option $o$ from state $x$ with a given probability. The variable time-length between option selections arises from the distance and stochasticity en route to a termination event. In this paper we will be using state-action LMDP policies for options, and we will be directly optimizing a state-action LMDP version of the meta-policy to schedule options. 
% Options gives an agent a semi-Markov program for when polices terminate and initiate new Markovian dynamics.
}

\section{Algorithm: Linearly-Solvable Goal Kernel Dynamic Programming}
% Here we outline a general scheme for using abstract operators to solve complex logical problems.  
\begin{problem}
In this paper we solve the problem of solving semi-MDP meta-policies over options for structured non-Markovian tasks, represented as a task state-space $\Sigma$. This task space is coupled to and controlled by the low-level states $\mc X$. We aim to mitigate the complexity of working in the Cartesian product space $\Sigma \times \mc X$ and show that the Linearly Solvable MDP framework lends some of its important properties for policy decomposition for our tasks. We seek a solution that enables zero-shot transfer from one problem definition to another, where task transfer is defined as the re-parameterization of high-level actions with low-level options.
\end{problem}
\subsection{State-Action LMDP}\label{section:saLMDP}
We introduce a novel extension of the LMDP, that operates on state-action pairs. This extension will be employed in two places throughout the algorithm, for both the low-level goal-conditioned policy ensemble, and the high-level TLMDP task policy that ``stitches'' option-policies from the ensemble together. Reintroducing the action variable will have non-trivial consequences for the TLMDP as we can introduce abstract initial-to-final state transition dynamics, and more expressive cost functions including precedence constraints on sub-goals and the cost of following a given policy to complete a sub-goal, while retaining a linear solution.
\begin{definition}[State-Action LMDP]\label{def:saLMDP}
    A state-action LMDP (saLMDP) is defined as the tuple $\mathscr{M}=(\mc X, \mc A, p_x, p_a, q)$, where $\mc X$ is the finite set of states, $\mc A$ is the finite set of actions, $p_x:(\mc X \times \mc A) \times \mc X \rightarrow [0,1]$ is a state-transition kernel, $p_a:(\mc X \times \mc A) \times (\mc X \times \mc A) \rightarrow [0,1]$ is the ``passive" action-transition kernel, and $q: \mc X \times \mc A \rightarrow \mathbb{R}_+$ is the cost function which has non-zero and non-infinite cost $c$ for all state-actions that are not goals or obstacles.
\end{definition}

What follows will be an abridged derivation of the solution which is similar to the original LMDP solution, but the reader may refer to Section~\ref{sec:saLMDP_append} for a full proof with the omitted steps.  We formulate a new objective function by defining the state in saLMDP as a state-action pair ${y} = (x,a)$, and then substitute ${y}$ for $x$ in (\ref{eq:original}). After substitution, we can decompose the resulting joint distributions $u_{xa}(x',a'|x,a)$ and $p_{xa}(x',a'|x,a)$ into $$u_{xa}:=u_x(x'|x,a)u_a(a'|x';x,a),\quad\quad p_{xa}:=p_x(x'|x,a)p_a(a'|x';x,a),$$ via the chain rule, where $u_x:\mc X \times \mc A \times \mc X \rightarrow [0,1]$ is, oddly and for lack of a better term, a state-transition ``policy", and where $u_a:\mc X \times \mc A \times \mc X \times \mc A \rightarrow [0,1]$ is an action transition policy. This results in the equation,
\begin{gather*}
v^*(x,a)=\min\limits_{\substack{u_a \\ u_x}} \Bigg[q(x,a) + \mathop{\mathbb{E}}\limits_{\substack{a'\sim u_a \\ x'\sim u_x}}\Bigg[\log\Bigg( \frac{u_{a}(a'|x';x,a)u_{x}(x'|x,a)}{p_{a}(a'|x';x,a)p_{x}(x'|x,a)}\Bigg)\Bigg] + \mathop{\mathbb{E}}\limits_{\substack{a'\sim u_a \\ x'\sim u_x}}[v^*(x',a')]\Bigg],
\end{gather*}
where the middle term is the KL-divergence. Note that $v(x,a)$ is the Q-function, $\mc Q(x,a)$, in RL~\cite{sutton1998introduction}. However, like the original LMDP, the saLMDP solution requires the use of the \textit{desirability function},
\begin{align*}
    z(x,a)=\exp(-v(x,a)),
\end{align*}
so we will refrain from using $\mc Q(x,a)$ since it does not play a primary role and risks being mistaken for the standard LMDP notation in which $q$ is a cost-function and $Q$ is a cost-matrix. We will also avoid creating a new symbol for ``Q-function desirability" and simply use $z$.

The idea of optimizing $u_x(x'|x,a)$---which would be interpreted as a state-transition ``policy" distribution---is not sensible, as we would be directly altering the state-transition probabilities encoded in the kernel $p_x$ which is a model of the physics of a system. Therefore, we enforce an optimization constraint that $u_{x}(x'|x,a) = p_{x}(x'|x,a)$ to constrain the agent to a fixed model---enforcing this condition makes these two distributions cancel out in the KL term. Thus, we only allow the controller to modify the distribution over actions, $u_a$, giving rise to the canonical saLMDP objective and solution:
% \begin{gather}
% \begin{split}
% v(x,a)=\min\limits_{\substack{u_a}} \Bigg[&q(x,a) + \mathop{\mathbb{E}}\limits_{\substack{a'\sim u_a}}\Bigg[\log\Bigg( \frac{u_{a}(a'|x',x,a)}{p_{a}(a'|x',x,a)}\Bigg)\Bigg] \\&+ \mathop{\mathbb{E}}\limits_{\substack{a'\sim u_a \\ x'\sim u_x=p_x}}[v(x',a')]\Bigg],\end{split}\\
% u_a^{*}(a'|x',x,a)=\frac{p_a(a'|x',x,a)z(x',a')}{\mathfrak{N}[z^*](x',x,a)}
% \label{eq:saPol},\\
% \mathbf{z}=QP_{xa}\mathbf{z},
% \label{eq:saLMDP_z}
% \end{gather}
\begin{align*}
     v^*(x,a)=\min\limits_{\substack{u_a}} \left[q(x,a) + \mathop{\mathbb{E}}\limits_{\substack{a'\sim u_a \\ x'\sim u_x=p_x}}\left[\log\left(\frac{u_{a}(a'|x';x,a)}{p_{a}(a'|x';x,a)}\right)\right] +~~ \expec\limits_{\mathclap{\substack{a'\sim u_a \\ x'\sim u_x=p_x}}}~~[v^*(x',a')]\right].
\end{align*}
After substituting $v(x,a)=-\log(z(x,a))$, the equation can be written in a simpler form by pulling $q$ out of the $\min(\cdot)$ function and combining the two expectations into one term:
\begin{align}
    -\log(z^*(x,a))= q(x,a) + \min\limits_{\substack{u_a}} \left[ \mathop{\mathbb{E}}\limits_{\substack{a'\sim u_a\\ x'\sim p_x}}\log\left(\frac{u_{a}(a'|x';x,a)}{p_{a}(a'|x';x,a)z^*(x,a)}\right)\right].\label{eq:saLMDP_simplified}
\end{align}
Notice that the term inside the minimization is similar to the KL-divergence, but the denominator needs to be normalized to sum to $1$ in order for it to be a true divergence measure on distributions. We can define a normalization function, $\mathfrak{N}$, which takes the expectation of future desirability under the passive dynamics,
\begin{align*}
    \mathfrak{N}[z](x',x,a) = \sum_{a'}p_a(a'|x';x,a)z(x',a') =\mathop{\mathbb{E}}_{a'\sim p_a}z(x',a').
\end{align*}
Then multiplying $\mathfrak{N}$ divided by itself, $\times \frac{\mathfrak{N}[z^*](x',x,a)}{\mathfrak{N}[z^*](x',x,a)}=1$, to the denominator in \eqref{eq:saLMDP_simplified},
we can pull the top $\mathfrak{N}[z^*](x',x,a)$ out of the action expectation and $\min(\cdot)$ function as $\mathfrak{N}$ only depends on $p_x$, but not on $u_a$, leaving the lower $\mathfrak{N}[z^*](x',x,a)$ for normalization. We now have $-\log(z^*(x,a))=$
\begin{align}
    q(x,a) + \mathop{\mathbb{E}}\limits_{x'\sim p_x }\Bigg[-\log\big(\mathfrak{N}[z^*](x',x,a)\big)+\min\limits_{\substack{u_a}} \Bigg[\mathop{\mathbb{E}}\limits_{a'\sim u_a}\log\Bigg( \frac{u_{a}(a'|x';x,a)}{\frac{p_{a}(a'|x';x,a)z(x',a')}{\mathfrak{N}[z^*](x',x,a)}}\Bigg)\Bigg]\Bigg].\label{eq:after_normalization}
\end{align}
Because the KL-divergence has a global minimum of $0$ when the numerator is equal to the denominator, we can set the optimal policy $u_a^*$ to be,
\begin{align}
    u_a^{*}(a'|x',x,a)&=\frac{p_a(a'|x',x,a)z^*(x',a')}{\mathfrak{N}[z^*](x',x,a)}\label{eq:salmdp_pol},
\end{align}
and the third term in \eqref{eq:after_normalization} goes to $0$. \tom{After multiplying both sides by $-1$ and exponentiating,} the Bellman equation takes a simpler form:
\begin{align*}
    z^*(x,a) = \exp(-q(x,a))\exp\left(\mathop{\mathbb{E}}\limits_{x'\sim p_x(\cdot|x,a) }\log\left(\mathfrak{N}[z^*](x',x,a)\right)\right).
\end{align*}
Observe that when $p_x$ is deterministic the expectation and $\exp(\cdot)$ function can be switched because the equality condition of Jensen's inequality ($f\left( \mathbb{E}[X] \right) \leq \mathbb{E}\left[ f(X) \right]$ for a convex $f$) holds. The $\exp(\cdot)$ and $\log(\cdot)$ functions will then cancel, resulting in the linear equation:
\begin{align}
    z^*(x,a) = \exp(-q(x,a))\mathop{\mathbb{E}}\limits_{x'\sim p_x}\mathfrak{N}[z^*](x',x,a)\label{eq:salmdp_linear_z}.
\end{align}
% \begin{align}
%     z(x,a) = \exp(-q(x,a))\mathop{\mathbb{E}}\limits_{x'\sim p_x(\cdot|x,a) }\mathfrak{N}[z^*](x',x,a)\label{eq:salmdp_linear_z}.
% \end{align}
% \begin{align}
%     z(x,a) = \exp(-q(x,a))\sum_{x'}p_x(x'|x,a) \mathfrak{N}[z^*](x',x,a)\label{eq:salmdp_linear_z}.
% \end{align}
In matrix-vector form, the desirability function is:
\begin{align}
    \lambda_1\mathbf{z}&=QP_{xa}\mathbf{z},\label{eq:salmdp_eig}
\end{align}
where $Q = \text{diag}(\exp(-\mathbf{q}))$ and $P_{xa}$ is the matrix form of the joint passive dynamics $p_{xa}(x',a'|x,a)$, and $\mathbf{z}$ is the desirability vector on $\mc X \times \mc A$ which is a principal eigenvector of $QP_{xa}$ with an eigenvalue of $1$. Pseudocode for the saLMDP is given in algorithm \ref{algor1} \tom{for computing the solution with z-iteration}.  \tom{As we can see in the algorithm, computing the solution can be performed by iterative updating the desirability vector $\mathbf{z}$ until convergence, just as one would with z-iteration for standard LMDPs.} Also, we will henceforth refer to $u^*_a$ as $u^*$ for compactness. For first-exit control the desirability function can be computed with the same methodologies described in the previous section, either z-iteration \eqref{eq:z-iter}, or as a linear system \eqref{eq:z-linsys}. In the case where $p_x$ is stochastic, the solution does not reduce down to a linear equation, but there exists an iterative non-linear mapping for computing the desirability function, 
% app.5.3 Eq.\eqref{eq:nonlinear_z}. 
see~\eqref{eq:nonlinear_z} in Section~\ref{sec:saLMDP_append}.

The policy $u_a(a'|x';x,a)$ is not conventional in the sense that we directly control the action $a'$ conditioned on the previous state and action along with the state $x'$.  While non-standard, this is an expanded modeling capacity and allows for the possibility of introducing constraints on the actuator transition structure in the passive prior $p_a(a'|x';x,a)$.  Furthermore, under the assumption that $p_a$ has the same distribution over $a'$ for all state-action pairs $(\tilde{x},a)$ with the same previous state $\Tilde{x}$, $u_a$ will be the same for each pair $(\tilde{x},a)$ because the desirability function $z$ is only a function of $x'$ and $a'$ in \eqref{eq:salmdp_pol}.  Therefore, under this condition the previous state variables $x$ and $a$ are uninformative and the policy will only depend on $x'$, reducing down to $u_a(a'|x')$, which is simply the form of a standard stochastic policy.

\begin{algorithm2e}[h]
\DontPrintSemicolon
\SetKw{return}{return}
\SetKwRepeat{Do}{do}{while}
%\SetKwFunction{assume}{assume}
%\SetKwFunction{isf}{isFeasible}
\SetKwData{conflict}{conflict}
\SetKwData{safe}{safe}
\SetKwData{sat}{sat}
\SetKwData{unsafe}{unsafe}
\SetKwData{unknown}{unknown}
\SetKwData{true}{true}
\SetKwInOut{Input}{input}
\SetKwInOut{Output}{output}
\SetKwFor{Loop}{Loop}{}{}
\SetKw{KwNot}{not}
\begin{small}
	\Input{saLMDP $\mathscr{M}= (\mc X, \mc A, p_x, p_a, q)$,~convergence constant $\epsilon$}
	\Output{Optimal Policy: $u^*(a'|x';x,a)$; Desirability function: $z$}
	Construct $P$ as the passive Markov chain matrix for $p_{xa}$\;
	Define non-terminal and terminal state indices, $\mc N, \mc T$\;
	Define $\mathbf{z}_{\mc T} = \exp(-q(x_\mc T))$\;
	$Q \leftarrow \text{diag}(\exp(-\mathbf{q}))$\;
	Initialize $\mathbf{z}$ to a one-hot vector with a $1$ on the terminal state.\;
	$delta \leftarrow \infty$\;
	\#Z-Iteration\;
	\While{$delta > \epsilon$}
	{
	    $\mathbf{z}_{\mc N,\text{new}} \leftarrow QP_{\mc N\mc N}\mathbf{z}_{\mc N} + QP_{\mc N\mc T}\mathbf{z}_{\mc T}$\;
	    $delta \leftarrow sum(abs(\mathbf{z}_{\mc N,\text{new}} - \mathbf{z}_{\mc N}))$\;
	    $\mathbf{z}_{\mc N} \leftarrow \mathbf{z}_{\mc N,\text{new}}$\;
	}
	$u^*_{a} \leftarrow \frac{p(a'|x',x,a)z(x',a')}{\mathfrak{N}[z^*](x',x,a)} $

\end{small}
\caption{saLMDP$\_$solve (Deterministic $p_x$)}
\label{algor1}
\end{algorithm2e}

It is also important to note that, similar to the LMDP, the saLMDP has a trajectory likelihood formula for a given state-action trajectory $\mathbf{xa}:=(x,a)_{t_0},(x,a)_{t_1},...,(x,a)_{T}$:
\begin{align}
Pr(\mathbf{xa}|u_a)=\frac{z(x_{T},a_{T})}{z(x_{t_0},a_{t_0})} \prod_{t=0}^{T-1} \exp (-q(x_{t},a_{t}))p_a(a_{t+1}|x_{t+1};x_t,a_t)p_x(x_{t+1}|x_t,a_t).\label{eq:saLMDP_likihood}    
\end{align}
While this will not be discussed in depth in this work (see \ref{likelihood_1} for derivation) it can be used for intent inference in future work, especially in the context of non-Markovian TLMDP tasks.

Note that, as the cost of a state-action in $\mathbf{q}$ approaches infinity, its negatively exponentiated counterpart, represented as a diagonal entry in $Q$, approaches zero. Thus, zero-entries on the diagonal effectively edit the transition structure of $QP_{xa}$, shaping the backwards propagation of desirability when computing \eqref{eq:salmdp_eig} with z-iteration, suppressing state-actions which would violate the infinite-cost transitions. Todorov \citeyear{todorov2009efficient} showed that in the first-exit formulation, $\mathbf{z}$ could approximate shortest-paths with arbitrary precision in the limit of driving the non-terminal state costs arbitrarily high towards infinity, i.e. $S(x) = \lim_{c\rightarrow \infty} \frac{v_c(x)}{c}$, where $S(x)$ is the shortest-path distance from $x$ to the goal state and $c$ is the non-terminal state cost. In this paper, obstacles are represented with infinite cost, non-terminal states have an arbitrarily high constant cost (resulting in zero desirability), and the terminal (goal) state-action has zero cost. Therefore, the saLMDP objective with this cost structure is a discrete reach-avoid shortest-path first-exit control problem, avoiding obstacles states while controlling to a terminal state-action. These policies are said to be \textit{goal-conditioned}, as we can explicitly interpret the policy by the goal's grounded state-action, rendering the policies \textit{remappable} for new tasks.

\subsection{Boolean Ordered Goal Task}\label{section:BOGtask}
\tom{For our agent to achieve non-Markovian tasks, it will need a state-space to record the progress of the tasks and specify objectives and rules. Consider a goal state-vector $\bs\in \Sigma$, which encodes a binary vector.  A transition kernel on the space of binary vectors $\Sigma$ can be encoded by a deterministic transition kernel $P_{\boldsymbol{\sigma}}(\bs'|\bs,\alpha_\dg)$, where the goal variable $\alpha_\dg \in \{\alpha_{\epsilon},\alpha_{1},\alpha_{2},...\}$ is an action that will flip the $\dg^{th}$ bit of the binary vector (where $\alpha_{\epsilon}$ does not flip a bit). For a binary vector off length three we would have the transition, $$(0,1,0)\xrightarrow{\alpha_3} (0,1,1),$$ under the kernel $P_{\boldsymbol{\sigma}}$, where a task might be to flip all of the bits to $(1,1,1)$}

\tom{If we have a kernel to move around the space $\Sigma$, we will need a task formalism to be able to specify the unacceptable transitions within the binary vector space as \textit{constraints}, and the \textit{acceptable final states} in the binary vector spaces constituting the task. A rule on space $\Sigma$ can be made to dictate which bits can be flipped conditioned on other bits in the vector. For example, we will formalize a \textit{constraint} $c(\sigma_i^0,\sigma_j^1)$ that specify that the bit $i$ of $\bs$ cannot be flipped from $0$ to $1$ (indicated by the superscript) if the $j$ bit is currently $1$:
$$(0,1,0)\not\to(0,1,1), \quad \text{If:} \quad c\xunderbrace{(~\sigma_3^{0}~ }_{\mathclap{\text{Control Bit}}},\xoverbrace{ ~\sigma_2^{1}~)}^{\mathclap{\text{Condition Bit}}}.$$
However, since goals can be associated with features in the world, we will want specify these constraints on the features first, which will then be inherited by the constraints on goals. For example, if we want the agent to complete red goals before blue goals, we will need to specify constraints on the these feature (e,g. $c(\text{Red}^0,\text{Blue}^1)$), which will be inhered by all goals that are mapped from a set of features under a feature function, where a goal's feature set $\bp = \{\psi_1,...,\psi_N\}$ is just a set of features associated with a goal inherited from the lower-level state-space $\mc X$. Lastly, because the binary vectors encode ones and zeros, we can specify which subset of $\Sigma$ are the accepting set by encoding them as Boolean statements. For example, we will want to express that goal 1 AND goal 2, OR goal 1 AND goal 3 will satisfy the task, which we can express as a set of accepting states. We now define a Boolean ordered goal task, which expresses these rules and tasks specifications, and elaborate on the specifics of the formalism:}

% We also will outline a system for encoding task rules into the cost function which will force the optimal policy to respects specified ordering constraints on the completion of goals.
\begin{definition}[BOG task]\label{def:BOG}
    A Boolean Ordered-Goal task (BOG task) is defined as the tuple $\mathscr{T}=(\Sigma, \mc A_{\sigma}, \mc C_\Psi, \Psi, H_{\alpha}, B)$, where $\Sigma$ is a the set of all $2^{|\mc A_{\sigma}|}$ binary vectors indicating sub-goal completion, $\mc A_{\sigma}$ is a set of goal variables which are \textit{actions} on $\Sigma$, $\mc C_{\Psi}$ is a set of constraints on features, $\Psi$ is a set of features, and $H_{\alpha}: 2^\Psi \rightarrow \mc A_{\sigma}$ is a bijective feature function which assigns sets of features to a goal, and $B$ is a Boolean logic statement used to form a set $\Sigma_{\mc T}$ of terminal task-completion states.
\end{definition}

\begin{figure}[b!]
\centerline{\includegraphics[scale=0.9]{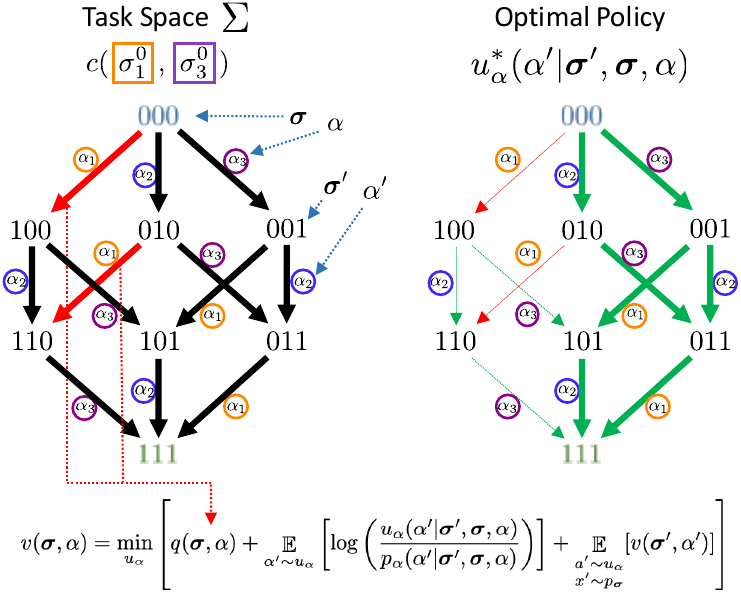}}
\caption{A BOG Task is shown for our running example where the vertices of the cube are binary state vectors $\bs$ in $\Sigma$.  We replaced the axe, water, and wood symbols with colors appended to the associated goals for cleaner visual presentation. The edges are labeled by goal-actions $\alpha_\dg$ which induce transitions between states via the kernel $P_{\boldsymbol{\sigma}}(\bs'|\bs,\alpha)$.  The Red lines denote constraint relations on goals, which prevent a goal from being carried out that violates the constraint: purple before orange.  Here, the orange goal $\alpha_1$ can not be used to flip the first bit from $0$ to $1$ when the third (purple) bit $\sigma_3$ is $0$.  These constraints can be encoded in the cost function which is used to derive the policy $u^*_\dg$.  The right cube shows the dynamics of the policy, where the thick green arrows are potential transitions under the optimal policy $u^*_\dg$ to the terminal state $\bs_{\mc T} = 111$, and the thin arrows will not be traversed because there isn't an efficient path that doesn't violate the constraints.}
\label{fig:bog}
\end{figure}

Here, $\mc C_\Psi$ is a set of task rules in the form of constraints $c(\cdot,\cdot)$, on features $\psi \in \Psi$, 
% e.g. $c(\psi_1, \psi_2) \in \mc C_\Psi \implies \psi_1 \prec \psi_2,~\psi_1,\psi_2 \in \Psi$, 
$\mc A_{\sigma}$ is a set of goals functioning as actions that induce transitions on $\Sigma$, a set of binary vector states $\bs$. \tom{Notationally, for the constraints, we will use the convention $\bs(j)=\sigma_j$ is the $j^{th}$ binary state that takes Boolean values $0$ and $1$, denoted $\sigma_j^0$ and $\sigma_j^1$.}

The function $H_{\alpha}$ maps \textit{sets of features} $\bp$ onto a goal variable, where a set of features $\bp\in 2^{\Psi}$ and a single feature $\psi \in \Psi$. Implicit to the BOG task definition is a bijective map $H_{\dg}$ between a goal $\alpha_i$ and the bit $\sigma_i=\bs(i)$ to be flipped: $H_{\dg}(\alpha_i)\rightarrow \sigma_i$, such that the goal and binary state-space share the same index $i$. Therefore, because of this bijective relationship between goals and bits, the bit can inherit the feature set through the composition $H_{\dg \alpha} = H_{\dg} \circ H_{\alpha}$, where $H_{\dg \alpha}(\bp)\rightarrow \sigma_i$. If our objective is to complete all goals, $\bs = [0,1,0,1] \in \Sigma$ indicates that $\alpha_2, \alpha_4 \in \mc A_{\sigma}$ have been activated to flip the goal bits and two other actions $\alpha_1, \alpha_3 \in \mc A_{\sigma}$ need to be induced to flip the other two. With these primary objects, we can derive a transition kernel $P_{\boldsymbol{\sigma}}(\bs'|\bs, \alpha)$ which specifies the dynamics of task-space under given sub-goals acting as actions on the space, with the restriction that $\alpha_\dg$ can only flip the $\dg^{th}$ bit of the vector, $\bs(\dg)$. 

The task-space transition kernel $P_{\boldsymbol{\sigma}}$ encodes the task-space hypercube (see Fig. \ref{fig:bog}). To impose structure on the task space in the form of rules (precedence conditions), the feature system gives the task states and actions richer semantics. For example, it might be the case that an agent needs to get an axe in order to chop wood, but it can chop wood in multiple places in the state space. Thus, we need features sets that allow us to group low-level states and actions which share the same semantic meaning, and express these constraints on both the low-level space and high level space. We will generically introduce constraints at the level of features which will be inherited both in the task space and the  low-level state-space which drives the task dynamics with calls to a policy. In this section, we focus on developing the feature system for the high-level space, but the connection low-level space will be made in Section~\ref{Section:Grounding}.

If we would like to specify that a bit associated with one feature $\psi_i$ can only be flipped conditional on the value of a associated with another feature, $\psi_j$, then these constraints are imposed directly on to the cost function. The feature relation $c(\psi_k^{\bd} , \psi_l^{\bd})$ takes in features, where the dot $\bd$ can be either $0$ or $1$ indicating the value of the bit associated with the feature. Therefore we have the following possible relations, $c(\psi_k^0, \psi_l^0)$, $c(\psi_k^0, \psi_l^1)$, $c(\psi_k^1, \psi_l^0)$, $c(\psi_k^1, \psi_l^1)$. 

We can then transfer the feature constraints onto the task space by defining an induced constraint set on the binary states,
\begin{align*}
    \mc C_{\Sigma}:=\{c(\sigma_i^{\bd},\sigma_j^{\bd})~|~c(\psi_k^{\boldsymbol{\cdot}},\psi_\ell^{\bd})\in \mc C_\Psi,~\exists (\psi_k,\psi_\ell)\in H^{-1}_{\alpha\dg}(\sigma_i) \times H^{-1}_{\alpha\dg}(\sigma_j)\},
\end{align*}
where $H^{-1}_{\alpha\dg} := H^{-1}_\alpha\circ H^{-1}_\dg$ such that $H^{-1}_{\alpha\dg}(\sigma_i)\rightarrow \bp$. The new set $\mc C_{\Sigma}$ is computed by adding bit relations $c(\sigma_i^{\bd},\sigma_j^{\bd})$ to the set if the bits have assigned features which exist in the set of feature relations $\mc C_\Psi$. Feature relations add an additional level of symbolic reasoning into the architecture, enabling us to make statements such as: ``complete the purple sub-goals before the orange sub-goals", or ``chop wood with an axe in addition to collecting water", as depicted in Fig. \ref{fig:bog} and \ref{fig:task_example} by red arrows.

Importantly, the precedence constraints aren't defined to apply to a history or sequence of bit-flips the agent has performed in the past, instead they are meant to refer to the state of the binary vector. When reversible bit flips ($1\rightarrow 0$) are disallowed, the constraints must necessarily act as precedence constraints on the agent's generated goal sequences, as there will be no way to repeat the same goal. The constraint, as it will be used to define the cost function $q_{\sigma \alpha}$ in \eqref{eq:BOG_order_cost}, is not symmetrical. Rather, $c(\sigma_i^{\bd},\sigma_j^{\bd})$ means that we cannot flip $\sigma_i^{\bd}$ with $\alpha_i$ when the vector $\bs$ has value $\bs(j)=\sigma_j^{\bd}$.
\begin{align*}
c\xunderbrace{(~\sigma_i^{\bd}~ }_{\mathclap{\text{Control Bit}}},\xoverbrace{ ~\sigma_j^{\bd}~)}^{\mathclap{\text{Condition Bit}}} \implies q_{\sigma \alpha}(\bs,\alpha_i) = \infty \quad\quad 
\text{when: }\quad\bs(i) = \sigma_i^{\bd},~ \bs(j) = \sigma_j^{\bd}.
\end{align*}

% \begin{align}
% c\xunderbrace{(~\alpha_i~|}_{\mathclap{\text{Control}}}\xoverbrace{\sigma_i, \sigma_j~}^{\mathclap{\text{Conditional}}}) 
% \end{align}

The notation above is written abstractly with the placeholder variable $\sigma_i^{\bd}$, but formally it takes binary states set to one of the Boolean values, leading to four possible constraints: $$c(\sigma_i^0,\sigma_j^0),~ c(\sigma_i^0,\sigma_j^1),~c(\sigma_i^1,\sigma_j^0),~c(\sigma_i^1,\sigma_j^1).$$

% We will use bar-notation, $\bar{\dg}_2$, in the conditional to refer to when we want to condition on $\bs(2)$ being $0$, and bar-notation on $\dg_1$ when we want to restrict bit flips of the first argument from one to zero, $\bs(1): 1 \rightarrow 0$. Thus, the four kinds of constraints are $c(\dg_1, \dg_2)$, $c(\bar{\dg}_1, \dg_2)$, $c(\dg_1, \bar{\dg}_2)$, $c(\bar{\dg}_1, \bar{\dg}_2)$.

A cost function on the task space $q_{\sigma}(\bs,\alpha)$ can be derived directly from the constraints. 
The cost can then be defined as $q = q_{\sigma \alpha} + q_{\sigma}$, where:
\begin{align}
    q_{\sigma \alpha}(\bs,\alpha_i) &= \{\infty~~ \mathbf{if}:~\exists c(\sigma_i^{\bd},\sigma_j^{\bd})\in \mc C_{\Sigma} ~s.t.~ (\bs(i)=\sigma_i^{\bd})\land (\bs(j)=\sigma_j^{\bd}) ;~\mathbf{else}:~ const.\},\label{eq:BOG_order_cost}\\
    q_{\sigma}(\bs) &= \{0~~\mathbf{if}:~ \bs = \bs_{\mc T}\in \Sigma_{\mc T};~\mathbf{else}:~ const.\}.
\end{align}
The function $q_{\sigma \alpha}$ penalizes ordering violations on the bit-flips of $\bs$, and $q_{\sigma}$ is a constant cost on task states with the exception of the task-space terminal state $\bs_{\mc T}$.

The terminal state set for the first-exit problem can be specified by any Boolean logic statement $B$ decomposed into disjunctive normal form (DNF):  $$B \implies w_1 \lor w_2 \lor ... \lor w_n \quad w \in \mc W,$$ where a clause $w := \sigma_i^{\bd} 
 \land \sigma_j^{\bd} \land ... \land \sigma_k^{\bd}$ is a conjunction of bits corresponding to the sub-goals to be achieved (e.g. $w = \sigma_i^1 \land \sigma_j^0$ means that bit $i$ must be $1$ and bit $j$ must be $0$ to be true; for compactness, sometimes we will use the convention that $\sigma_i$ and $\bar{\sigma}_i$ (bar denoting $\lnot$, \texttt{NOT}) denote True and False, and we will drop the $\land$ so the above conjunction $\sigma_i^1 \land \sigma_j^0 \equiv \sigma_i\bar{\sigma}_j$). The final state set $\Sigma_{\mc T}$ is the set of binary vectors encoding the conjunctive clauses $\mc W$.  .

Since the objects of a BOG task $\mathscr{T}$ can be directly used to define a transition kernel and a cost function, much like Definition \ref{def:saLMDP}, we could introduce a passive transition dynamics $p_{\alpha}(\alpha'|\bs',\bs,\alpha)$ and solve an saLMDP policy in the task space, $u_{\alpha}^*(\alpha'|\bs',\bs,\alpha)$, which satisfies our constraints, as illustrated in Fig. \ref{fig:coupled}.A. However, synthesizing a policy only in the task space is not sufficiently useful for an agent. To make task-states and actions useful representations, we show how to map goal-actions to the agent's underlying controllable space, in this case, the base space $\mc X \times \mc A$ in section \ref{Section:Grounding}.

\subsubsection{Connections to LTL Control}
\tom{Naturally, these non-Markovian BOG tasks appear related to Linear Temporal Logic (LTL) control tasks \cite{hasanbeig2020deep, certified}.  A major difference is that our agent is trying to control the state of a higher-level state-space $\Sigma$ without encoding hard constraints on the low-level dynamics on $\mc X$ as one might with LTL. Furthermore, with BOG tasks we are also minimizing the length or cost of controlling to the final states, where as obstacles to be avoided are encoded into a cost function. LTL tasks on the other hand often use automata to constrain acceptable trajectories on $\mc X$ including obstacle states, but do not necessarily minimize costs. However, BOG tasks can represent Co-Safe LTL fragments. The precedence constraint $c(\sigma^0_i,\sigma^1_j)$ is equivalent to $[\sigma_i^0, x]  \Diamond [\sigma_j^1, x] ~ \forall x \in \mc X$ (where $\Diamond$ is ``eventually"). Boolean formulas encoded in terminal states can express fragments as well. For example, encoding $\sigma^1_i \land \sigma^1_j$ into the terminal state set $\Sigma_f$ is equivalent to  $\Diamond(\sigma^1_i \land \sigma^1_j$), and similarly $\sigma^1_i \lor \sigma^1_j$ is equivalent to $\Diamond (\sigma^1_i \lor \sigma^1_j)$.
}

\subsection{Feature Functions and Affordance Functions}\label{Section:Grounding}

In order to handle the mapping between a low-level state-space and a high-level transition kernel, we introduce an \textit{affordance function}, previously mentioned in Eq. \eqref{eq:product}, which mediates the interaction. An affordance function, $F: (\mc X \times \mc A) \times \mc A_{\sigma} \rightarrow [0,1]$ is a deterministic probability distribution $F(\alpha|x,a)$ which associates a state-action pair with a high-level goal-action variable $\alpha$ in $\mc A_{\sigma}$. We use the name \textit{affordance function} because the higher-level system ``affords" certain actions from given lower-level state spaces mediated through features, which we will soon explain. We borrow the term from Gibson's theory of affordances \cite{gibson1977theory}, which has already been introduced to RL and the Options Framework, where \textit{affordance sets} are used to group state-actions that induce intended state-transitions \cite{khetarpal2020can,khetarpal2021temporally,xu2021deep}. The affordance \textit{function} here can be used to group state-actions in terms of the higher-level transition dynamics they induce. We will often refer to a high-level task-space $\Sigma$ as being \textit{grounded} to a low-level state-space $\mc X$ through the affordance function, and a state-action which induces a high-level variable $\alpha_{\dg}$ to flip a bit is a \textit{grounded state-action}.

For an affordance function to be well-formed for a Boolean task, it must include a ``null action" $\alpha_\epsilon\in \mc A_{\sigma}$ which has no influence on the bit-vector $\bs$ when the agent is at any non-goal state-action $(x,a)_g$ not in the set of goal state-actions $\mc X\mc A_{\mc G}$ that flip bits. This ensures that each state-action pair $(x,a)$ is linked to some high-level action $\alpha$. 
% We will not show $\alpha_\epsilon$ in the figures, and just assume that it is included in $\mc A_{\sigma}$ and $P_\sigma$ is appropriately defined.
\subsubsection{Feature Functions}
In the problems that we consider, it is natural to parameterize the affordance function in terms of features so that we can 1) associate low-level state-actions with features (such as attributes or items), and 2) map sets of features to goal variables independently of the low-level state-space.  For example, it is useful to be able to map many state-actions to the same feature (such as the color ``Red" or the item ``water") and have precedence rules based on color. Features are simply ways of grouping sub-goals into subsets by a feature such as the color of the state. 

In the BOG-task definition \eqref{def:BOG}, we already introduced a feature function $H_{\alpha}$ for mapping features to actions. We also will introduce an $xa$-to-feature function $H_{\psi}:\mc X \mc A \rightarrow 2^{\Psi}$ which maps state-actions to features.  Like $F$, the functions $H_{\alpha}$ and $H_{\psi}$ are deterministic, and we will often write them as deterministic probability distributions, $H_{\alpha}(\alpha|\bp)$ and $H_{\psi}(\bp|x,a)$, for visual clarity.

\begin{figure}[h!]
    \centering
    \includegraphics[width=\linewidth]{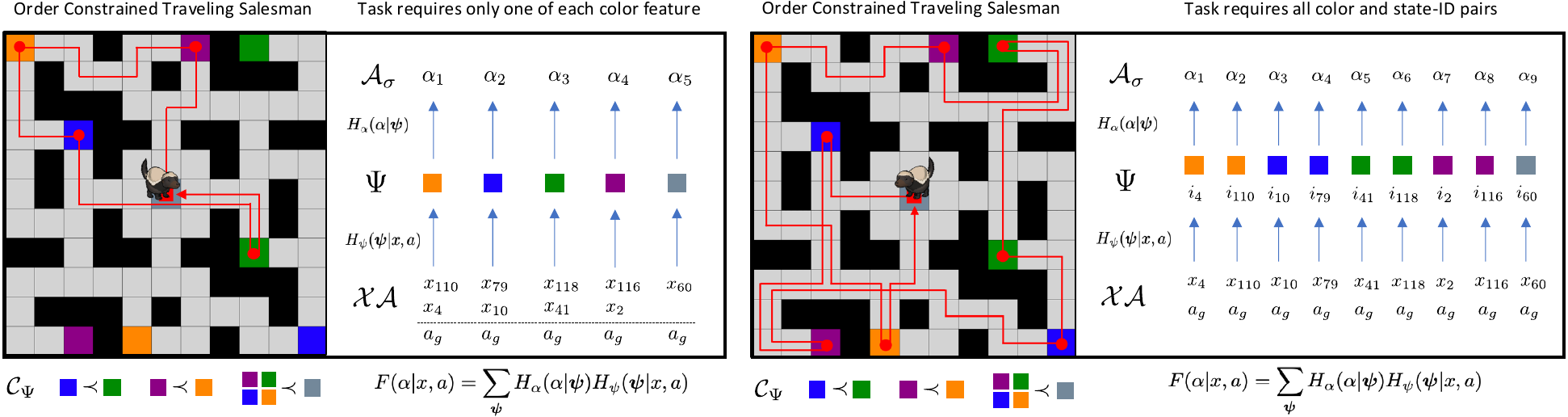}
    \caption{This traveling salesman example from the experiments section shows an agent that must complete all goals with precedence constraints on the colors. The final bit-vector state is the state of all ones, where the agent achieves all goals and returns back to the grey square. The expression $\psi_i\prec \psi_j$ indicates $c(\psi_i^0,\psi_j^1)$, meaning that all goals of color $\psi_i$ can not be flipped when goals of color $\psi_j$ are $1$. The left panel show how mapping a single color feature to a high-level action creates a problem where the agent needs to achieve \textit{one} of each color, but any low-level state-action is acceptable as a goal. The right panel shows that a state-ID (e.g. $i_{41}$) can be included into the feature sets that map to high-level actions so that there is a bit for each state-ID that needs to be completed.}
    \label{fig:traveling}
\end{figure}

\begin{figure}[t]
\centerline{\includegraphics[scale=0.75]{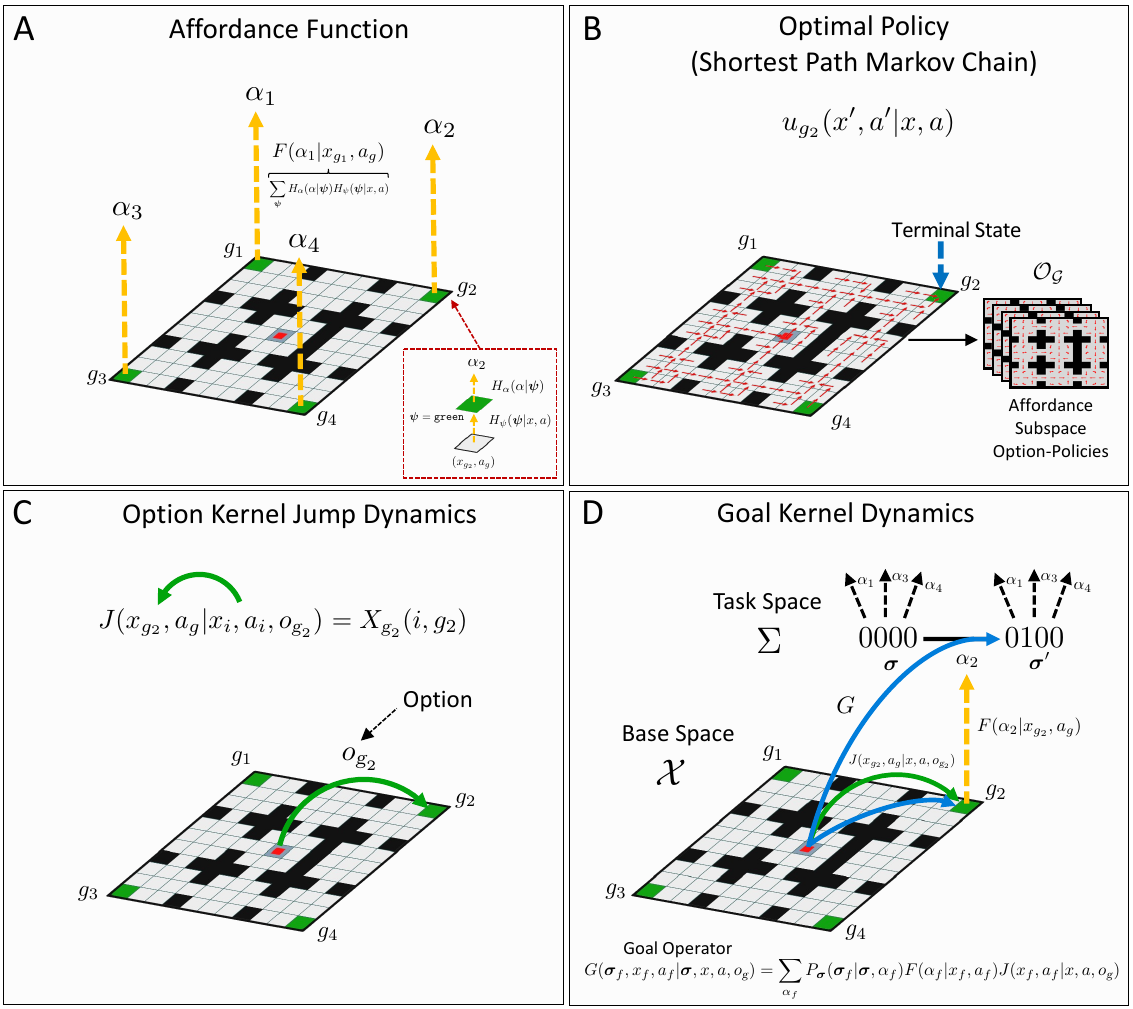}}
\caption{For our running example, we have constructed the following functions, policies, and transition kernel: \textbf{A}: Affordance functions $F$ map state-actions to goals. \textbf{B}: terminal state-actions used to define members of an ensemble of cost-minimizing policies. \textbf{C}: Optimal policies in an ensemble can be used to compute a option kernel $X_{\dg}$. \textbf{D}: The option kernel and affordance functions are combined with the task dynamics $P_{\boldsymbol{\sigma}}$ to create the option kernel~$G$.}
\label{fig:coupled}
\end{figure}

The feature mapping is illustrated in Fig.\ref{fig:coupled}.A where the a low-level state is associated with the color green, which maps to an $\alpha$ variable. The mapping is also illustrated in Fig. \ref{fig:traveling}, which shows a simple traveling salesman problem where the agent needs to complete all goals, but the ordering constraints are specified on the colors of the goals, \textit{not} the state-indices. As mentioned with we introduced the BOG task, an advantage of using this system is that it allows us to be more expressive in our ordering rules that we specify on the sets of features.  If we want the agent to complete all of the goals associated with state-actions $\mc X \mc A_g$, then we simply associate the low-level state-action with a unique identifier feature (index) $Id$; this allows us to map the unique feature to a unique goal.  For example $$H_{\psi}((x,a)_i) \rightarrow \{i,Red\},\quad \text{and}\quad H_{\alpha}(\{i,Red\}) \rightarrow \alpha_{i,red},$$ produces an affordance function $F$, which would map $F((x,a)_i)\rightarrow \alpha_{i,red}$, where $\alpha_{i,red}$ is the only goal variable associated with $(x,a)_i$. Alternatively, if we simply map $(x,a)_i$ to a color without an ID-feature and then mapped the color feature to a goal, e.g. $$H_{\psi}((x,a)_i) \rightarrow \{Red\},\quad \text{and}\quad H_{\alpha}(\{Red\}) \rightarrow \alpha_{red},$$ then $F$ would be non-injective by extension of $H_{\psi}$ being non-injective, and $F$ would map all state-actions with the feature Red to the same goal $\alpha_{red}$. This is useful when there are many state-actions on the low-level state-space which would satisfy the goal, as we avoid needing to represent extra goal variables in the task-space. However, when using unique Id-features, we can state how many goals of a given feature type need to be completed by defining the Boolean statement $B$, which dictates the terminal states $\Sigma_{\mc T}$. 

These two scenarios provided here are both illustrated in the figures of this paper.  The first example is shown in the traveling salesman problem which requires the agent to visit all goals with constraints on goal color in Fig. \ref{fig:results_fig}.A (page \pageref{fig:results_fig}), and the second scenario is shown in the office environment example where the robot only needs to get one of many possible cups of water to bring to the central office, shown in Fig. \ref{fig:office_world} (page \pageref{fig:office_world}).

\subsubsection{Affordance Function}
We can define the affordance function $F$ in terms of the composition of $H_{\alpha}$ and $H_{\psi}$.
% \begin{align*}
%     % F := H_{\alpha} \circ H_{\psi}: \mc X \mc A \rightarrow \mc A_{\sigma}.\\    
%     F(\alpha|x,a):=\sum_{\bp}H_{\alpha}(\alpha|\bp)H_{\psi}(\bp|x,a),
% \end{align*}
For the grid-world examples in Fig. \ref{fig:coupled}, the base action-space $\mc A \allowbreak=\allowbreak \{a_{u},\allowbreak a_{d},a_{r},a_{l},a_{n},a_{g}\}$ includes an action $a_g$, which is a special neutral action for completing a sub-goal in addition to five standard directional actions: up, down, right, left and neutral.  We can create a deterministic affordance function, 
\begin{align*}
F(\alpha|x,a)=\sum_{\bp}H_{\alpha}(\alpha|\bp)H_{\psi}(\bp|x,a).
\end{align*}
% $$F(\alpha|x,a):=\delta(\alpha,h(x,a)),$$
% defined by a Kronecker delta function, returning $1$ if the arguments match, $0$ otherwise. Note that $F$ is simply the affordance function with deterministic probability notation. 
Consider an saLMDP transition kernel $p_{xa}(x',a'|x,a)$. We can now compose this kernel with the affordance function and BOG-task kernel $P_{\boldsymbol{\sigma}}$ to obtain the following joint passive dynamics $p_{\bs xa}$, called the \textit{full passive dynamics}:
$$p_{\bs xa}(\bs'\allowbreak\allowbreak,x'\allowbreak,a'|\bs,x,a)\allowbreak=\sum_{\alpha}P_{\boldsymbol{\sigma}}(\bs'|\bs,\alpha)F(\alpha|x,a)p_{xa}(x',a'|x,a).$$
% \begin{align}
% P_{\bs \dg xa}(\bs',\alpha',x',a'|\bs,\alpha,x,a)=T(\bs'|\bs,\alpha')P_{\dg xa}(\alpha',x',a'|\alpha,x,a)
% \end{align}
Note that this transition kernel on a space of size $|\Sigma \times \mc X \times \mc A|$. Since $\Sigma$ grows exponentially in the number of sub-goals, i.e. $2^{|\mc A_{\sigma}|}$, full transition kernel can introduce considerable space and time complexity into the computation of the optimal policy. However, the added complexity can be mitigated by creating specialized transition kernels called \textit{option kernels} which summarize the resulting dynamics of the a policy from its initial state to the state-action of sub-goal completion. In doing so, we can avoid representing non-terminal states of $\mc X$ and compute a  solution in a lower dimensional subspace of the full space (discussed in Section~\ref{section:TLMDP}). However, first we discuss how to define options from saLMDP solutions.

\subsubsection{Defining Options of from saLMDP policies}
Now we can create a set of saLMDP problems for many goals and turn their policies and into a set of options. Let $\mathscr{M}^{N}=\{\mathscr{M}_{\dg_1},...,\mathscr{M}_{\dg_N}\}$, be a set of $N$ first-exit saLMDPs, with $\mc G = \{\dg_1,...,\dg_N\}$ being a set of goal indices for each saLMDP. Each saLMDP has a cost function $q_{\dg_i}$ for a termination state for a state-action in the \textit{low-level goal state-action set} $\mc X\mc A_{\mc G}$ which induce bit flips on $\Sigma$. That is:
$$\mc X\mc A_{\mc G} = \{(x,a)_{g_i}:F(\alpha_i|(x,a)_{g_i})=1; ~\exists \alpha_i \neq \alpha_{\epsilon}\},$$
and each saLMDP is defined as:
$$\mathscr{M}_{\dg_i} = (\mc X, \mc A, p_{xa},p_a, q_{\dg_i}), \quad \dg_i \in \mc G. $$
We can then derive a set of policies and desirability functions for each saLMDP:
% $$\mc U_{\mc G} = \{u^*_{g_1},...,u^*_{g_N},...\},$$  $$\mc Z=\{z_{\dg_1},...,z_{\dg_N}\}.$$
\begin{align*}
    \mc U_{\mc G} = \{u^*_{\dg_1},...,u^*_{\dg_N}\},\quad \mc Z_{\mc G} = \{z_{\dg_1}^*,...,z_{\dg_N}^*\}.
\end{align*}
\tom{We can now define our options from the set $\mc U_{\mc G}$ and $\mc Z_{\mc G}$. Recall, an option is typically defined as $o=(\pi, \beta)$ for the policy $\pi$ and termination function $\beta$. In this paper, for a saLMDP policy $u$, the deterministic termination function $\beta_{\dg}(x,a)$ is defined with the desirability function $z_{\dg}$ which encodes states in the termination state-actions for the first-exit problem as zeros for goal-failure events and $\exp(-c_{\dg})$ (with terminal cost  $c_{\dg}$) for goal-success events:}
\begin{align}\label{eq:termination}
    \beta_{\dg}(x,a) = \begin{cases}
        1 &\text{if} \quad (z^*_\dg(x,a) = 0) \lor (z^*_\dg(x,a) = \exp(-c_{\dg}))\\
        0 &\text{otherwise}
    \end{cases}, 
\end{align}
where $\beta_{\dg}$ is a function of $z_{\dg}$, which is suppressed and represented as $\dg$ on the subscript. Thus, a set of options $\mc O$ can be defined from the ensembles $\mc U_{\mc G}$ and $\mc Z_{\mc G}$, given as:
\begin{align*}
    \mc O = \{o_{\dg_1},...,o_{\dg_N}\} ~~\text{where: } o_{\dg_i} = (u^*_{\dg_i}, \beta_{\dg_i}),\quad u^*_{\dg_i}\in \mc U_{\mc G},~z_{\dg_i}^*\in \mc Z_{\mc G}.
\end{align*}
\tom{We can now use a set of options to construct an option kernel, which allows us to chain together options to satisfy Boolean tasks.}

\subsection{Option Kernel}\label{Section:JumpFeas}
Recall the joint distribution $u_{xa}$ over the base space $\mc X \times \mc A$ in Section~\ref{section:saLMDP} that expresses the control policy. Once a control policy is optimized with a saLMDP, a Markov chain is induced. Consider an optimal shortest-path first-exit Markov chain matrix $U_{\dg}$, induced by applying the distribution ${u_{xa}}_g^*(x',a'|x,a) := {u^*_{a}}_{g}(a'|x';x,a)p_x(x'|x,a)$ that controls to the goal $\dg$'s state-action pair $(x,a)_g$\footnote{We use stylized `$\dg$' to refer to goal variables and `$g$' to refer to the goal index on a state-action (Fig. \ref{fig:coupled}.A)}. To simply notation we will sometimes use $(x_g,a_g)$ instead of $(x,a)_g$. \tom{Let $\mc T =\{(x,a) : \beta(x,a)=1\}$ be the terminal state-actions, which includes both goal success and failure terminations from the option's termination function $\beta$ in Eq. \eqref{eq:termination}. Let $\mc N = (\mc X \times \mc A) \setminus \mc T$ be the set of non-terminal state-actions. We represent the Markov chain $u_{xa_g}$ with enforced termination conditions as the Markov matrix $U_{\dg}$ which, using the sets $\mc N$ and $\mc T$, can be written as a block matrix,}
\begin{align*}
\centering
    U_{\dg} = \begin{bmatrix}
        A_{\mc N \mc N}& B_{\mc N \mc T} \\ 
        0_{\mc T \mc N}& I_{\mc T \mc T} 
        \end{bmatrix},
\end{align*}
where $A_{\mc N \mc N}$ is a sub-matrix of $U_{\dg}$ for non-terminal to non-terminal state dynamics, and $B_{\mc N \mc T}$ is sub-matrix of non-terminal to terminal state dynamics. By taking an arbitrary power, $U_{\dg}^t$, the elements represent the probability of existing in a non-goal state-action or absorbing into one after $t$ time steps. The power of $U_{\dg}$ in the infinite limit is~\cite{bremaud2013markov}:
\begin{align*}
X_{\dg}=\lim\limits_{\mathclap{t \rightarrow \infty}}U_{\dg}^t=
\begin{bmatrix}
\lim\limits_{t \rightarrow \infty}A_{\mc N \mc N}^t& \lim\limits_{t \rightarrow \infty}\left(\sum_{\tau=0}^t A_{\mc N \mc N}^\tau\right)B_{\mc N \mc T} \\ 
0_{\mc T \mc N}& I_{\mc T \mc T} 
\end{bmatrix} = \begin{bmatrix}
0_{\mc N \mc N}& (I- A_{\mc N \mc N})^{-1}B_{\mc N \mc T} \\ 
0_{\mc T \mc N}& I_{\mc T \mc T} 
\end{bmatrix},
\end{align*}
where $(I-A_{\mc N \mc N})^{-1}$ is the fundamental matrix, which is the solution to the Neumann series, $\sum_{\tau=0}^\infty A_{\mc N \mc N}^\tau$. \tom{The equality between the block matrices hinges on the equation $\lim_{t \rightarrow \infty}A_{\mc N \mc N}^t=0_{\mc N \mc N}$ being true and whether $I-A_{\mc N \mc N}$ is actually invertible. Both are true if $U_{\dg}$ is an \textit{absorbing} Markov chain \cite{bremaud2013markov}, which requires all state-actions of $\mc N$ to be \textit{transient} under the dynamics of $U_{\dg}$---that is, there is a positive probability that a trajectory starting from any $(x,a)_i$ in $\mc N$ will never revisit it. In our case this is true because if a state-action in $\mc N$ were not transient (\textit{recurrent}) then it would be revisited an infinite number of times and have an infinite cost-to-go and zero desirability, which implies by Eq. \eqref{eq:termination} that it is in $\mc T$, not $\mc N$, resulting in a contradiction. All state-actions of $\mc N$ are therefore transient which implies that the spectral radius of $A_{\mc N \mc N}$ is less than one, and so the upper-left block is a matrix of zeros $0_{\mc N \mc N}$ and $I-A_{\mc N \mc N}$ is invertible. For a detailed explanation, see Appx. \eqref{sec:invertability}.}

The matrix $X_{\dg}(i,f)$ is a \textit{primitive option transition kernel} (or simply \textit{option kernel}) which maps, or \textit{jumps}, \underline{i}nitial state-actions $(x,a)_i$ to \underline{f}inal termination state-actions $(x,a)_f$ of a single option $o_{\dg}=(u_{xa_g},\beta_{\dg})$ \cite{precup1998theoretical,silver2012compositional}. We now discuss how we can compute many options and their primitive kernels to create an aggregated option kernel.

\subsubsection{Defining the Aggregated Option Kernel}

Using our set of $N$ options $\mc O_{\mc G}$, comprised of policies $\mc U_{\mc G}$ and desirability functions $\mc Z_{\mc G}$, we can define an aggregated \textit{option kernel} $J$ out of primitive option kernels $\mathscr{X}_{\mc G}=\{X_{g_1},...,X_{g_N}\}$:
\begin{align*}
    J((x,a)_f|(x,a)_i, o_{\dg}) = X_{\dg}(i,f),\quad X_{\dg} \in \mathscr{X}_{\mc G}, ~~ o_{\dg} \in \mc O_{\mc G}
\end{align*} 
The aggregated option kernel $J$ is a transition kernel \textit{jumps} the agent from an initial state-action to a terminal state-action with the probability that it ends there under the option, specified by the primitive option kernel $X_{\dg}$. 
% This operator can also be written as $J((x,a)_j|(x,a)_i, u_{xa_g})$ given that $o_\dg$ is simply an index variable--this will be important when we remap low-level solutions to new problems. 
\subsubsection{Defining the Option Kernel with Desirability Functions}
While $X_{\dg}$ is simple to compute, under the conditions of a deterministic transition kernel $p_x$ and a single terminal goal, we need not compute it because we can just use a desirability function. We show this with a simple theorem:
\begin{theorem}[Goal-state Hitting Probability Theorem]
    Let $\mathscr{M}$ be an saLMDP. Under the conditions of deterministic dynamics, $p_x \in \mathscr{M}$, and a single terminal state $x_g$, any state-action which has non-zero desirability will hit the goal with probability one under the optimal policy $u^*_{xa,g}$.\label{simple_thm}
\end{theorem}
% \begin{mytheo}{Goal-state Hitting Probability Theorem}{}
%     Let $\mathscr{M}$ be an saLMDP. Under the conditions of deterministic dynamics, $p_x \in \mathscr{M}$, and a single terminal state $x_g$, any state-action which has non-zero desirability will hit the goal with probability one under the optimal policy $u^*_{xa,g}$.\label{simple_thm}
% \end{mytheo}
\textbf{\textit{Proof:}} Recall that $z$ is a non-negative function. If the function $z(x,a)$ is equal to $0$ when evaluated on $(x,a)$, then the terminal state-action $(x_g,a_g)$ is not reachable under the policy $u^*_{xa,g}$ from $(x,a)$, as each trajectory would have infinite accumulated cost and $0$ desirability. This is because, even though $z$ is an expectation over all possible trajectories, if there were one trajectory generated by $u^*_{xa,g}$ which hit $(x_g,a_g)$, then $z$ would be finite and positive. Thus, $z(x,a)=0$ implies that the terminal state-action is not reachable. Because \eqref{eq:salmdp_pol} directly sets the distribution $u_{xa,g}^*$, and there is a deterministic mapping between the chosen action and the next state, we can see that if $z(x',a')=0$, then there is $0$ probability that $u^*_{xa,g}(a'|x';x,a)$ will choose \textit{any} action $a'$ from which $(x_g,a_g)$ is not reachable. It follows that if $z(x,a)>0$ and $u^*_{xa,g}$ only generates trajectories where all states-actions communicate with $(x_g,a_g)$, then all trajectories from $(x,a)$ must reach $(x_g,a_g)$. Therefore the probability that the agent reaches $(x_g,a_g)$ when $z(x,a)>0$ must be one.$\hfill\square$\\
% ($z(x,a)=0 \implies v(x,a)=\infty$, $z(x,a)>0 \implies v(x,a)\in \mathbb{R}^{+}$)

Following Theorem \ref{simple_thm} we need not compute the fundamental matrix under deterministic $p_x$ and a single terminal goal state-action $(x,a)_g$, rather we can simply use the logical inequality function to define the aggregated option kernel with the sets $\mc O_{\mc G}$ and $\mc Z_{\mc G}$: 
\begin{align*}
    J((x,a)_j|(x,a)_i, o_{\dg}) = \{1 \text{~if~} \big((z_\dg(i)>0)\land (j=g)\big) \lor \big((z_\dg(i)=0) \land (i=j)\big) ; ~~0 \text{~otherwise} \}.
\end{align*}
Thus, with this result we also establish a connection between the desirability function and the fundamental matrix through the definition of $J$ under determinism.
% \footnote{It is easy to see that even if there are multiple terminal states for a control problem, then: $X_{\dg}(i,g) = \{1 \text{~if~} (\mathbf{z}(i)>0)\land (x_g\in \mc T); 0 \text{~otherwise} \}$. This relates $z$ to the total probability that the agent hits any state in $\mc T$.}. 
\tom{For the remainder of the paper, our algorithm will assume that there is a single terminal state for each option, as a hierarchical decomposition proof will justify this choice.} We now show how $J$ can be combined with $F$ to construct a goal kernel.
\subsection{Goal Kernel}\label{section:goalop}

Now that we have defined the option kernel we can form the \textit{goal kernel}, $G$, which summarizes the resulting full state-action vector on the product-space after achieving the goal:
\begin{align}
    % &\kappa(\alpha',x',a'|x,a,o_{\dg}) = F(\alpha'|x',a')J(x',a'|x,a,o_{\dg}) \label{eq:feasibility},\\
    G(\bs_f,x_f,a_f|\bs,x,a,o_{\dg})=\sum_{\alpha_f}P_{\boldsymbol{\sigma}}(\bs_f|\bs,\alpha_f)F(\alpha_f|x_f,a_f)J(x_f,a_f|x,a,o_{\dg}).
    % \kappa(\alpha_f,x_f,a_f|x,a,o_{\dg}). 
    \label{eq:coupleddyn}
\end{align}
A goal kernels, $G(\mathbf{s}_f|\mathbf{s},o_{\dg})$ (where $\mathbf{s}=(\bs,x,a)$) can simply be thought of as the same as an option kernel, but for the entire product-space, and will allow us to compute an exact non-Markovian solution via backwards induction.

There are a couple of constraints on the form of the transition kernels and policies that are used to form $G$ (through $J$).  In order for the call of a policy to induce a single change in a $\bs \rightarrow \bs'$ transition, the policy must not be able to induce a different high-level goal variable through the affordance function $F$ en route to completing the target goal of the policy. This requires that the low-level ensemble policies either 1) exhibit an orthogonality condition on the affordance functions, meaning the action in $\mc X \mc A_g$ is unnecessary in order to produce cost- or path-minimizing policies on $\mc X$ for the base-policy $u$, or 2) that other possible goal state-actions are encoded as obstacles with an arbitrarily high cost so that they are avoided. If the orthogonality condition does not hold and we do not encode the alternative goals as obstacles, there would be a possible interference effect in which scheduling a policy would induce undesirable $\bs$-transitions which violate ordering constraints.

The second and third constraints are on $P_{\boldsymbol{\sigma}}$, which will require that $P_{\boldsymbol{\sigma}}$ is not a function of time (stationarity) and that the update for $P_{\boldsymbol{\sigma}}$ happens in the same time step as action selection. If $P_{\boldsymbol{\sigma}}$ is not stationary, then we can't readily perform dynamic programming. In such conditions,~\citeauthor{ringstrom2023reward} \citeyear{ringstrom2023reward} presents a forward solution for solving non-Markovian tasks with non-stationary goals and obstacles.

\subsection{Task-LMDP}\label{section:TLMDP}

With the definition of the goal kernel, $G$, we can repurpose the saLMDP to optimize a task-policy over the joint space, stitching together policies from the ensemble to complete the task. This formalism is called the task-LMDP, which is used to compute a meta-policy $\mu$ over options (it can also be thought of as a \textit{state-option LMDP}).
\begin{definition}[task-LMDP]
    A task-LMDP is defined by the tuple $\mathscr{L}=(\mathscr{M}^{N}, \mathscr{T}, H_{\psi},H_{\alpha})$, where $\mathscr{M}^{N}$ is a set of first-exit saLMDP control problems of size $N$, the number of states in $\mc X$ to be controlled to, $\mathscr{T}$ is an BOG task, and $H_{\psi}: \mc X \mc A \rightarrow 2^{\Psi}$ is the state-action-to-feature function, $H_{\alpha}: 2^{\Psi} \rightarrow \mc A_{\sigma}$, is the feature-to-goal-action function, used to define the affordance function $F$. $\mathscr{L}$ is used to derive  the objects of an saLMDP, $(\mc S, \mc O, G, p_o, q_o)$, where $\mc S$ is the product space, $\mc O$ is a set of options, $p_o$ is a passive dynamics over options, and $q$ is a cost function, which defines the objective function for the problem. 
\end{definition}

From these three components, we can derive a desirability ensemble $\mc Z_i = \{z_1,z_2,...\}$, policy ensemble $\mc U_{\mc G} = \{u_{1},u_{2},...\}$, option kernel $J$, and affordance function $F$
% , and feasibility function, $\kappa$, 
in order to construct the coupled dynamics $G$ in (\ref{eq:coupleddyn}). $N$ can either be $N_{All}$, for a \textit{complete} ensemble of \textit{all} non-obstacle states, or $N_{g}$, the number of goal states defined by the affordance function. The choice depends on whether or not the solution is intended for task-transfer, where policies in the complete ensemble can be remapped to different problems, as we will soon explain. By defining a uniform passive and controlled dynamics distribution $p_{o}, \mu$, we arrive at the TLMDP objective:

\begin{equation*}
    v(\mathbf{s},o_{\dg})=\min\limits_{\substack{\mu}} \Bigg[q(\mathbf{s},o_{\dg})+ \mc{D}_{KL}(\mu(o'|\mathbf{s}';\mathbf{s},o)||p_{o}(o'|\mathbf{s}';\mathbf{s},o))+ \mathop{\mathbb{E}}\limits_{\substack{o_\dg'\sim \mu \\ \mathbf{s}'\sim G}}[v(\mathbf{s}',o_\dg')]\Bigg],
\end{equation*}
where $\mathbf{s} = (\bs, x,a) \in \mc S$. This is an saLMDP objective function, however, the cost function $q(\mathbf{s},o_{\dg})$ is an additive mixture of functions: $$q(\mathbf{s},o_{\dg})=q_{\sigma \alpha}(\bs,o_{\dg}) + q_{\sigma}(\bs) + q_o(x,a,o_{\dg}).$$ Just like in Section~\ref{section:BOGtask}, $q_{\sigma \alpha}$ encodes the ordering constraints similar to \eqref{eq:BOG_order_cost}, this time enforced on the low-level policy call which will induce the goal variable:
\begin{align*}
    q_{\sigma \alpha}(\bs,o_{\dg}) = \{\infty~~ \mathbf{if}:~\exists c(\sigma_{\dg}^{\bd},\sigma_j^{\bd})\in \mc C_{\Sigma} ~s.t.~ (\bs(\dg)=\sigma_\dg^{\bd})\land (\bs(j)=\sigma_j^{\bd}) ;~\mathbf{else}:~ const.\}.
\end{align*}
Note that the called option, $o_\dg$, will induce the goal $\alpha_\dg$, through $F$ and thus the $\dg$-index on $\sigma_\dg$ is inherited from the index of action from the affordance function $F(x_g,a_g)\rightarrow \alpha_\dg$, where $(x_g,a_g)$ is the terminal state-action of the option call. The function $q_{\sigma}$ is the constant state-cost, and $q_o(x,a,o_{\dg}) = v_{o_\dg}(x,a)$ encodes the cost of following the option $o_\dg$ from $(x,a)$ until inducing $\alpha_\dg$, which is the value function of the policy. This is significant because option kernels combined with a value-function-cost form compact representations which preserve low-level cost information. 
% \newpage

\subsection{Task-LMDP Optimal Desirability Function and Meta Policy}
\begin{figure}[h]
\centerline{\includegraphics[width=\columnwidth]{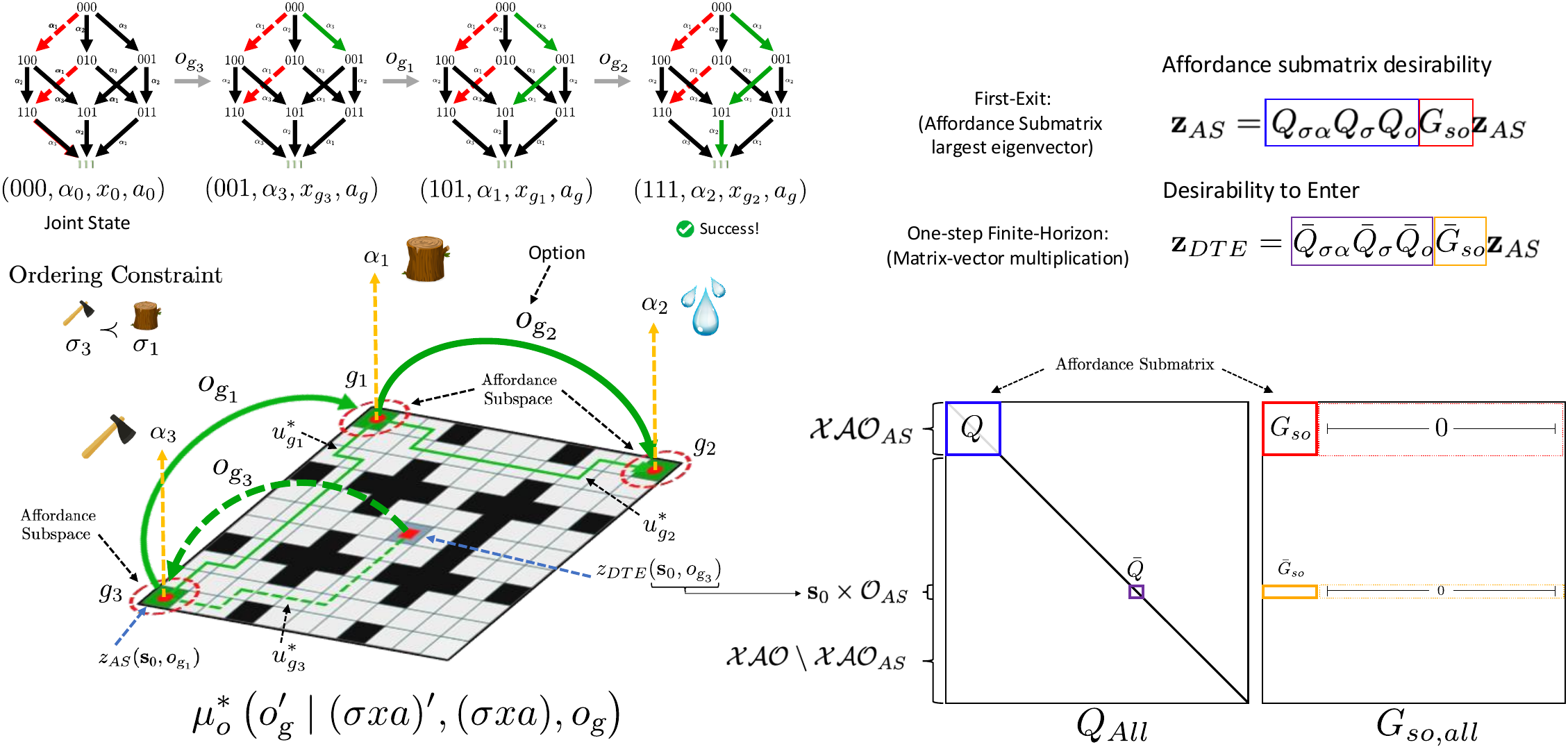}}
\caption{First-exit/Finite-horizon Decomposition (affordance subspace). In our running example, a TLMDP policy $\mu$ generates coupled trajectories for a task. Green arrows on the base state-space depict the jump trajectories whereas green arrows on the top cube are the corresponding dynamics induced in task space. The green line on $\mc X$ shows the agent's trajectory when following the sampled policies until termination. The $\prec$ relation indicates $c(\sigma_i^0,\sigma_j^0)$ and $c(\psi^0_{wood},\psi^0_{axe})$. Dashed green arrows indicated that the policy was sampled using the DTE, which was computed from outside the affordance subspace. $\mc{XA} \mc O_{AS} := \mc{XA}_{\alpha_\dg} \times \mc O_{AS}$ is shorthand for the set of indices $(x,a,o_{\dg})$ which are restricted to the affordance sub-space, where $\mc{XA}_{\alpha_\dg} := \{(x,a) | H_{\psi}(x,a)=\alpha_\dg, ~\forall \alpha \in \mc A_{\sigma}\}$, $\mc O_{AS} := \{o_\dg ~|~ \forall \alpha \in \mc A_{\sigma}\}$, and $\mc{XA}\mc O := \mc X \times \mc A \times \mc O$ is the full product space.  $G_{so,{All}}$ is the passive dynamics matrix over the complete ensemble $\mc U_{All}$, and $Q_{All}$ is a cost matrix on the same space.  By computing the solution in the affordance subspace, we can compute the principal eigenvector solution on the \textit{affordance submatrix} without ever needing to form the full matrices. Because the grounded-subspace (row-space in the complete matrices) has zero entries outside of the grounded state-action-policies, an agent that starts at $(x,a)_g$ and calls option $o_\dg$ will remain within $\mc{XA} \mc O_{AS}$. When the agent starts outside the AS at state $\mathbf{s}_0$, it chooses a policy in $\mc O_{AS}$, which will only communicate with the affordance sub-matrix.} 
\label{fig:task_example}
\end{figure}

Given that the goal kernel $G_{so}$ is only defined on the subset $\mc X\mc A_{\mc G}$ of termination state-actions involved with flipping the bits of the task space, we call this the affordance subspace (AS), we compute the optimal desirability function $z_{AS}$ and meta-policy $\mu$ for the coupled-state-space as using the saLMDP solution (derived in Section \ref{app:TLMDP_Derivation}):
\begin{align}
    \lambda_1\mathbf{z}_{AS}&= Q_{\sigma \alpha}Q_{\sigma} Q_o G_{so}\mathbf{z}_{AS} \label{eq:TLMDP},\\
    \mu^*(o_\dg'|\mathbf{s}',\mathbf{s},o_{\dg}) &= \frac{p_{o}(o_\dg'|\mathbf{s}';\mathbf{s},o_{\dg})z_{AS}(\mathbf{s}',o_\dg')}{\mathfrak{N}[z^*](\mathbf{s}',\mathbf{s},o_{\dg})},
\end{align}
the $Q$-matrices are diagonal matrices of each negatively exponentiated cost function, and $G$ is the matrix for the joint \textit{state-option passive dynamics} of $G_{so}(\mathbf{s}',o'|\mathbf{s},o):=G(\mathbf{s}'|\mathbf{s},o)p_{o}(o'|\mathbf{s}',\mathbf{s},o)$, analogous to $P_{xa}$ in \eqref{eq:salmdp_eig}. In (\ref{eq:TLMDP}), all matrices have the indexing scheme $(\bs,\alpha,x_g,a_g,o_{\dg})$. However, recall that the affordance function $F$ is potentially non-injective between state-actions and high-level actions, and so the number of policies $N_o$ is the size of the preimage of $F$ ($N_o = N_\dg$ when $F$ is bijective). Thus, there are $N_o$ $(\alpha,x,a)$ tuples, and there are $N_o$ options $o_\dg$, which advance $\bs$ by inducing the high-level action from these state-action pairs in the affordance subspace. Hence, the transition matrix $G$ can be represented as a sparse matrix of size $|\Sigma| N_{o}^2 \times |\Sigma| N_{o}^2$. We can compute the first-exit problem restricted to this sub-space, then compute the desirability for the agent's best initial state-option pair $(\mathbf{s},o_{\dg})_0$ which lies outside the AS (for all possible $o_\dg$ from the state). This is the desirability to enter affordance subspace (and then complete the task) at a given grounded state-action under an option from the initial $(x,a,o_{\dg})_0$. Henceforth, we call this the \textit{Desirability-To-Enter} (DTE):
\begin{align}
    \mathbf{z}_{DTE} = \bar{Q}_{\bs \dg}\bar{Q}_{\bs} \bar{Q}_o \bar{G}_{so} \mathbf{z}_{AS},
\end{align}
which is a single matrix-vector multiplication where $\mathbf{z}_{AS}$ is the result of (\ref{eq:TLMDP}) computed for the affordance subspace (see Fig. \ref{fig:task_example} for visualization). $\bar{G}_{so}$ is wide matrix of size $N_o \times |\Sigma| N_{o}^2$ where rows correspond to the agent's current state $\mathbf{s}$ paired with members of the set of AS-policies $\mc O_{AS} = \{o_{\dg_1}, ..., o_{\dg_N}\}$, and $\bar{Q}$-matrices are $N_o \times N_o$ diagonal cost matrices specific to the starting state. By computing $\mathbf{z}_{AS}$ separately from $\mathbf{z}_{DTE}$ we effectively break the problem into two parts, a one-step finite horizon problem for interior states (where the ``step" is a single policy call, not one discrete time-step) and a first-exit problem for the AS. This enables the agent to be anywhere in the state-space while requiring only one additional linear transformation to act optimally.

\begin{algorithm2e}[h]
\DontPrintSemicolon
\SetKw{return}{return}
\SetKwRepeat{Do}{do}{while}
%\SetKwFunction{assume}{assume}
%\SetKwFunction{isf}{isFeasible}
\SetKwData{conflict}{conflict}
\SetKwData{safe}{safe}
\SetKwData{sat}{sat}
\SetKwData{unsafe}{unsafe}
\SetKwData{unknown}{unknown}
\SetKwData{true}{true}
\SetKwInOut{Input}{input}
\SetKwInOut{Output}{output}
\SetKwFor{Loop}{Loop}{}{}
\SetKw{KwNot}{not}
\begin{small}
	\Input{saLMDP ensemble $\mathscr{M}^N$, BOG task $\mathscr{T}=  (\Sigma, \mc A_{\sigma}, \mc C_{\Psi}, \Psi, H_{\alpha},H_{\psi},B)$, affordance function $F$}
	\Output{Task Policy: $\mu^*$}
	$nxa: $ number of state-action pairs\;
% 	$ng: $ number of goals\;
	$npi: $ number of policies\;
	$J \leftarrow zeros(nxa, nxa, npi)$\;
	$\Sigma_{\mc T} \leftarrow \mathtt{set\_terminal\_states\_as\_{DNF}\_clauses}(B)$\;
	$\mc C_{\Sigma}\leftarrow \{c(\sigma_i^{\bd},\sigma_j^{\bd})~|~c(\psi_k^{\boldsymbol{\cdot}},\psi_\ell^{\bd})\in \mc C_\Psi,~\exists (\psi_k,\psi_\ell)\in H^{-1}_{\Psi\alpha}(\sigma_i) \times H^{-1}_{\Psi\alpha}(\sigma_j)\}$\;
	
	Construct $P_{\boldsymbol{\sigma}}$: binary vector transition kernel on $\Sigma$ \hfill\# Section~\ref{section:BOGtask}\;
    \For{$\mathscr{M}_g$ \textbf{in} $\mathscr{M}^N$}
	{   
	    $u_{x_g,a_g}, z_\dg \leftarrow \mathtt{saLMDP\_solve}(\mathscr{M}_g)$\;
	    
	    store $(u_{x_g,a_g}$, $z_\dg)$ in $(\mc U$, $\mc Z)$\;
	}
    	
    \For{$U_{\dg}$ \textbf{in} $\mc U$}
	{   
            \# Terminal and non-terminal state index sets, $\mc T, \mc N$\;
	    $X_{\dg}(\mc N, \mc N) \leftarrow (I-U_{\mc N \mc N})^{-1}U_{\mc N \mc T}$\;
	    $J(\mc N,\mc N,o_\dg) \leftarrow X_\dg(\mc N, \mc N)$\;
            $J(\mc T,\mc T,o_\dg) \leftarrow 1$
	}
    Define $p_o$ as uniform passive dynamics\;
    % $\kappa \leftarrow J \circ F$\g\# \;
    $G \gets P_{\boldsymbol{\sigma}}\circ F \circ J $\hfill\# \eqref{eq:coupleddyn}\;
    $G_{so} \leftarrow G \circ p_o$\hfill\# Section~\ref{section:TLMDP}\;
    $q_{\sigma \alpha}(\bs,o_{\alpha_i}) = \{\infty~~ \mathbf{if}:~\exists c(\sigma_{\dg},\sigma_j)\in \mc C_{\Sigma} ~s.t.~ (\bs(\dg)=\sigma_\dg)\land (\bs(j)=\sigma_j) ;~\mathbf{else}:~ const.\}$\;
    $Q_{\sigma \alpha} \leftarrow \text{diag}(\exp(-\mathbf{q}_{\bs \dg}))$\;
    $q_{\sigma}(\bs) \leftarrow 0 \text{ if } \bs \in \Sigma_{\mc T}, \text{ else } const.$\;
    $Q_{\sigma} \leftarrow \text{diag}(\exp(-\mathbf{q}_{\bs}))$\;
    $Q_o \leftarrow \text{diag}(\text{vec}(\mc Z_{AS}[\mc X \mc A_{\mc G},\mc X \mc A_{\mc G}]))$\hfill\# $\mc X \mc A_{\mc G}$ is the set of all goal state-action indices\;
    $Q \leftarrow Q_{\sigma \alpha}Q_{\sigma}Q_{o}$\;
    Set $\mathbf{z}$ to zero vector with ones on $\bs_{\mc T}$ terminal states.\;
    Define $\mathbf{z}_{\mc T} = \exp(-q(x_\mc T))$\;
    \#Z-Iteration\;
    \While{$delta > \epsilon$}
	{
	    $\mathbf{z}_{\mc N,\text{new}} \leftarrow QG_{so,\mc N\mc N}\mathbf{z}_{\mc N} + QG_{so, \mc N\mc T}\mathbf{z}_{\mc T}$\;
	    $delta \leftarrow sum(abs(\mathbf{z}_{\mc N,\text{new}} - \mathbf{z}_{\mc N}))$\;
	    $\mathbf{z}_{\mc N} \leftarrow \mathbf{z}_{\mc N,\text{new}}$\;
	}
	$\mu^*_{\dg} \leftarrow \frac{p(o_\dg'|\mathbf{s}',\mathbf{s},o_{\dg})z(\mathbf{s}',o_\dg')}{\mathfrak{N}[z^*](\mathbf{s}',\mathbf{s},o_{\dg})} $

\end{small}
\caption{TLMDP$\_$solve}
\label{algor2}
\end{algorithm2e}
% \newpage

\tom{For an overview of all components of the TLMDP, see fig \ref{fig:function_graph}. We also provide the pseudocode for the TLMDP in Alg.\ref{algor2}. In lines 8-12 of the algorithm we compute a set of individual goal-conditioned saLMDP solutions with their desirability functions, and then in lines 12-15 we compute the time-to-go vectors for each policy to construct the option kernel $J$. The rest of the algorithm constructs the goal kernel $G$ and the cost function matrices, and then solves the full problem in lines 29-33 with z-iteration.}

Note that the ordering constraints in $Q_{\sigma \alpha}$ are acting on the row space of $G$. Also, since the cost function $q_o$ equals the value function, i.e. $q_o(o_\dg,x,a)=v_{o_\dg}(x,a)$, the cost matrix $Q_o$ has diagonal entries which are the desirability functions of each sub-problem concatenated into a vector: $Q_o = \text{diag}(\exp(-\mathbf{q}_o)) = \text{diag}(\exp(-\mathbf{v}_o)) = \text{diag}(\mathbf{z}_o)$, where $\mathbf{z}_o$ is a vector with each entry $(x_i,a,g)\rightarrow z_\dg(x_i,a)$ 
% (see app.5.2 for architecture visualization).
(see Section~\ref{sec:function_graph} for architecture visualization.).
Once $\mu^*$ has been computed, the agent follows the option indexed by each sampled $o_\dg$, until the goal state-action has been reached, followed by a newly sampled $o_\dg$. Like the saLMDP \eqref{eq:saLMDP_likihood}, the TLMDP has the same likelihood function for a state-option trajectory (see section \ref{likelihood_2} for details).

Fig. \ref{fig:task_example}.A shows an example of coupled state trajectory and jump dynamics, drawn from the optimal policy which satisfies the task of getting an axe before chopping wood, and collecting water. 
Because the TLMDP solves precedence-constrained multi-goal tasks, it is structurally analogous to (precedence-constrained) Traveling Salesman Problems~\cite{bianco1994exact} given a final goal that grounds to the agent's initial state. Fig. \ref{fig:results_fig}.A shows a problem with nine sub-goals with precedence on sub-goal colors as features, $\psi$. 

\subsubsection{Optimality}

In \ref{app:tMDP} we define a standard objective called the full-task-MDP (where ``full" refers to the usage of a complete fine-grained transition kernel with no jump transitions), which is a direct analogue of the TLMDP that can be solved via Value-Iteration. We motivate our approach to the TLMDP with a proof that, in the case of deterministic transition kernels and policies, value functions based on the full transition kernel are equivalent to value functions based on the option kernel when the option kernel cost-function equals the value function of the low-level policy. This Task-MDP decomposition result can be summarized by the following decomposition theorem:
\begin{theorem}[Task-MDP Equivalency]
    Let $\mathscr{M}=$ $(\Sigma \times \mc X,\mc A,P_{\boldsymbol{\sigma}}\circ F \circ P_x,q,\Sigma_{\mc T})$ be a full-task-MDP (ftMDP) and $\bar{\mathscr{M}}=(\mc M^n, \mathscr{T},H_{\psi},H_{\alpha})$ be a deterministic Task-MDP (where $\mc M = (\mc X,\mc A, P_x, q) \in \mc M^n$ is an MDP with a first-exit control solution to one of $n$ terminal states in $\mc X_g$ with a cost function $q(x)=const.$). When $\Sigma$, $\mc X$, $\mc A$, $P_x$, $P_{\boldsymbol{\sigma}}$, $F$, and $\Sigma_{\mc T}$ are the same for both problems and the cost function used to optimize $\bar{\mathscr{M}}$ is $q(x,o_{\dg})=v_{o_{\dg}}(x),~ \forall (x,o_{\dg}) \in \mc X \times \mc O$, then the optimal value function $\bar{v}$ derived for $\bar{\mathscr{M}}$ and the optimal value function $v$ derived for the Task-MDP are equivalent, $v=\bar{v}$. (see \ref{app:tMDP} for proof).
\end{theorem}

With respect to the TLMDP, given the presence of the KL-cost, it is difficult to know \textit{a priori} what to set the high-level policy cost to be in order for the full problem to be equal to the decomposed semi-MDP solution. However, this decomposition theorem is valuable since the LMDP can approximate any shortest path policy arbitrarily well in the limit of sending the cost function to infinity, as the state-costs dominates the KL-cost in its contribution to the value function \cite{todorov2009efficient}.  

\subsubsection{Boolean Logic Decomposition}
\begin{figure}[h]
\centerline{\includegraphics[width=0.8\columnwidth]{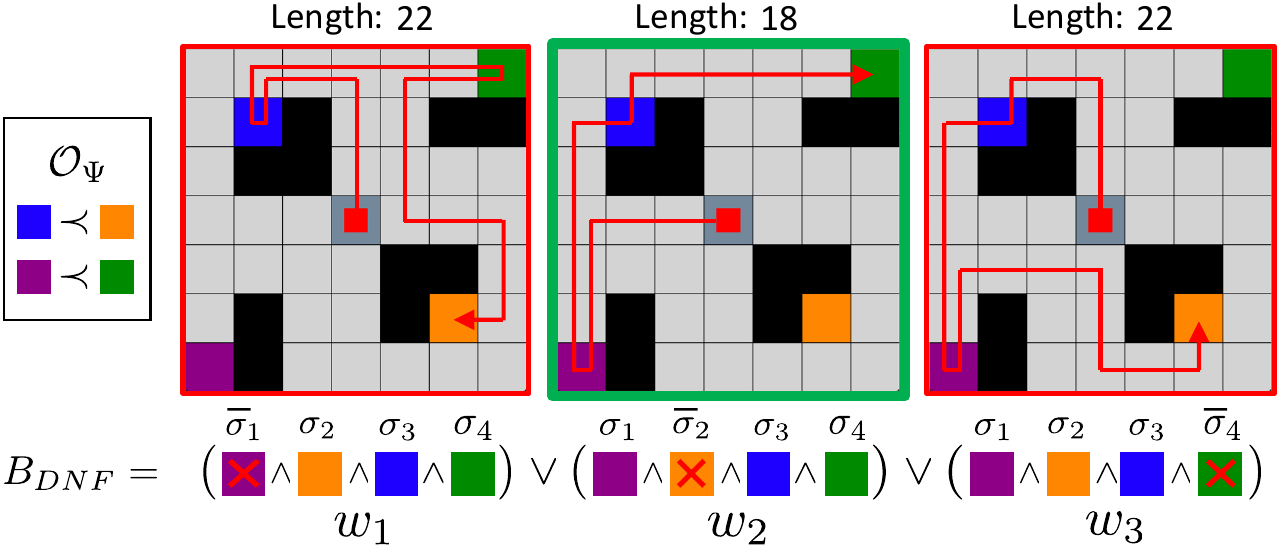}}
\caption{Trajectories for DNF clauses, $w$, of a Boolean formula $B = (\sigma_1\oplus\sigma_2)\sigma_3\sigma_4 \lor \sigma_1\sigma_2\sigma_3\bar{\sigma}_4\implies w_1 \lor w_2 \lor w_3$. Each panel shows a trajectory of a \textit{clause policy} $\mu^*_{w_k}$, computed individually. The green panel shows the optimal clause ($w_2$) for the full task policy, which can be formed by the LMDP policy composition law, which combines the component clause policies into a composite task policy.
% $\oplus$ is XOR, $\alpha_i\alpha_j$ indicates conjunction. 
}
\label{fig:dnf}
\end{figure}

To illustrate the Boolean logic decomposition, Fig. \ref{fig:dnf} shows the task $B = (\sigma_1 \oplus\sigma_2)\sigma_3\sigma_4 \lor \sigma_1\sigma_2\sigma_3\neg\sigma_4$ (where $\oplus$ is XOR, the subscript on $\sigma_i$ indicates the bit position, and the binary superscript indicates  \texttt{True} or \texttt{False}) expanded into DNF: $w_1 \lor w_2 \lor w_3$. Each panel shows an optimal trajectory for a TLMDP \textit{clause policy} $\mu^*_{w_i}$ computed for each clause $w_i$ separately. However, when the task policy is computed for the full formula, OR-ing all disjunctions, the task policy will pursue clause $w_1$ from the initial state because it is cost-minimizing. The remainder of this section will explain how compositions of clauses can be mixed in order to obtain a composite policy.

One can use clause policies $\mu_{w}$ as primitives for the composition law~\cite{todorov2009compositionality} for LMDPs superposition.  This law states that if a set of component problems, $\{\mc M_1,\mc M_2,...\}$, only differ in terminal-state costs $f_k(x_{\mc T})$ for terminal states $x_{\mc T} \in \mc T$, then when they are combined to form a composite final cost $f(x_{\mc T})$, it results in a linear combination of the terminal state desirability \eqref{eq:boundary-mixture}. It follows from the linearity of the solution \eqref{eq:LMDP_linear}, that there exists a linear state-dependent policy mixture \eqref{eq:LMDP-pol-mixture} on the non-terminal states to produce an optimal composite policy. This mixture law for the original LMDP is expressed as:
% (See app.5.9 for discussion).:
\begin{gather}
    f(x_{\mc T})=-\log \left(\sum_{k=1}^{K} c_{k} \exp \left(-f_{k}(x_{\mc T})\right)\right) \implies z(x_{\mc T}) = \sum_{k=1}^K c_k z(x_{\mc T}), \label{eq:boundary-mixture}\\
    \implies u^{*}\left(x^{\prime}|x\right)=\sum_{k} \frac{c_{k} z_{k}(x)}{\sum_{n} c_{n} 
    z_{n}(x)} u_k^*(x'|x)=\sum_{k} m_k(x) u_k^*(x'|x),\label{eq:LMDP-pol-mixture}
\end{gather}
where $c_k$ is a mixture weight for the component desirability $z_k$ and policy $u_k$, $m_k(x)=\frac{c_{k} z_{k}(x)}{\sum_{n} c_{n}z_{n}(x)}$ is a state-dependent mixture weight, and $u$ is the composite policy.  This composition law requires that all of the non-terminal state costs and passive dynamics are shared across the component problems and only vary in the terminal costs. Due to the linearity of the solution, the saLMDP and TLMDP inherit this superposition principal. If we have a TLMDP policy for each clause-option $\mc O = \{o_{w_1},...,o_{w_K}\}$, we have meta policies $\{\mu_{w_1},...,\mu_{w_K}\}$ and desirability functions $\{z_{w_1},...,z_{w_K}\}$. Therefore, we can write a composition law as, 
\begin{align*}
    \mu^*(o'|\mathbf{s}';\mathbf{s},o)=\sum_{k} m_k(\mathbf{s},o) \mu_{w_k}^*(o'|\mathbf{s}';\mathbf{s},o),
\end{align*}
% $$\mu^*(o'|\mathbf{s}';\mathbf{s},o)=\sum_{k} m_k(\mathbf{s},o) \mu_{w_k}^*(o'|\mathbf{s}';\mathbf{s},o),$$ 
where $k$ indexes the component clause policies (See app.\ref{DNF_compo} for full details and discussion). For the TLMDP, because the passive dynamics and costs are shared, the component problems must encode the same precedence constraints and policy-costs. The superposition principal is naturally an OR operation on the potential terminal state costs and is thus suitable for logical statements in disjunctive normal form. By equally weighting each clause $w_k$ with $c_{k}$, we can compute a full task policy by using this policy mixture, as shown in app.\ref{DNF_compo}. One could compute a basis of $2^{|\mc A_{\sigma}|}$ clause policies, one for each possible conjunctive clause, which can then be linearly combined into a task policy for any given formula $B$. Though such a basis is large, the upshot is that, given any logical statement in disjunctive normal form, we can compute a composite policy with a simple mixture law.  The agent need not even compute the full policy mixture for all states in the state-space, because at run-time it only needs to compute the policy for any given state it currently occupies using the mixture function $m_k$. For control, it is never necessary to compute all of the clause policies separately and then combine them (all the clause terminal states can be grouped into one problem, as specified in the TLMDP problem formulation), however, computing clause policies independently could be beneficial in inverse planning, where we may want to infer which clause in a set of possible clauses is being pursued for the purpose of predicting the final low-level state.  For this reason, we included the trajectory inference formula in \eqref{eq:saLMDP_likihood} and appendix \ref{appx:saLMDP_likihood}.

\begin{figure}[h]\centering
\includegraphics[width=0.7\columnwidth]{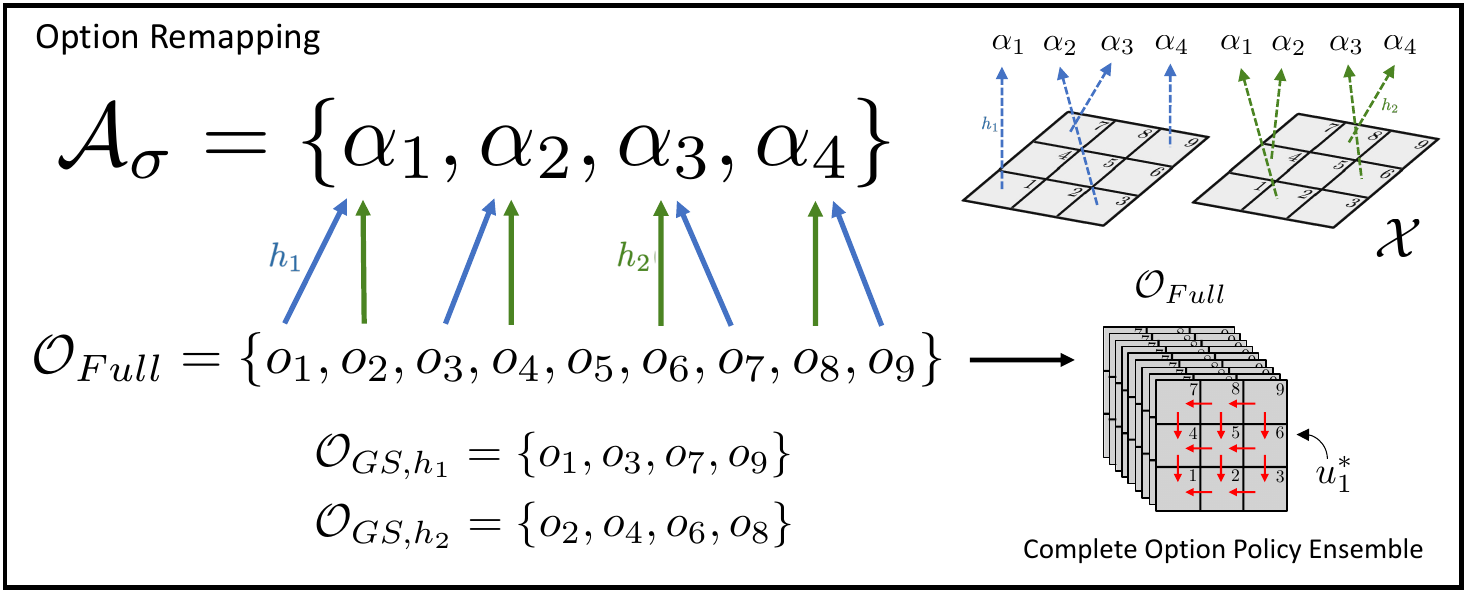}
\caption{Option Remapping: A simple $3\times3$ state-space with a complete option ensemble and two different affordance functions that map options to $\alpha_\dg$ variables. This remapping allows agents to solve a variety of tasks by reusing options.}
\label{fig:remap}
\end{figure}

\subsubsection{Key Properties of the TLMDP solution}

The TLMDP solution (\ref{eq:TLMDP}) has three key properties that make it particularly flexible in the context of efficient task transfer:

\begin{enumerate}
    \item[$P1$] \textit{Policy Remappability}: \tom{As we show in Fig. \ref{fig:remap}}, given a complete policy-desirability ensemble and option kernel, under a new affordance function we only need to re-index the policies (where $o_{\dg}$ indexes different low-level elements of $\mc U_{All}$ and $\mc Z_{All}$) to derive new matrices for (\ref{eq:TLMDP}). Consequently for $N_{\dg}$ goals there is a space of ${N_x\choose N_\dg}2^{N_\dg^2}$ possible tasks one could formulate\footnote{This holds for $N_\dg$ goals grounded onto $N_x$ possible states along with a set of ordering constraints.}, not requiring recomputation of $\mc U_{All}, \mc Z_{All}$ or $J_{All}$.
    \item[$P2$] \textit{Affordance Invariance}: Since the optimal desirability function is a principal eigenvector, there is an invariance of the solution to scalar multiples of the diagonal matrices, assuming $G$ remains constant under the new grounding.
    \item[$P3$] \textit{Minimal Solution Distance}: The agent can compute the optimal desirability (the desirability-to-enter, DTE) for any given initial state, $\mathbf{s}_0$, outside the AS with one matrix-vector product. Importantly, this also holds true after the zero-shot regrounding of a previous grounded-subspace solution. 
\end{enumerate}

% $(\mathbf{P1})$ \textbf{Policy Remappability:} Given a complete policy-desirability ensemble and option kernel, under a new grounding function we only need to re-index the policies (where $o_{\dg}$ references different low-level elements of $\mc U_{All}$ and $\mc Z_{All}$) to derive new matrices for equation (\ref{eq:TLMDP}). Consequently, for $N_{\dg}$ goals there is a space of ${N_x\choose N_\dg}2^{N_\dg^2}$ possible tasks one could formulate (for $N_\dg$ goals grounded onto $N_x$ possible states along with a set of ordering constraints) which would not require recomputation of $\mc U_{All}, \mc Z_{All}$ or $J_{All}$. $(\mathbf{P2})$. \textbf{Affordance Invariance:} Since the optimal desirability function is a principal eigenvector, there is an invariance of the solution to scalar multiples of the diagonal matrices, assuming $G$ remains constant under the new grounding. $(\mathbf{P3})$ \textbf{Minimal Solution Distance:} The agent can compute the optimal desirability (the desirability-to-enter, DTE) for any given initial state, $s_0$, outside the AS with one matrix-vector product. Importantly, this also holds true after the zero-shot regrounding of a previous grounded-subspace solution. 

These properties all allow for the fast transfer of task solutions within the same environment, or across different environments with distinct obstacle configurations and transition kernels. The following sections will elaborate on the empirical and theoretical implications of these properties.

% $P1:$ \textit{Minimal Solution Distance}: The agent can compute the optimal desirability (the desirability-to-enter, DTE) for any given state outside the AS with one matrix-vector product; $P2$: \textit{Policy Remappability}: Given a full policy/desirability ensemble and option kernel, under a new grounding function we only need to re-index the AS-policies in order to derive new matrices for equation (\ref{eq:TLMDP}); $P3$: \textit{Affordance Invariance}: The optimal desirability function, being a principal eigenvector, implies an invariance of the solution to scalar multiples of the diagonal matrices, given $G$, which encodes the affordance sub-goal connectivity, remains constant under the new grounding.

\begin{figure}[!h]
\centerline{\includegraphics[width=\columnwidth]{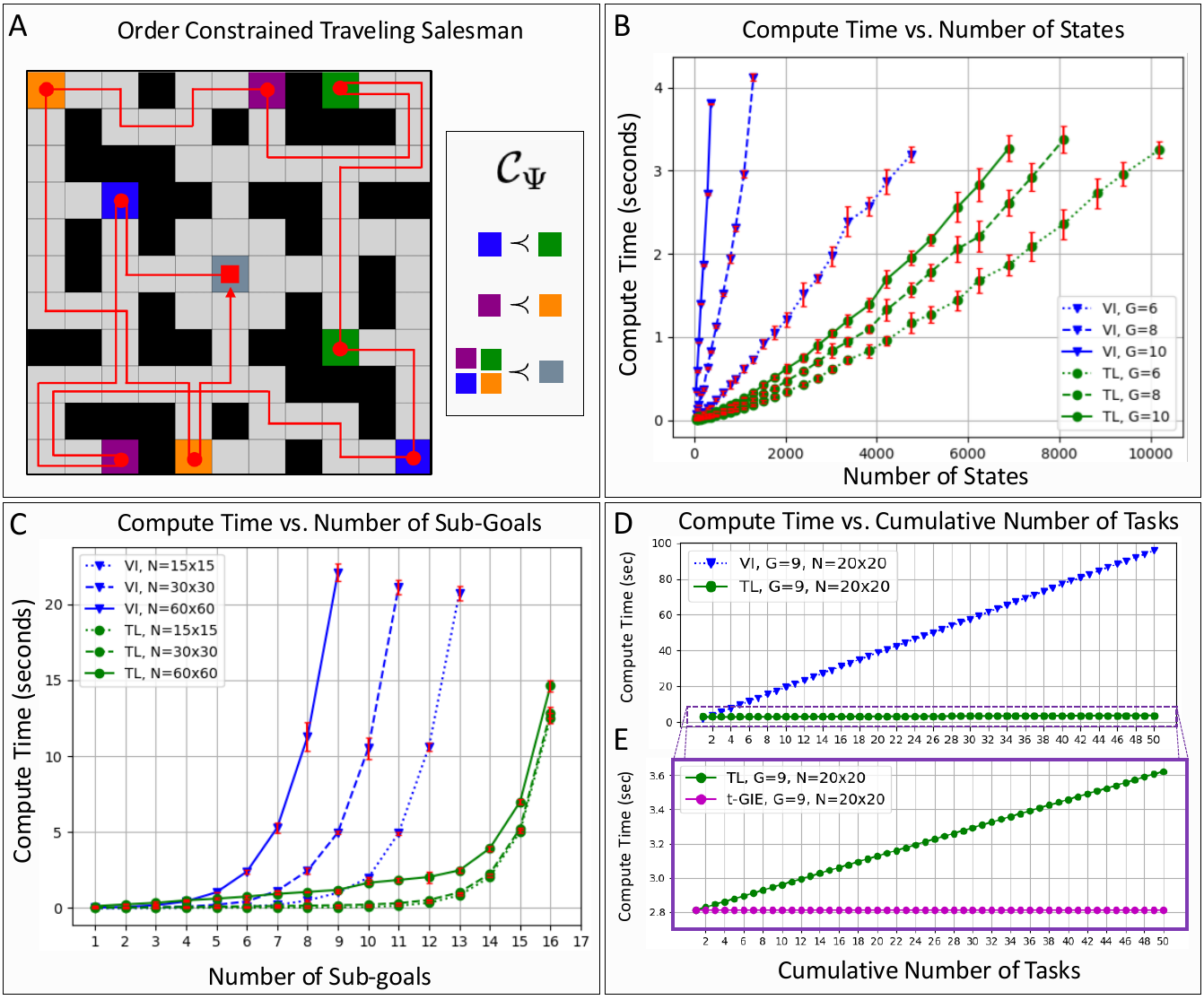}}
\caption{ \textbf{A}: A precedence-constrained Traveling Salesman Problem with constraints defined over sub-goal colors. In \textbf{B},\textbf{C} we demonstrate the \textit{scalability} of the algorithm. \textbf{B}: Computation time when fixing the number of sub-goals and increasing the size of $\mc X$. \textbf{C}: Computation time when fixing the state-space sizes while varying the number of sub-goals. In \textbf{D},\textbf{E} we demonstrate the benefits of policy reuse. \textbf{D}: Cumulative time for solving N re-grounded tasks when fixing the size of $\mc X$ to a $20\times20$ gridworld, and the sub-goal count to nine with two ordering constraints. The TLMDP is efficient by exploiting representational reuse of a full policy ensemble and the option kernel. \textbf{E}: The figure shows a close up of the cumulative time of $\mathscr{L}_{AS}$ compared to the zero-shot $\mathscr{L}_{tGIE}$ solution, requiring no additional computation. In \textbf{B},\textbf{C}, only the AS base-policies were computed, whereas \textbf{D},\textbf{E} uses complete ensembles and option kernel. Sub-goal locations were randomly sampled and in \textbf{B},\textbf{C} we plotted the mean time over 10 episodes. Error bars were plotted with $\alpha$=0.95 (Apple M1 Max, 32 GB RAM, Single Core).}
\label{fig:results_fig}
\end{figure}
% \comment{I think a figure similar to the results presented in Fig. \ref{fig:results_fig} is needed for the office world, right?}

\subsection{Running Example}

For our running example, in Fig. \ref{fig:results_fig}.B-E we empirically show the scaling advantage of computing TLMDP policies in the AS ($\mathscr{L}_{AS}$) relative to computing policies with value-iteration on the full product space, denoted as $\mathscr{V}_{Full}$. \tom{In these examples we aim to show how our algorithm scales with the number of sub-goals (i.e. the size of $\Sigma$), when altering either the size of the state-space $\mc X$, and how it scales when we reuse solutions from previously solved problems.} 

First, in Fig. \ref{fig:results_fig}.B, we hold the number of sub-goals constant at, $N_\dg = 6,8,10$, and vary the number of states in the grid-world. We can see the clear advantage $\mathscr{L}_{AS}$ has from separating the low-level AS-ensemble and option kernel computation from the task-policy optimization. The plots show that by working in the lower-dimensional AS, $\mathscr{L}_{AS}$ can optimize over much larger base state-spaces with a large number of sub-goals. $\mathscr{V}_{Full}$, however, optimizes the task policy over the full product space, hindering scalability. 

In Fig. \ref{fig:results_fig}.C, we fix the number of grid-world states at $15\times15$, $30\times30$, and $60\times60$, and only vary the number of sub-goals: we expect an exponential increase in run-time for both $\mathscr{L}_{AS}$ and $\mathscr{V}_{Full}$, given that the complexity is dominated by $|\Sigma| = 2^{N_\dg}$. However, since $\mathscr{L}_{AS}$ is working in the AS it only needs to compute $N_\dg$ base-policies and $J_{AS}$, whereas $\mathscr{V}_{Full}$ is again burdened by the extra dimensionality of the non-AS non-terminal states. $\mathscr{L}_{AS}$ plots for all grid-world sizes remain relatively similar, due to the knowledge that the AS dimensionality is not a function of non-terminal state count, whereas $\mathscr{V}_{Full}$ sees a progressively larger leftward shift of the curve for each state-space size.

We demonstrate the flexibility granted by $\mathbf{P1}$ in Fig. \ref{fig:results_fig}.D, which plots the total time for computing $N_{\mathscr{L}}$ different tasks (with $N_\dg = 9$ sub-goals and $|\mc C_{\dg}| = 2$ constraints) grounded in the same $20\times20$ environment. $\mathscr{V}_{Full}$ does not have a policy ensemble at its disposal, so it must solve new tasks from scratch, resulting in a worst-case constant-time increase per task proportional to $(2^{N_\dg}|\mc X|)^3|\mc A|$~\cite{puterman2014markov}. Alternatively, $\mathscr{L}_{AS}$ can remap the goal variables to reusable ensemble policies to form a new AS, and therefore it does not require the computation of new $\mc U_{All}$, $\mc Z_{All}$, or $J_{All}$ (which contribute to a one-time upfront cost, resulting in a higher intercept) it only requires recomputation of (\ref{eq:TLMDP}). This results in a constant time-per-task increase proportional to $2^{N_\dg} N_\dg^4$ (see \ref{Complexity}).
For a modest number of sub-goals this solution can be recomputed very quickly. The green line, $\mathscr{L}_{AS}$, shows a negligible growth in time, where each solution to the regrounded nine-subgoal task was computed in approximately $1.5\times10^{-2} s$. The speed with which this remapping and task reoptimization is possible could have important applications for real-time systems which can learn and rapidly adapt to changes in the structure of the task.

\subsection{An Additional Example: Office World}

\begin{figure}[h]
\centerline{\includegraphics[width=\columnwidth]{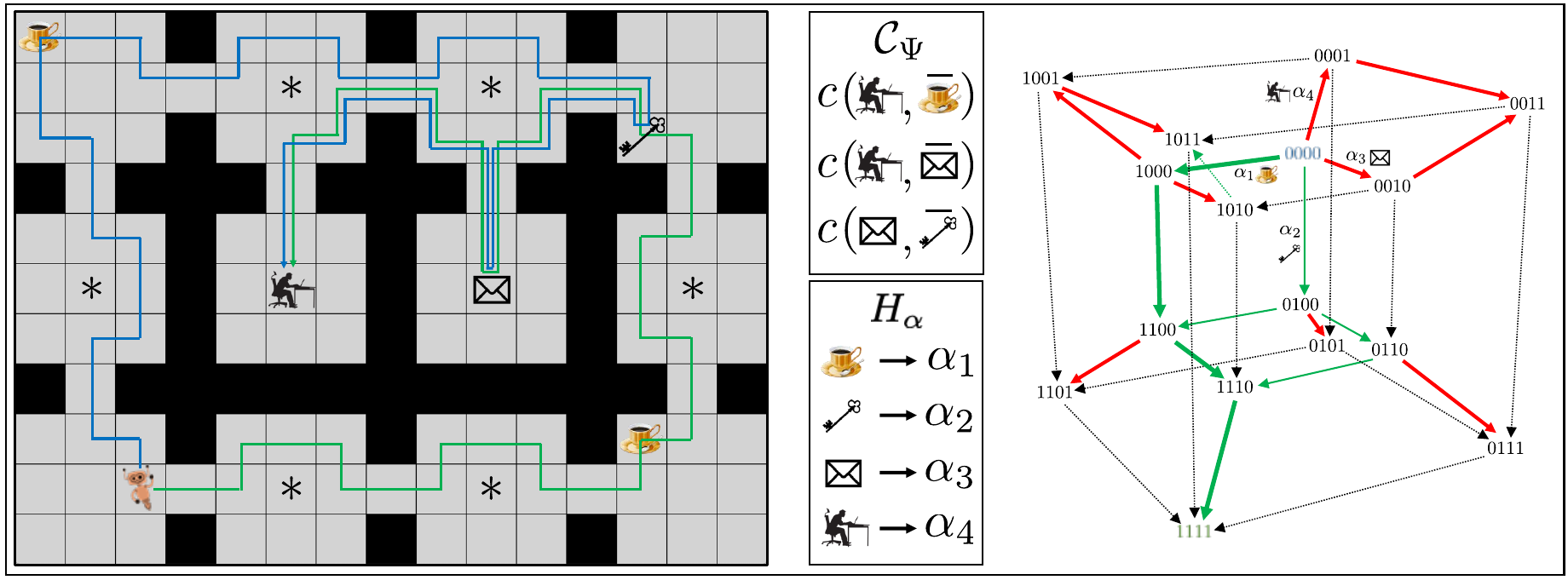}}
\caption{Office world: A robot navigates to satisfy the task of bringing coffee and mail to the central office.  The mail requires obtaining a key first. The green trajectory is the optimal trajectory through the low-level space, where as the blue trajectory is an example of a sub-optimal trajectory. This example demonstrates that multiple objects (coffee in this case) can map to the same bit-flip. The task space shows the constraints $\mc C_\Psi$ (Red arrows) along with the optimal policy dynamics (Green arrows).}
\label{fig:office_world}
\end{figure}

We also show an additional example to demonstrate scenarios where multiple features map to the same variable. Fig. \ref{fig:office_world} shows the office world where a robot must navigate an office to perform a task while avoiding obstacles, indicated by $*$. The task is to bring a coffee and the mail to the desk worker in the center office. However, a key is also needed in order to unlock the mail box.  Therefore the task consists of four goals with the following constraints: The robot cannot go to the desk without the coffee and mail, and also can only obtain the mail when it possesses the key. The red trajectory shows the optimal trajectory for the task.  Notice that there are two states which are mapped to coffee with the function $H_{\psi}(x,a)\rightarrow \psi_{\text{coffee} }$. The function $H_{\alpha}$ only maps single features (no Id) to goal variables, meaning both coffee state-actions map to $\alpha_1$. Thus, the top-left coffee isn't necessary to complete the task and is part of a plan with a sub-optimal cost solution, shown by the blue trajectory.  There are more than two acceptable paths, however the others involve significant backtracking through the state-space. The 4D task cube shows the constraints on bit-flips with red arrows indicating that the transition is unacceptable and won't occur under the optimal policy $\mu^*$. The thin green arrows are acceptable but low probability transitions, and the solid green arrows the high probability transitions through the task-space under $\mu^*$ (corresponding to the red trajectory in the gridworld), whereas the black dotted arrows are acceptable transitions that are inaccessible from the starting task state $\bs_s = 0000$, and therefore occur with zero probability when starting from this state.

\subsection{Compositionality and Transfer}\label{sec:compandtransfer}
A key advantage of computing the solution in the AS pertains to \textit{Affordance Invariance} (property $\mathbf{P2}$), conferred by \textit{Policy Remappability} (property $\mathbf{P1}$). Consider a solution for the TLMDP problem $\mathscr{L}_1$ under affordance function $F_1$. $\mathbf{P1}$ means that under a new affordance function $F_2$ for $\mathscr{L}_2$, the AS-index $o_\dg$ now references a different policy and desirability vector, and also conditions different dynamics in $J_{All}$. Since the matrices $Q_o$ and $G_{so}$ are the only two matrices whose elements are functions of $F$, if these matrices are exactly the same under $F_1$ and $F_2$, then \eqref{eq:TLMDP} is the same, holding the task-matrices $Q_{\sigma \alpha}$ and $Q_{\sigma}$ constant. Furthermore, \eqref{eq:TLMDP} being a principal eigenvector solution implies that if $Q_{o,F_2}$ is a scalar multiple of $Q_{o,F_1}$, it has the same solution, the solution for $\mathscr{L}_1$ will preserve cost-minimization for the new grounding $F_2$, which can change distances between sub-goals. This invariance is called a \underline{t}ask and \underline{c}ost preserving Grounding Invariant Equation (tc-GIE),
\begin{align}
    &\text{tc-GIE:}~ \mathbf{z}_{AS} =\gamma Q_{\sigma \alpha}Q_{\sigma} Q_o G~\!\mathbf{z}_{AS}:\exists \gamma \in \mathbb{R}^{+},Q_{o,F_1}=\gamma Q_{o,F_2},G_{so,1} = G_{so,2} \label{eq:GIE1},\\
    &\text{t-GIE:}~\mathbf{z}_{AS} = Q_{\sigma \alpha}Q_{\sigma} G~\!\mathbf{z}_{AS}:Q_{o,F_1}, Q_{o,F_2} \leftarrow I,~G_{so,1} = G_{so,2}. \label{eq:GIE2}
    % &\text{t-GIE:}~\mathbf{z}_{AS} = Q_{\sigma \alpha}Q_{\sigma} G\mathbf{z}_{AS}: ~\nexists \gamma \in \mathbb{R}_{\setminus{\{0\}}},~ Q_{o,F_1}=\gamma Q_{o,F_2}, ~G_{s_1} = G_{s_2}. \label{eq:GIE2}
\end{align}
% ~ Q_{o,F_1},Q_{o,F_2} = I
If, however, $Q_{o,F_2}$ is not a scalar multiple of $Q_{o,F_1}$, it is still possible for the two problems to share the same solution if $Q_{o}$ is set to the identity matrix for both problems (policy cost equal to zero). This means that, while the task will be solved, the policy will not minimize the low-level trajectory costs on $\mc X$. This is a relaxed version of grounding-invariance called the \underline{t}ask preserving GIE (t-GIE), \eqref{eq:GIE2}. Fig. \ref{fig:results_fig}.E shows the relationship between $\mathscr{L}_{AS}$ (from Fig. \ref{fig:results_fig}.D) and $\mathscr{L}_{tGIE}$, a t-GIE problem with the same task specification. Whereas $\mathscr{L}_{AS}$ sees a slight additional time increase for optimizing \eqref{eq:TLMDP} for every new grounding, $\mathscr{L}_{tGIE}$ solves the task with no further optimization, resulting in zero-shot transfer. Furthermore, regardless of whether two problem formulations \textit{are} related by \eqref{eq:GIE1} or \eqref{eq:GIE2}, the ordering violation costs in $Q_{\sigma \alpha}$ are hard constraints (infinite cost), so as long as $G$ and $Q_{\sigma \alpha}$ remain constant between $\mathscr{L}_1$ and $\mathscr{L}_2$, a policy $u^*_{o,1}$ for $\mathscr{L}_1$ can be used to solve $\mathscr{L}_2$. However, the sub-goal dynamics will retain the biases of $\mathscr{L}_1$'s low-level policy costs, dictated by the obstacle configuration.

% A cliff example is presented in the supplementary material to further illustrate Affordance Invariance (app.5.1).
% % (app \ref{app:comp_transfer}). 
\begin{figure}[h]\centering
\includegraphics[width=\textwidth]{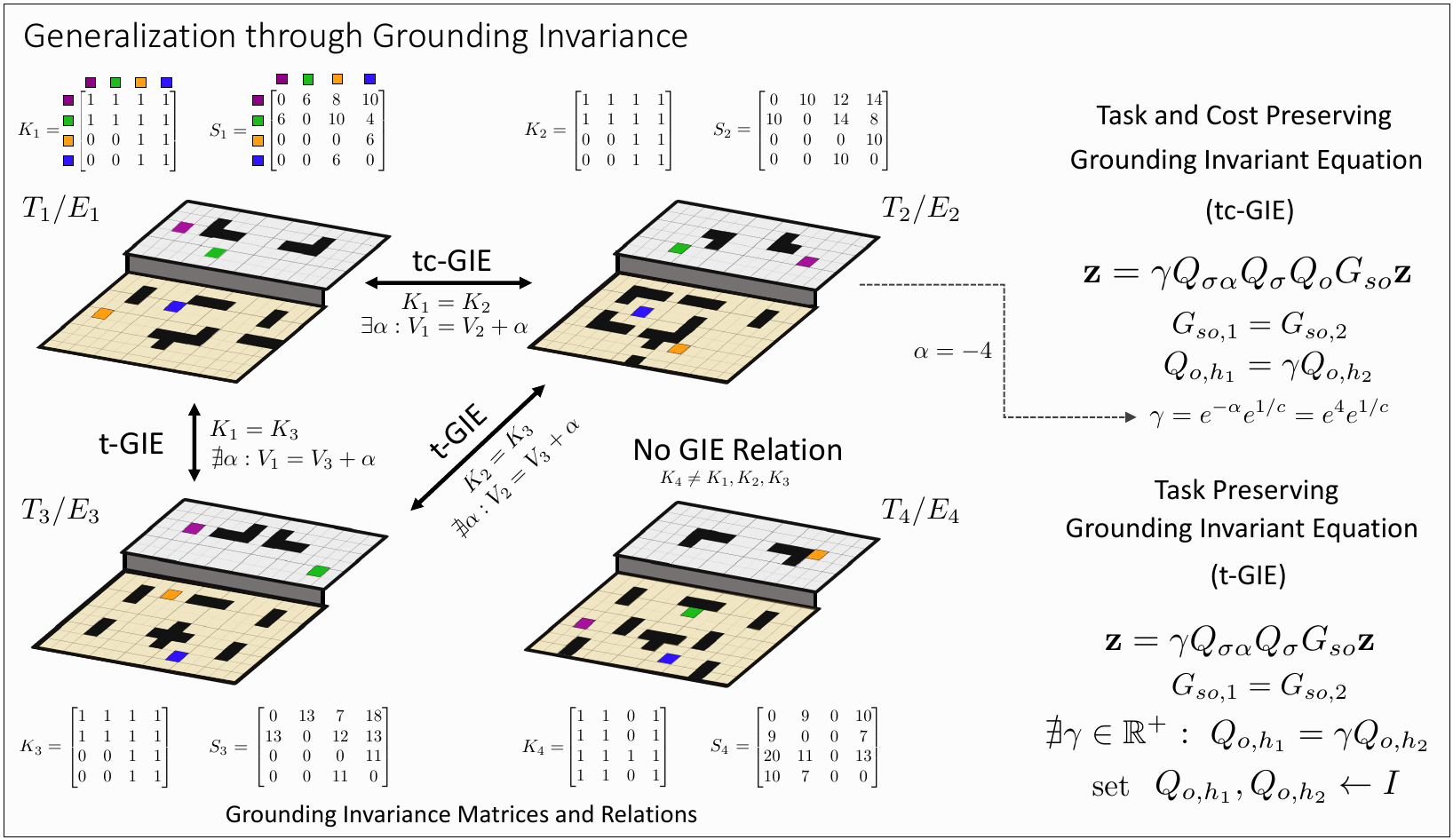}
\caption[width=\columnwidth]{A cliff example, in which the agent can descend but is not able to ascend the cliff. Four different tasks ($T_1$-$T_4$) defined on four cliff environments ($E_1$-$E_4$) that are related by 2 possible GIEs. $S$ is a shortest-path matrix. If $S_1$ is the same as $S_2$ up to an additive constant and they share the same feasibility matrix $K$, Task 2 will have the same solution as Task 1, but mapped to different underlying policies (tc-GIE); this solution will minimize trajectory length on $\mc X$. If no such additive constant exists but the tasks share the same $K$-matrix, the tasks can only share an optimal solution where trajectory length is not minimized (t-GIE) ($Q_o$ is set to $I$). Task 4 does not share a $K$-matrix with other tasks and therefore doesn't share a GIE solution.}
\label{fig:GIE_main}
\end{figure}

Figure~\ref{fig:GIE_main} demonstrates both GIE equations for an environment with a cliff that the agent can descend, but not ascend, resulting in different $G$-matrices depending on the grounding. Since $G$ is a function of $F$, differences can only arise in $G$ under different sub-goal connectivity structures. These connectivity structures are stored in the \textit{feasibility matrix} $K(i,g) = J((x,a)_g|(x,a)_i,o_{\dg})$, which represents the global connectivity of the sub-goals. Furthermore, recall that $Q_{o_{\dg}} = \text{diag}(\mathbf{z}_\dg)$ where $\mathbf{z}_\dg$ is a concatenated state-action vector of all low-level desirability functions in $\mc Z_{AS}$.  Because the desirability function is defined as $z(x) = e^{-v(x)}$, constant multiples of $z$ translate into constant additives of $v(x)$, (i.e. $\gamma z(x) = e^{-(v(x)+\alpha)}$ where $\gamma = e^{-\alpha}$).  Since we are working with shortest path policies, to better illustrate the tc-GIE relationship in Fig. \ref{fig:GIE_main}, we construct a matrix $S$ of shortest path distances which are derived from the value function divided by the non-terminal state cost, which gives us the shortest path distance, i.e. $S(i,g) = \frac{v_g((x,a)_i)}{c}$, where $c$ is the constant non-terminal state-action cost used to define $q(x,a)$, and $v_g = -\log(z_\dg)$ is the value function corresponding to the low-level policy with terminal state $(x,a)_g$.  The new affordance functions preserve the solution under an additive constant $\alpha$ to inter-subgoal distances, which translates into a multiplicative constant $e^{-a}e^{1/c}$ between desirability functions used to define $Q_o$. Therefore, even between environments with different obstacle constraints, we can compute that two tasks in the affordance subspace are exactly the same because they share the same underlying objective function. We emphasize that this will only occur in the case where the sub-goal distances happen to be the same across the environments up to an additive constant, any deviation from this condition may mean that the optimal solution has changed. The t-GIE is also depicted in Fig. \ref{fig:GIE_main}: task $T_3$ in $E_3$ does not share a policy cost matrix with a multiplicative constant with $T_1,T_2$; however, the feasibility structure remains the same and $Q_o$ is set to $I$ for both problems to obtain the same objective function.

\section{Discussion and Conclusions}

We have demonstrated how to solve long-horizon non-Markovian Boolean tasks with a Task LMDPs that are solved with Linearly-solvable Goal Kernel Dynamic programming, a control paradigm which enables task transfer.
% The construction of the policy ensemble and option kernel is computationally decoupled from the global TLMDP task-optimization.  This decoupling is necessary for scalability: both objects can be parallelized and the option kernel is required to work in the grounded-subspace, which is critical for the method to scale to large state-space sizes.
Since option kernels are paired with cost functions derived from the low-level value function, our method performs multi-step tasks planning while integrating all of the low-level cost information for each successive jump, constituting a closed-form solution for long-horizon problems.
% which is a new abstraction technique for hierarchical control.
While the focus of this paper was on exact dynamic programming methods, the motifs have important implications for learning frameworks.

As we have shown, the task can be a structured object which can be traversed by a first-exit policy, where task rules correspond to a cost function shaping permissible dynamics on this space. There are limitations with the task model that could be expanded in future work. For instance, in the office gridworld example, there is no state representation for the state of the coffee or mail being placed on the desk, these detail are abstracted away. Therefore, it is not natural to express plan constraints that would make the robot first bring the coffee to the desk and then fetch the mail (after getting the key) and place it on the desk.  However it would certainly be possible to expand the state representation to accommodate this. A key addition would be to allow the goal variables in the BOG task to induce more complicated bit flips.  For example, the aforementioned act of transferring the coffee from the robot to the desk may involve an action that performs a bit swap $([0,1]\rightarrow [1,0])$ between bits that represent the state of the cup transferring from the hand to the top of the table. Of course, these modeling and representation problems have long been studied in the planning literature often using various kinds of predicate logic \cite{russell2002artificial}; our contribution here has been to show how we can solve some of these more classical logical reasoning problems when they are embedded in Markov Decision Processes.

It is critical to recognize that first-exit value functions are endowed with additional \textit{explicit} meaning: the accumulated cost \textit{until hitting the terminal state}. This means expected path costs or lengths of each option are summed for each policy, which is what allows the problem to be decomposed. Importantly, there is no natural analogue for reward-maximization problems of this kind. A sparse reward problem which only rewards a terminal logical task state does not permit the summation of value functions of each option to equal the total reward of completing the true sparse-reward task optimally. This is because long-horizon sparse reward problems \textit{do not have rewards for sub-goals} which can be used to compute value functions for sub-problems that can be summed, despite the prevalent use of pseudo-rewards in RL. On the other hand, path lengths or costs can be summed for sub-goals because they carry information about the agent's trajectory distributions under the each option, which are stitched together for the global problem. Naturally, if a sub-goal is not reachable the option will never be chosen because it contributes an infinite cost.

Lastly, the explicit event space of the option kernel is necessary for  \textit{remappability}, allowing low-level policies to generalize to new tasks by mapping the terminal state-actions to high-level actions. Indeed, the task in the definition of the TLMDP need not be a BOG task, and future work could scale up these principles to more complex tasks specified by appropriately grounded automata, as well as exploit the TLMDPs favorable properties for task inference.

\newpage
\acks{The authors wish to thank Keno Juchems and Andrew Saxe for their insightful comments and feedback.
}
% \newpage
% \appendix
% \section*{Appendix A. Probability Distributions for N-Queens}

% [section ommitted]

\vskip 0.2in
\bibliography{references}
\bibliographystyle{apacite}

\clearpage
\section{Appendix}

\subsection{Function Dependency Graph}\label{sec:function_graph}

Here we present an illustration of all of the functions that we have discussed in this paper:
\begin{figure}[h]
\centerline{\includegraphics[width=\columnwidth]{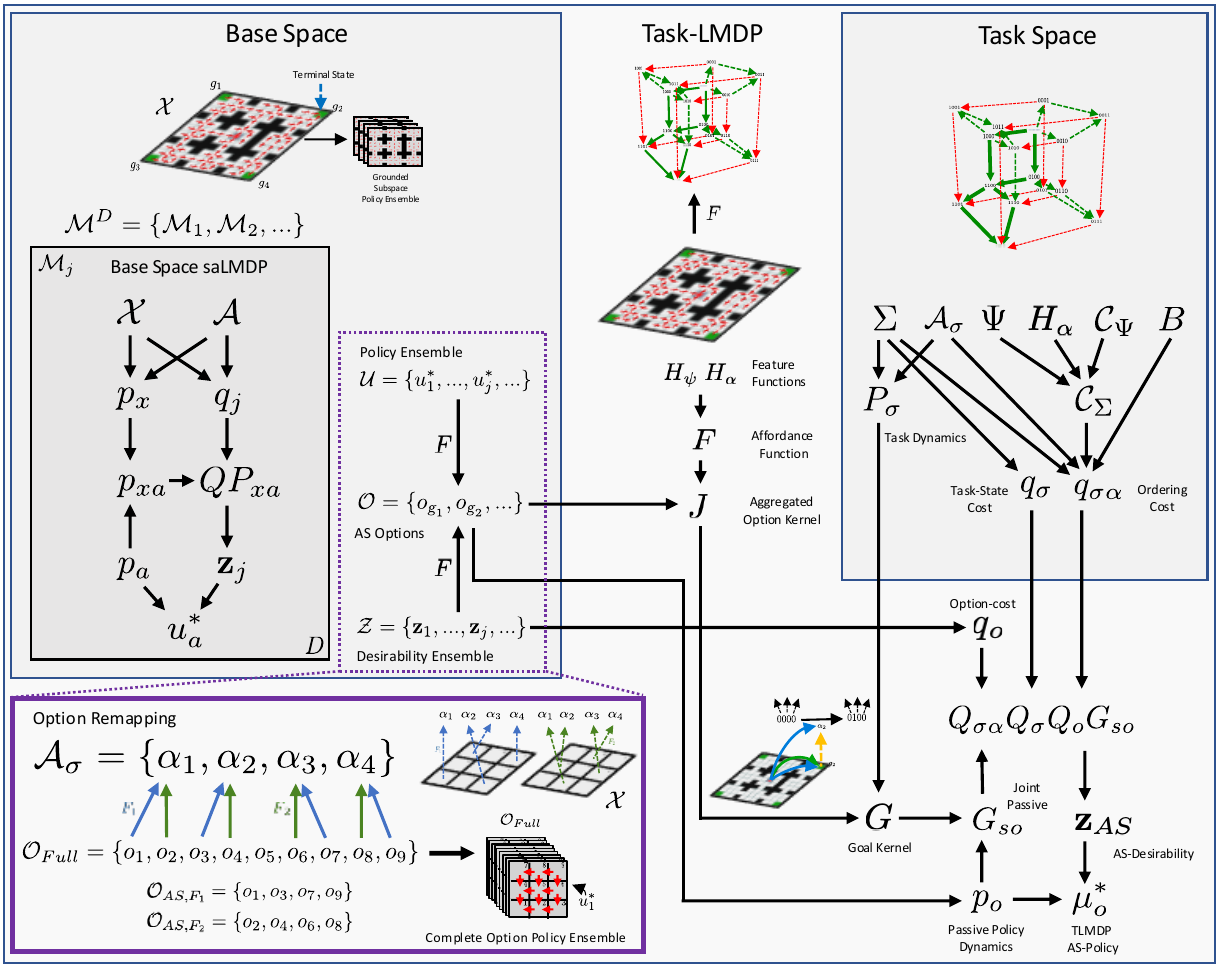}}
\caption{Function Dependency Graph: This graph shows the dependency structure of the different objects in this paper. Arrows denote that either one object was derived from another, or it takes another object as an argument. Not every relationship is plotted, only the essential structure for understanding the architecture.}
\label{fig:function_graph}
\end{figure}
\clearpage

\subsection{saLMDP Derivation}\label{sec:saLMDP_append}

We begin by creating a state vector ${y} = (x,a)$ and substitute this state in the equation for an LMDP:

\begin{align*}
&v^*(x)=\min\limits_{\substack{u}} \Big[q(x) + \mc{D}_{KL}(u(x'|x)||p(x'|x)) + ~\expec\limits_{\mathclap{{x'\sim u}}}~[v^*(x')]\Big],\\
&v^*({y}) = v^*(x,a)=\min_{u_{xa}} \Big[q(x,a) + \mc{D}_{KL}(u_{xa}(x',a'|x,a)||p_{xa}(x',a'|x,a)) \,\,+\,\, \mathop{\mathbb{E}}_{\mathclap{x',a'\sim u_{xa}}}\,\,[v^*(x',a')]\Big].
\end{align*}

We now decompose the joint distributions using the product rule and impose a dynamics constraint. We would like the low level dynamics of the saLMDP, $p_{x}(x'|x,a)$, to behave like an uncontrollable physical system and control to only be achieved over the state-space by manipulating the action distribution, $u_a$. To enforce this condition we introduce the constraint that $u_x = p_x$, and we minimize over $u_a$, obtaining 

\begin{align}
&= \min\limits_{\substack{u_a \\ u_x}} \Bigg[q(x,a) + \mathop{\mathbb{E}}\limits_{\substack{a'\sim u_a \\ x'\sim u_x}}\Bigg[\log\Bigg( \frac{u_{a}(a'|x';x,a)u_{x}(x'|x,a)}{p_{a}(a'|x';x,a)p_{x}(x'|x,a)}\Bigg)\Bigg] + \mathop{\mathbb{E}}\limits_{\substack{a'\sim u_a \\ x'\sim u_x}}[v^*(x',a')]\Bigg] \quad \stackrel{s.t.}{u_x = p_x},
\label{eq:HD-LMDP}\\
&= \min\limits_{\substack{u_a}} \Bigg[q(x,a) + \mathop{\mathbb{E}}\limits_{\substack{a'\sim u_a}}\Bigg[\log\Bigg( \frac{u_{a}(a'|x';x,a)}{p_{a}(a'|x';x,a)}\Bigg)\Bigg] + \mathop{\mathbb{E}}\limits_{\substack{a'\sim u_a \\ x'\sim u_x = p_x}}[v^*(x',a')]\Bigg].
\label{eq:HD-LMDP_reduced}
\end{align}

We consider \eqref{eq:HD-LMDP_reduced} to be the canonical form of the saLMDP objective function.

After transforming the cost-to-go, $v(x,a)$, into desirability via $v(x,a) = -\log(z(x,a))$ the derivation proceeds as follows:

\begin{align}
&=\min\limits_{\substack{u_a}} \Bigg[q(x,a) + \mathop{\mathbb{E}}\limits_{\substack{a'\sim u_a}}\Bigg[\log\Bigg( \frac{u_{a}(a'|x';x,a)}{p_{a}(a'|x';x,a)}\Bigg)\Bigg] + \mathop{\mathbb{E}}\limits_{\substack{a'\sim u_a \\ x'\sim p_x}}[-\log(z^*(x',a'))]\Bigg],
\label{eq:HD-LMDP1}\\
&=\min\limits_{\substack{u_a}} \Bigg[q(x,a) + \mathop{\mathbb{E}}\limits_{x'\sim p_x }\Bigg[\mathop{\mathbb{E}}\limits_{a'\sim u_a}\Bigg[\log\Bigg( \frac{u_{a}(a'|x';x,a)}{p_{a}(a'|x';x,a)z^*(x',a')}\Bigg)\Bigg]\Bigg]\Bigg],
\label{eq:HD-LMDP2}\\
&=\min\limits_{\substack{u_a}} \Bigg[q(x,a) + \mathop{\mathbb{E}}\limits_{x'\sim p_x }\Bigg[\mathop{\mathbb{E}}\limits_{a'\sim u_a}\Bigg[\log\Bigg( \frac{u_{a}(a'|x';x,a)}{p_{a}(a'|x';x,a)z^*(x',a')\frac{\mathfrak{N}[z^*](x',x,a)}{\mathfrak{N}[z^*](x',x,a)}}\Bigg)\Bigg]\Bigg]\Bigg],
\label{eq:HD-LMDP3}\\
&=\min\limits_{\substack{u_a}} \Bigg[q(x,a) + \mathop{\mathbb{E}}\limits_{x'\sim p_x }\Bigg[-\log\big(\mathfrak{N}[z^*](x',x,a)\big)+\mathop{\mathbb{E}}\limits_{a'\sim u_a}\Bigg[\log\Bigg( \frac{u_{a}(a'|x';x,a)}{\frac{p_{a}(a'|x';x,a)z^*(x',a')}{\mathfrak{N}[z^*](x',x,a)}}\Bigg)\Bigg]\Bigg]\Bigg], 
\label{eq:HD-LMDP4}
\end{align}
where $\mathfrak{N}[\cdot](\cdot)$ is a normalization term that satisfies the KL divergence requirement that the terms sum to 1. 
\begin{eqnarray}
\mathfrak{N}[z](x',x,a) = \sum_{a'\in \mc A}p_a(a'|x';x,a)z(x',a').\label{eq:app_salmdp_linear}
\end{eqnarray}
By setting the policy $u_a^{*}$ to the denominator (\ref{eq:HD-LMDP4}), the third term goes to zero: 
\begin{eqnarray}
u_a^{*}(a'|x';x,a) = \frac{p_a(a'|x';x,a)z^*(x',a')}{\mathfrak{N}[z^*](x',x,a)}.
\label{eq:pol_derivation}
\end{eqnarray}
This leaves two terms that do not depend on the minimization argument $u_a$. The resulting equation is:
\begin{eqnarray}
\nonumber-\log(z^*(x,a)) = q(x,a) - \mathop{\mathbb{E}}\limits_{x'\sim p_x }\big[\log\big(\mathfrak{N}[z^*](x',x,a)\big)\big],\\
z^*(x,a) = \exp(-q(x,a))\exp\Big(\mathop{\mathbb{E}}\limits_{x'\sim p_x }\big[\log\big(\mathfrak{N}[z^*](x',x,a)\big)\big]\Big)\label{eq:app_nonlinear_z}.
\end{eqnarray}

The above equation can be written in matrix-vector notation with nested non-linearities, but it requires us to construct non-square matrices. This is because if $\mathbf{z}$ is a state-action desirability vector, then $\mathfrak{N}[z](x',x,a) = \mathop{\mathbb{E}}\limits_{a'\sim p_a} z(x',a')$ requires us to only sum over only the $a'$ variables of the vector. 

% In order to understand the above expression over all state-action pairs, it may be helpful to first view it in the notation of (paired) tensor contractions and then show an equivalent matrix-vector multiplication with nested non-linear functions. In tensor notation, we can write eq. (\ref{eq:finalSingleVar}) as:

% \begin{eqnarray}
% Z = Q \odot \exp(P_{x'}[log(H_{a'}Z'^{a'})]^{x'})
% \label{eq:TensContr}
% \end{eqnarray}

% Where $Z$ is an $X\times A$ state desirability matrix (2-tensor), $P$ is an $X\times A\times X'$ 3-tensor which represents the state dynamics distribution $p_x$, $H$ is an $X\times A\times X'\times A'$ 4-tensor representing the action dynamics distribution $p_a$, and $Q$ is an $X\times A$ matrix of the state-action cost function with entries $Q_{i,j} = \exp(-q(x_i,a_j))$. Here, $\exp$, $\log$, and the Hadamard product $\odot$ are element-wise operations. The notation $H_{a'}Z'^{a'}$ and $P_{x'}B^{x'}$ denotes a tensor contraction, which involves matching the variables of the tensors and performing point-wise multiplication and summation of the $a'$ and $x'$ dimensions, where $B=log(H_{a'}Z'^{a'})$ is a 3-tensor. 

To construct the appropriate matrices, we encode the distributions $p_x$ and $p_a$ into two matrices, $M$ and $W$ respectively. $M$ is an $XA\times XAX'$ matrix (where $X = |\mc X|, A = |\mc A|$) which encodes $p_x(x'|x,a)$, where the additional $XA$ in the column space are dummy variables. That is, the indexing scheme for the rows is $(x,a)$, and $(x,a,x')$ for the columns, so elements with $(x,a)$ in the columns can only have non-zero entries if they match the $(x,a)$ pairs on the row. Similarly, $W$ is an $XAX'\times X'A'$ matrix encoding $p_a(a'|x',x,a)$ where the matrix can only have non-zero entries when the $x'$ on the row matches the $x'$ on the column. It is straightforward to check that multiplying $M$ and $W$ forms a matrix of the joint distribution $P_{xa} = MW$. $W$ allows us to take the expectation over the $a'$ variables in $\mathbf{z}$, followed by the point-wise $\log$ nonlinearity, then $M$ takes the expectation over the $x'$ variables.

Now written in matrix-vector notation we have:
\begin{align}
\mathbf{z} = Q\exp(M\log(W\mathbf{z})).   
\label{eq:nonlinear_z}
\end{align}

Since the $\exp(\cdot)$ and $\log(\cdot)$ are element-wise functions, Jensen's inequality implies the following element-wise inequality:\\
\begin{align*}
    \mathbf{z}_1 = Q\exp(M\log(W\mathbf{z})) \preceq QMW\mathbf{z} = \mathbf{z}_2.
\end{align*}

% \textcolor{red}{It would be great if we could get the stochastic problem to reduce to a linear equation. One of the questions I ask myself is: Given that we know via Jensen's inequality that the RHS is larger, is it possible that we can produce a modified Q matrix such that we can use the linear equation and get the same result as the non-linear? That is: $z =\hat{Q}MHz'$ equals $z = Q\exp(M\log(Hz'))$ for some $\hat{Q}?$. Seems unlikely.}

However, Jenson's inequality holds as equality when the expectation is over a deterministic distribution. Therefore if every row of the state dynamics matrix $M$ is a deterministic distribution, then we can exchange the order of the point-wise exponential with the matrix multiplication:\\
\begin{align}
&\mathbf{z} = Q\exp(M(\log(W\mathbf{z}))),\\
\implies &\mathbf{z} = QM\exp(\log(W\mathbf{z})),\\
\implies &\mathbf{z} = QMW\mathbf{z},\\
\implies &\mathbf{z} = QP_{xa}\mathbf{z}, 
\label{eq:linear2}
\end{align}
where $P_{xa}=MW$ is the matrix for the joint distribution $p(x',a'|x,a)$. The original non-linear mapping reduces to a linear equation:

Thus \eqref{eq:app_nonlinear_z} can be written as the linear equation:
\begin{align}
    z(x,a) = \exp(-q(x,a))\mathop{\mathbb{E}}\limits_{x'\sim p_x }\mathfrak{N}[z^*](x',x,a)),
\label{eq:salmdp_linear_det}
\end{align}
and $z$ can be solved for as the principal eigenvector of $QP_{xa}$. We can solve for $\mathbf{z}$ by the iterating either (\ref{eq:nonlinear_z}), in the stochastic case, or (\ref{eq:linear2}) in the deterministic case, until convergence. Then plug $\mathbf{z}$ into (\ref{eq:pol_derivation}) to obtain the optimal policy.

\subsubsection{TLMDP Derivation}\label{app:TLMDP_Derivation}
% \comment{can be moved before the proof?}

The TLMDP only differs in the transition kernels and number of cost functions. Following the same derivation above we will arrive at:

\begin{align*}
&-\log(z(\mathbf{s},o_{\dg})) = q(\mathbf{s},o_{\dg}) - \mathop{\mathbb{E}}\limits_{\mathbf{s}'\sim G }\big[\log\big(\mathfrak{N}[z^*](\mathbf{s}',\mathbf{s},o_{\dg})\big)\big],
\\
&z(\mathbf{s},o_{\dg}) = \exp(-q_{\sigma \alpha}(\bs,o_{\dg}))\exp(-q_{\sigma}(\bs))\exp(-q_{o}(o_\dg,x,a))\exp\Big(\mathop{\mathbb{E}}\limits_{\mathbf{s}'\sim G }\big[\log\big(\mathfrak{N}[z^*](\mathbf{s}',\mathbf{s},o_{\dg})\big)\big]\Big),
\end{align*}

Following the previous methods for the saLMDP solution derivation, in matrix-vector notation and under deterministic $G$ this results in:

\begin{align*}
\lambda_1\mathbf{z} = Q_{\sigma \alpha}Q_{\sigma}Q_{o}G_{so}\mathbf{z}.
\end{align*}

\subsubsection{TLMDP Finite-Horizon/First-Exit Decomposition (affordance subspace)} \label{section:GroundedSub}

Because we are working with shortest-path first-exit control problems, all trajectories from any given $\bs_k \rightarrow \bs_{k+1}$ must be shortest paths (where $k$ is a lumped time-step called a \textit{period} which corresponds to the window of time $\bs$ remains constant).  This means instead of working with the complete ensembles, $\mc U_{All}, \mc Z_{All}$, and $J_{All}$ we only need to work with $\mc U_{AS}, \mc Z_{AS}, J_{AS}$.  This is because any policy $u_i^*$ which controls to an non-terminal state-action $(x,a)_i$ followed by a policy that controls to $(x,a)_g$ by $u_g^*$ can never have an accumulated cost-to-go $v_i + v_g$ which is lower than the value function $v_g$ for $u^*_g$. (because shortest-path policies have the lowest cost-to-go).  The consequence of the AS restriction is that an agent that controls from a state inside the AS will remain inside the AS (given that the $J_{AS}$ can only jump the agent to AS state-actions). Therefore, a first-exit TLMDP problem can be decomposed by separating the first period out of the Bellman equation and solving a first-exit (FE) problem starting at $k=1$, while solving the policy for the period $k=0$ as a one-step finite-horizon (FH) problem which controls into the affordance subspace.

We define the Bellman equation in two parts, one for controlling within the affordance subspace \eqref{eq:AS_Bellman}, and one for controlling from outside the affordance subspace into the affordance subspace \eqref{eq:out_to_in}:

% \begin{eqnarray}
% v_{t_0}^*(\mathbf{s},o_{\dg})= \min\limits_{\substack{u_{o}}} \Bigg[q(\mathbf{s},o_{\dg})
% + \mathop{\mathbb{E}}\limits_{\substack{o_\dg\sim \mu}}\Bigg[\log\Bigg( \frac{\mu(o_\dg'|\mathbf{s}',\mathbf{s},o_{\dg})}{p_{o}(o_\dg'|\mathbf{s}'
% ,\mathbf{s},o_{\dg})}\Bigg)\Bigg]+ \mathop{\mathbb{E}}\limits_{\substack{o_\dg'\sim \mu \\ \mathbf{s}'\sim G}}[v_{FE}(\mathbf{s}'
% ,o_\dg')]\Bigg]\\
% v_{t_0}^*(\mathbf{s},o_{\dg})=\min_{u_{so_\dg}} \Big[q(\mathbf{s},o_{\dg}) + \mc{D}_{KL}(u_{so_{\dg}}(\mathbf{s}',o_{\dg}'|\mathbf{s}_{t_0},o_{\dg}_{t_0})||p_{so_{\dg}}(\mathbf{s}',o_{\dg}'|\mathbf{s}_{t_0},o_{\dg}_{t_0})) \,\,+\,\, \mathop{\mathbb{E}}_{\mathclap{\mathbf{s}',o_\dg'\sim u_{so_{\dg}}}}\,\,[v_{FE}^*(\mathbf{s}',o_{\dg}')]\Big]
% \end{eqnarray}
% where:
% \begin{eqnarray}
% v_{FE}^*(\mathbf{s},o_{\dg})=\min_{u_{so_{\dg}}} \Big[q(\mathbf{s},o_{\dg}) + \mc{D}_{KL}(u_{so_\dg}(\mathbf{s}',o_{\dg}'|\mathbf{s}_{t_0},o_{\dg}_{t_0})||p_{so_{\dg}}(\mathbf{s}',o_{\dg}'|\mathbf{s}_{t_0},o_{\dg}_{t_0})) \,\,+\,\, \mathop{\mathbb{E}}_{\mathclap{\mathbf{s}',o_\dg'\sim u_{so_{\dg}}}}\,\,[v_{FE}^*(\mathbf{s}',o_{\dg}')]\Big]
% \end{eqnarray}

\begin{align}
v_{k_0}^*(\mathbf{s}_0,o_{\dg})= \min\limits_{\substack{\mu}} \Bigg[q(\mathbf{s}_0,o_{\dg})
+ \mathop{\mathbb{E}}\limits_{\substack{o_\dg\sim \mu}}\Bigg[\log\Bigg( \frac{\mu(o_\dg'|\mathbf{s}',\mathbf{s}_0,o_{\dg})}{p_{o}(o_\dg'|\mathbf{s}'
,\mathbf{s}_0,o_{\dg})}\Bigg)\Bigg]+ \mathop{\mathbb{E}}\limits_{\substack{o_\dg'\sim \mu \\ \mathbf{s}'\sim G}}[v_{FE}(\mathbf{s}'
,o_\dg')]\Bigg],\label{eq:out_to_in}
\end{align}
where:
\begin{align}
v_{FE}^*(\mathbf{s},o_{\dg})= \min\limits_{\substack{\mu}} \Bigg[q(\mathbf{s},o_{\dg})
+ \mathop{\mathbb{E}}\limits_{\substack{o_\dg\sim \mu}}\Bigg[\log\Bigg( \frac{\mu(o_\dg'|\mathbf{s}',\mathbf{s},o_{\dg})}{p_{o}(o_\dg'|\mathbf{s}'
,\mathbf{s},o_{\dg})}\Bigg)\Bigg]+ \mathop{\mathbb{E}}\limits_{\substack{o_\dg'\sim \mu \\ \mathbf{s}'\sim G}}[v_{FE}(\mathbf{s}'
,o_\dg')]\Bigg],\label{eq:AS_Bellman}
\end{align}

where in \eqref{eq:AS_Bellman}, $(\mathbf{s},o) \in \mc{XA}\mc O_{AS}$, and $G$ and $\mu$ in both equations are restricted to policies in the affordance subspace, $\mc O_{AS}$.

\subsection{Option Kernel Analysis and Invertability of $I-A_{\mc N \mc N}$}\label{sec:invertability}
\tom{Recall:
\begin{align*}
\centering
    U_{\dg} = \begin{bmatrix}
        A_{\mc N \mc N}& B_{\mc N \mc T} \\ 
        0_{\mc T \mc N}& I_{\mc T \mc T} 
        \end{bmatrix},
\end{align*}
and:
\begin{align*}
X_{\dg}=\lim\limits_{\mathclap{t \rightarrow \infty}}U_{\dg}^t=
\begin{bmatrix}
\lim\limits_{t \rightarrow \infty}A_{\mc N \mc N}^t& \lim\limits_{t \rightarrow \infty}\left(\sum_{\tau=0}^t A_{\mc N \mc N}^\tau\right)B_{\mc N \mc T} \\ 
0_{\mc T \mc N}& I_{\mc T \mc T} 
\end{bmatrix} = \begin{bmatrix}
0_{\mc N \mc N}& (I- A_{\mc N \mc N})^{-1}B_{\mc N \mc T} \\ 
0_{\mc T \mc N}& I_{\mc T \mc T} 
\end{bmatrix},
\end{align*}
}
\tom{There are a few facts to note about $U_\dg$, $A_{\mc N \mc N}$, and the block matrices.} 

\begin{enumerate}
    \item \tom{All state-actions in $\mc N$ must be \textit{transient} state-actions of the Markov chain $U_{\dg}$, meaning there is a non-zero probability that a trajectory never revisits the state-action in the future. The proof is: if we assume $(x,a)_i$ is recurrent (i.e. not transient), then a trajectory from it has a probability of one of never leaving the set $\mc N$ since all trajectories that leave $\mc N$ are absorbed into $\mc T$, but recurrence also implies the state-action has a desirability of $z((x,a)_i)=0$ (i.e. the cost-to-go $v((x,a)_i)=\infty$ due to positive costs from the cost function $q(x,a)$ on state-actions in $\mc N$). This, however, is a contradiction because zero desirability means that $(x,a)_i$ should be in the terminal set $\mc T$ as defined in Eq. \eqref{eq:termination}, not $\mc N$. Thus, all state-actions in $\mc N$ must be transient.}
    \item \tom{All state-actions in $\mc N$ being transient implies that the matrix $A_{\mc N \mc N}$ must have a spectrum $\boldsymbol{\lambda}$ with a spectral radius $\rho(A_{\mc N \mc N})<1$, because an eigenvalue of $1$ or $-1$ implies the existence of a recurrent state---that is, there would exist a distribution $\mathbf{v}$ over $\mc N$ that is invariant to $\mathbf{v}^TA_{\mc N \mc N}$ and those state-actions have infinite visitations.}
    \item \tom{Given $\rho(A_{\mc N \mc N})<1$, the upper-left sub-matrix $\lim_{t \rightarrow \infty}A_{\mc N \mc N}^t$ must be a sub-matrix of zeros $0_{\mc N \mc N}$ because when using the eigendecomposition, $A_{\mc N \mc N}^t=Q\Lambda^t Q^{-1}$, the matrix $\Lambda$ is a diagonal matrix of the eigenvalues $\boldsymbol{\lambda}$ that go to zero in the infinite limit.}
    \item  \tom{The spectrum of $I- A_{\mc N \mc N}$ is $\boldsymbol{\nu} = 1-\boldsymbol{\lambda}$, meaning all eigenvalues in $\boldsymbol{\nu}$ are non-zero and $I- A_{\mc N \mc N}$ is therefore invertible in the top-right block.}
\end{enumerate}

\subsection{TLMDP Complexity}\label{Complexity}

The run-time complexity of the TLMDP solution \eqref{eq:TLMDP} depends on the specialized structure of the task.  If the ordering constraints preclude the agent from needing to ``undo" sub-goals encoded in $\bs$, the agent is never more than $N_\dg$ policy-calls away from arriving at $\bs_{\mc T} = [1,1,...]$ because it takes at most $N_\dg$ bit flips to solve the task.  With this assumption, we can initialize $\mathbf{z}_0 = \mathbf{b}_\dg$, where $\mathbf{b}_\dg$ is a vector of zeros except for the terminal state-actions for $\bs_{\mc T}$, which are set to the known exponentiated terminal value of $\exp(0)=1$ for a terminal cost of $0$.  $Q_{\sigma \alpha}Q_{\sigma}Q_{o}G_{so}$ is a sparse matrix of size $2^{N_\dg}N_\dg^2 \times 2^{N_\dg}N_\dg^2$, with only $2^{N_\dg}N_\dg^3$ non-zero entries, and power-iteration starting from this initialization converges within $N_\dg$ iterations.  Therefore, we have $N_\dg$ sparse matrix-vector multiplications, resulting in a total of $2^{N_\dg}N_\dg^4$ floating-point operations.
%(Note to reviews: we mistakenly said $2^{N_\dg}N_\dg^3$ in the body of the paper, in reality we expect the algorithm to scale according to $2^{N_\dg}N_\dg^4$.)

\subsection{Compositionality Laws for saLMDP and TLMDP DNF Decomposition}\label{DNF_compo}

Here we describe the (sa/t)LMDPs composition property. Briefly, the compositionality property allows us to take convex mixtures of terminal state (set) costs for component policies and derive a convex state-dependent policy mixture law to form a composite policy.  In the context of the TLMDP, this will allow us to show that a disjunctive normal form policy basis (consisting of a policy for each clause) can be computed independently and combined into a single task policy.

Following \cite{todorov2009compositionality}, we derive the saLMDP and TLMDP compositionality law. Because the saLMDP and TLMDP have the same linear solution (under deterministic transition models),
\begin{align*}
    \mathbf{z}_{\mc N} &= (\text{diag}(\exp(\mathbf{q}_{\mc N}))-P_{xa,\mc N,\mc N})^{-1}P_{xa,\mc N,\mc T}\mathbf{z}_{\mc T},\\
    \mathbf{z}_{\mc N} &= (\text{diag}(\exp(\mathbf{q}_{\mc N}))-G_{s,\mc N,\mc N})^{-1}G_{s,\mc N,\mc T}\mathbf{z}_{\mc T},
\end{align*}
we can extend \cite{todorov2009compositionality} to our new problem formulations.  Note here that $\mathbf{z}_{\mc T} = \exp(-\mathbf{q}_{\mc T})$ is the known terminal state desirability.
\subsubsection{saLMDP composition law}
Consider a set of \textit{component} saLMDPs $\{\mc M_1,\mc M_2,...\}$ which share the same non-terminal cost function $q$ and passive dynamics $p$. Any given convex combination of final state-action costs $f_k(\mathbf{(x,a)}_{\mc T})$ over the terminal state-actons ${(x,a)}_{\mc T}$ with weights $c_1,...,c_K$ forming the \textit{composite} final cost $f({(x,a)_{\mc T}})$ in the form of \eqref{eq:app_terminal_mix}, implies a linearity in the terminal state desirability \eqref{eq:weighted_z}:

\begin{gather}
f({x})=-\log \left(\sum_{k=1}^{K} c_{k} \exp \left(-f_{k}((x,a)_{\mc T})\right)\right),\label{eq:app_terminal_mix}\\
\implies z((x,a)_{\mc T}) = \sum_{k=1}^K c_k z((x,a)_{\mc T}).\label{eq:weighted_z}
\end{gather}

Since $z$ has a linear solution, we can see that a weighted combination of terminal states implies that this linearity must carry over to the desirability of the non-terminal states:

\begin{align*}
    \mathbf{z}_{\mc N} &= (\text{diag}(\exp(\mathbf{q}_{\mc N}))-P_{\mc N,\mc N})^{-1}P_{\mc N,\mc T}\mathbf{z}_{\mc T},\\
    \implies \sum_{k} c_k\mathbf{z}_{k,\mc N} &=  \sum_{k}c_k(\text{diag}(\exp(\mathbf{q}_{\mc N}))-P_{\mc N,\mc N})^{-1}P_{\mc N,\mc T}  \mathbf{z}_{k,\mc T},\\
    \implies \sum_{k} c_k\mathbf{z}_{k,\mc N} &=  (\text{diag}(\exp(\mathbf{q}_{\mc N}))-P_{\mc N,\mc N})^{-1}P_{\mc N,\mc T} \left(\sum_{k}c_k \mathbf{z}_{k,\mc T}\right).
\end{align*}

With this result, we can see that \eqref{eq:weighted_z} applies to non-terminal states.  By combining \eqref{eq:weighted_z} with the definition of the optimal policy \eqref{eq:pol_derivation}, we obtain a state-dependent policy mixture over the $K$ \textit{component policies}, $u_k^*$, for constructing an optimal \textit{composite policy} $u^*$:
\begin{align}
     u^*(a'|x';x,a)&=\sum_k \frac{c_k \mathfrak{N}[z^*_k](x',x,a)}{\sum_n c_n \mathfrak{N}[z^*_n](x',x,a)}\frac{p(a'|x';x,a)z_k(x',a')}{\mathfrak{N}[z^*_k](x',x,a)},\label{eq:pol_mix_1}\\
     &=\sum_k \frac{c_k z_k(x,a)\exp(q(x,a))}{\sum_n c_n z_n(x,a)\exp(q(x,a))}\frac{p(a'|x';x,a)z_k(x',a')}{\mathfrak{N}[z^*_k](x',x,a)},\label{eq:pol_mix_2}\\
     &=\sum_k \frac{c_k z(x,a)}{\sum_n c_n z_n(x,a)}\frac{p(a'|x';x,a)z_k(x',a')}{\mathfrak{N}[z^*_k](x',x,a)},\label{eq:pol_mix_3}
\end{align}
where \eqref{eq:pol_mix_1} implies \eqref{eq:pol_mix_2} by using the identity derived from \eqref{eq:salmdp_linear_det}:
\begin{align*}
     \mathop{\mathbb{E}}\limits_{x'\sim p_x }\mathfrak{N}[z^*](x',x,a) &= z(x,a)\exp(q(x,a)),\\
     \implies \mathfrak{N}[z^*](x',x,a) &= z(x,a)\exp(q(x,a)) \quad (\text{when~} x'\sim p_x, \text{~for deterministic~}p_x).
\end{align*}
Note, that we are not interested in computing a policy composition for tuples $(x',x,a)$ in the conditional when $x'$ can not be reached from $x$ under some action $a$, (i.e. $p_x(x'|x,a)\neq 1$ for deterministic $p_x$) because the policy $u_a$ will never encounter these generated tuples under $p_x$. In \eqref{eq:pol_mix_2}, since $q$ is shared across all component problems it is independent of $k$ and drops out resulting in \eqref{eq:pol_mix_3}. By letting the function,
\begin{align*}
    m_k(x,a)=\frac{c_{k} z_{k}(x,a)}{\sum_{n} c_{n}z_{n}(x,a)},
\end{align*}
be a state-dependent mixture weight (with $\sum_k m_k(x)=1$, $m_k(x)>1$), we can then write the policy composition law as:
\begin{align*}
    u^{*}(a'|x';x,a)=\sum_{k} m_k(x,a) u_k^*(a'|x';x,a).
\end{align*}

% For a combination of final state-action costs in an saLMDP we have the following mixture law:
% \begin{gather}
%     f(\mathbf{x,a})=-\log \left(\sum_{k=1}^{K} c_{k} \exp \left(-f_{k}(\mathbf{x,a})\right)\right)\\
%     u^{*}\left(a^{\prime}|x',x,a\right)=\sum_{k} \frac{c_{k} z_{k}(x,a)}{\sum_{n} c_{n} 
%     z_{n}(x,a)} u_k(a'|x',x,a)=\sum_{k} m_k(x,a) u_k(a'|x';x,a)
% \end{gather}
\subsubsection{TLMDP composition law}
Since the TLMDP is simply a change in the parameters, we can use the previously described method to derive the TLMDP policy mixture law corresponding for a linear combination of terminal state-option costs:
\begin{gather*}
    % f(\mathbf{s},\boldsymbolo)=-\log \left(\sum_{k=1}^{K} c_{k} \exp \left(-f_{k}(\mathbf{s},\boldsymbolo)\right)\right)\\
    \mu^{*}\left(o^{\prime}|\mathbf{s}';\mathbf{s},o\right)=\sum_{k} \frac{c_{k} z_{k}(\mathbf{s},o)}{\sum_{n} c_{n} 
    z_{n}(\mathbf{s},o)} \mu_{k}(o'|\mathbf{s}',\mathbf{s},o)=\sum_{k} m_k(\mathbf{s},o) \mu_{k}(o'|\mathbf{s}';\mathbf{s},o),
\end{gather*}

where each component policy $\mu_{k}$ corresponds to a clause $w_k$ in a BOG-task DNF formula.

\subsection{Trajectory Likelihood Function}\label{appx:saLMDP_likihood}

Here we derive the trajectory liklihood functions for the saLMDP and TLMDP:

\subsubsection{saLMDP trajectory likelihood}\label{likelihood_1}

Consider a hierarchical trajectory,
\begin{align*}
    \mathbf{xa} = (x,a)_{t_0},(x,a)_{t_1},...,(x,a)_{T}.
\end{align*}
% $$\mathbf{s} = s_{t_{k_0}},\mathbf{s}_{t_{k_1}},...,\mathbf{s}_{t_{K}} = (\bs_{t_{k_0}},x_{t_{k_0}}),(\bs_{t_{k_1}},x_{t_{k_1}}),...,(\bs_{t_K},x_{t_K})$$
% where $k$ denotes the \textit{period}, and $t_k$ is the start time of the period. 
Using the identities:
\begin{align*}
   &u_a^*(a'|x';x,a) = \frac{z(x',a')p_a(a'|x';x,a)}{\mathfrak{N}[z^*](x',x,a)},\\
    &z(x,a) = \exp(-q(x,a))\mathfrak{N}[z^*](x',x,a)\implies \mathfrak{N}[z^*](x',x,a)=z(x,a)\exp(q(x,a)),
\end{align*}
we obtain:
\begin{align*}
    u_a^{*}(a'|x';x,a) = \exp(-q(x,a))p_{a}(a'|x';x,a)\frac{z(x',a')}{z(x,a)},
\end{align*}

For the likelihood function for the saLMDP we can put a probability on a state-action trajectory:

\begin{align*}
    p(\mathbf{xa}|u_{xa}) &= \prod_{t=0}^{T-1} u_{xa}(x_{t+1},a_{t+1}|x_t,a_t)= \\p(\mathbf{xa}|u_a,p_x)&= \prod_{t=0}^{T-1} u_{a}(a_{t+1}|x_{t+1};x_t,a_t)p_x(x_{t+1}|x_t,a_t)\\
    &= \prod_{t=0}^{T-1} \exp (-q(x_t,a_t)) p_a(a_{t+1}|x_{t+1};x_t,a_t)p_x(x_{t+1}|x_t,a_t) \frac{z\left(x_{t+1},a_{t+1}\right)}{z(x_t,a_{t})}\\
    &=\frac{\exp(-q(x_{t_0},a_{t_0}))p(a'_{t_1}|x_{t_1},a_{t_0},x_{t_0})\cancel{z(x_{t_1},a_{t_1})}}{z(x_{t_0},a_{t_0})}\times\\ \nonumber&\quad\quad\quad\quad\frac{\exp(-q(x_{t_1},a_{t_1}))p(a_{t_2}|x_{t_2},x_{t_1},a_{t_1})\cancel{z(x_{t_2},a_{t_2})}}{\cancel{z(x_{t_1},a_{t_1})}}\times...\times\\
    \nonumber&\quad\quad\quad\quad\frac{\exp(-q(x_{T-1},a_{T-1}))p(a_{T}|x_{T},x_{T-1},a_{T-1})z(x_{T},a_{T})}{\cancel{z(x_{T-1},a_{T-1})}}\\
    &=\frac{z(x_{T},a_{T})}{z(x_{t_0},a_{t_0})} \prod_{t=0}^{T-1} \exp (-q(x_{t},a_{t}))p_a(a_{t+1}|x_{t+1};x_t,a_t)p_x(x_{t+1}| x_t,a_t).
\end{align*}

% However, typically we do not actually observe actions. If we have computed a state-action desirability function and we only observe $\mathbf{x}$ we can infer the initial and final actions.  Under the assumption of a deterministic transition kernel and with a unique $a$ transitioning $x$ to $x'$ through $p_x(x'|x,a)$, there will be a unique $a$.  We will make this assumption for the rest of the exposition, but for stochastic transition kernels, we can simply estimate the action for a state pair $(x,x')$ with maximum likelihood: $\hat{a}=\argmax_a p_x(x'|x,a)$. We call the inferred state-action pair $(x,\hat{a})$.  The probability over the trajectory $\mathbf{x}$ is then:

% \begin{align}
%     p(\mathbf{x}|u_a,p_x) = \frac{z(x_{T},\hat{a}_{T})}{z(x_{t_0},\hat{a}_{t_0})} \prod_{t=0}^{T-1} \exp (-q(x_{t},\hat{a}_{t}))p_a(\hat{a}_{t+1}|x_{t+1};x_t,\hat{a}_t)p_x(x_{t+1}| x_t,\hat{a}_t)\label{eq:inferred_a}
% \end{align}

\subsubsection{Hierarchical Trajectories}\label{likelihood_2}
Having derived the likelihood function for the saLMDP, we can extend this result to a hierarchical trajectory.  We define the trajectory:
\begin{align*}
    \mathbf{s}o = (\mathbf{s},o)_{k_0},(\mathbf{s},o)_{k_1},...,(\mathbf{s},o)_{k_f},
\end{align*}
% $$\mathbf{s} = s_{k_0},\mathbf{s}_{k_1},...,\mathbf{s}_{k_f}$$ 
where $\mathbf{s} = [\bs, \alpha,x,a]$ and $k$ is the \textit{period}, an abstract time variable that increments for each change the state $\bs$. Since the TLMDP is simply an saLMDP with different transition kernels, states, and actions (policies), it has the same form:

% And because $o$ is never actually observed, following \eqref{eq:inferred_a} we can create a probability over $\mathbf{s}$:

% \begin{align}
%     p(\mathbf{s}|\mu,p_s)=\frac{z\left(\mathbf{s}_{T},\hat{o}_T\right)}{z\left(\mathbf{s}_{t_0},\hat{o}_{t_0}\right)} \prod_{t=0}^{T-1} \exp \left(-q\left(\mathbf{s}_{t},\hat{o}_t\right)\right)p_{o}\left(\hat{o}_{t+1},|\mathbf{s}_{t+1};\mathbf{s}_{t},\hat{o}_{t}\right)p_s\left(\mathbf{s}_{t+1}|\mathbf{s}_t,\hat{o}_t\right)
% \end{align}

% \begin{align}
%     p(\mathbf{s}o|\mu,p_s)=\xunderbrace{\frac{z\left(\mathbf{s}_{T},o_T\right)}{z\left(\mathbf{s}_{t_0},o_{t_0}\right)}}_{r(z,\mathbf{s}o)} \xunderbrace{\prod_{t=0}^{T-1} \exp \left(-q\left(\mathbf{s}_{t},o_t\right)\right)p_{o}\left(o_{t+1},|\mathbf{s}_{t+1};\mathbf{s}_{t},o_{t}\right)p_s\left(\mathbf{s}_{t+1}|\mathbf{s}_t,o_t\right)}_{c(z,p,q)} \label{eq:TLMDP_inference}
% \end{align}

\begin{align}
    p(\mathbf{s}o|\mu,p_s)=\frac{z\left(\mathbf{s}_{T},o_T\right)}{z\left(\mathbf{s}_{t_0},o_{t_0}\right)}\prod_{t=0}^{T-1} \exp \left(-q\left(\mathbf{s}_{t},o_t\right)\right)p_{o}\left(o_{t+1},|\mathbf{s}_{t+1};\mathbf{s}_{t},o_{t}\right)p_s\left(\mathbf{s}_{t+1}|\mathbf{s}_t,o_t\right). \label{eq:TLMDP_inference}
\end{align}

This formulation could prove to be very useful for task inference in the context of non-Markovian tasks.  We leave the development of this to future work.

\subsection{Proof of Task-MDP Decomposition}\label{app:tMDP}

% \comment{Is the discussion below important to understand the theorem on equivalency? If yes maybe we can move it up to the main body?}

To assess the optimality between the goal kernel and the full transition kernel, we formulate the problem in the non-entropy regularized regime by defining a Task-MDP. 

% We use this proof to analyze the TLMDP considering the result in \cite{todorov2009efficient} that an LMDP solution can approximate a shortest-path policy arbitrarily well as the cost is increased to infinity.

\begin{definition}[task-MDP]
    A task-MDP (tMDP) is defined by the tuple $\mathscr{L}=(\mathscr{M}^{N}, \mathscr{T}, h)$, where $\mathscr{M}^{N}$ is a set of first-exit MDP control problems of size $N$, the number of states in $\mc X$ to be controlled to, $\mathscr{T}$ is an BOG task, and $F:\mc X \mc A \rightarrow  \mc A_{\sigma}$ is the affordance function.
\end{definition}

Each MDP $\mathscr{M}$ in $\mathscr{M}^{N}$ is defined as a first-exit problem to one of $N$ goal (terminal) states.  $\mathscr{T}$ is the same BOG task as used in the TLMDP, and implies the kernel $P_{\boldsymbol{\sigma}}(\bs'|\bs,\alpha)$. The only major difference between the tMDP and the TLMDP is that in the tMDP we require the policy at the terminal state to map to the high-level action.  We can construct the full dynamics as:
\begin{align*}
    P(\bs',x'|\bs,x,a) := \sum_{\alpha}P_{\boldsymbol{\sigma}}(\bs'|\bs,\alpha')F(\alpha|x,a)P_x(x'|x,a).
\end{align*}

We also define a full-task-MDP (where ``full" refers to using the ``full kernel", i.e. non-semi-MDP) as:
\begin{definition}[full-task-MDP]
    A full-task-MDP (ftMDP) $\mathscr{M}=(\Sigma \times \mc X,\mc A,P_{\boldsymbol{\sigma}} \circ F \circ  P_x,q,\Sigma_{\mc T})$ is an MDP where $\Sigma \times \mc X$ is the extended state space, $\mc A$ is the set of actions, $T = P_{\boldsymbol{\sigma}} \circ F \circ P_x$ is the transition kernel, $q$ is the cost function, and $\Sigma_{\mc T}$ is the set of final terminal states.
\end{definition}

We can now state the decomposition theorem: 

\begin{theorem}[Task-MDP Decomposition]
  Let $\mathscr{M}=$ $(\Sigma \times \mc X,\mc A,P_{\boldsymbol{\sigma}}\circ F \circ P_x,q,\Sigma_{\mc T})$ be a full-task-MDP (ftMDP) and $\bar{\mathscr{M}}=(\mc M^n, P_{\boldsymbol{\sigma}},h,\Sigma_{\mc T})$ be a deterministic Task-MDP (where $\mc M = (\mc X,\mc A, P_x, q) \in \mc M^n$ is an MDP with a first-exit control solution to one of $n$ terminal states on $\mc X$). When $\mc X$, $\mc A$, $P_x$, $P_{\boldsymbol{\sigma}}$, $F$, and $\Sigma_{\mc T}$ are the same for both problems, the optimal value function $\bar{v}$ derived for $\bar{\mathscr{M}}$ and the optimal value function $v$ derived for the Task-MDP are equivalent, $v=\bar{v}$.
\end{theorem}
% \begin{remark}
% \label{thm:fulljumpthm}
% The value function derived from a Bellman equation with a full kernel is equivalent to a value function derived from a Bellman equation with an option kernel
% \end{remark}
\textbf{\textit{Proof:}} Here we show that with a tMDP, a Bellman equation based on a deterministic full transition kernel with a deterministic policy achieve the same value function as a Bellman equation based on the option kernel when the option kernel cost function is set to be the value function of the underlying ensemble policy. 

Let us assume that the ``full kernel" and the policy $\pi(a|x)$ are deterministic. 
% $$T(\bs',x'|\bs,x,a) = P_{\boldsymbol{\sigma}}(\bs'|\bs,\alpha')h(\dg|x,a)P_x(x'|x,a).$$
We begin with the definition of the optimal value function for a deterministic first-exit control problem:
\begin{align*}
     v^*(\bs,x) = 
    q(\bs,x,a)+v^*(\bs',x')T(\bs',x'|\bs,x,a)\pi^*(a|\bs,x).%\hspace{30mm}
\end{align*}
Unrolling the recursion we have:
\begin{align*}
    v^*(\bs,x) = 
    q(\bs_{t_f},x_{t_f},a_{t_f}) + \sum_{t=0}^{t_f-1}q(\bs_t,x_t,a_t)T(\bs_t,x_t|\bs_{t-1},x_{t-1},a_{t-1})\pi^*(a|\bs,x). %\hspace{30mm}
\end{align*}
We can decompose this summation by isolating constant $\bs$ variables within each $\bs \rightarrow \bs'$ transition by creating a new index $k$ for each \textit{period}  when $\bs$ remains constant, and where $\tau_k$ indexes inter-period discrete time steps:
\begin{align*}
    & v^*(\bs,x) = \nonumber \\
    & q(\bs_{t_f},x_{t_f},a_{t_f}) + \sum_{k=0}^{k_f-1}\sum_{\tau_k=0}^{\tau_{k,f}}q(\bs_{\tau_k},x_{\tau_k},a_{\tau_k})T(\bs_{\tau_k},x_{\tau_k}|\bs_{{\tau_k}-1},x_{{\tau_k}-1},a_{{\tau_k}-1})\pi^*(a|\bs,x) = \\
    & q(\bs_{t_f},x_{t_f},a_{t_f}) + \sum_{k=0}^{k_f-1} \Bigg[q(\bs_{\tau_{k,f}},x_{\tau_{k,f}},a_{\tau_{k,f}})T(\bs_{\tau_{k,f}},x_{\tau_{k,f}}|\bs_{\tau_{k,f}-1},x_{\tau_{k,f}-1},a_{\tau_{k,f}-1})\pi^*(a|\bs,x)\nonumber\\&\qquad\qquad\quad+\sum_{\tau_k=0}^{\tau_{k,f}-1}q(\bs_{\tau_{k}},x_{\tau_{k}},a_{\tau_{k}})T(\bs_{\tau_k},x_{\tau_k}|\bs_{\tau_k-1},x_{\tau_k-1},a_{\tau_k-1})\pi^*(a|\bs,x)\Bigg]=\\
    &q(\bs_{t_f},x_{t_f},a_{t_f}) + \sum_{k=0}^{k_f-1}\Bigg[q(\bs_{\tau_{k,f}},x_{\tau_{k,f}},a_{\tau_{k,f}})T_{\pi^*}(\bs_{\tau_{k,f}},x_{\tau_{k,f}},a_{\tau_{k,f}}|\bs_{\tau_{k,f}-1},x_{\tau_{k,f}-1},a_{\tau_{k,f}-1})\nonumber\\
    &\qquad\qquad\quad+\sum_{\tau_k=0}^{\tau_{k,f}-1}q(\bs_{\tau_{k,f}},x_{\tau_{k,f}},a_{\tau_{k,f}})T_{\pi^*}(\bs_{\tau_k},x_{\tau_k},a_{\tau_k}|\bs_{{\tau_k}-1},x_{{\tau_k}-1},a_{\tau_{k}-1})\Bigg],
\end{align*}
    where $T_{\pi^*}: (\Sigma \times \mc X \times \mc A) \times (\Sigma \times \mc X \times \mc A) \rightarrow [0,1]$ is the induced Markov chain by applying the policy $\pi^*$. By conditioning the transition kernel on the first time step of period $k$, we have:
\begin{align}  
    & v^*(\bs,x) = q(\bs_{t_f},x_{t_f},a_{t_f}) + \sum_{k=0}^{k_f-1}\Bigg[q(\bs_{\tau_{k,f}},x_{\tau_{k,f}},a_{\tau_{k,f}})T^{\tau_{k,f}}_{\pi^*}(\bs_{\tau_k},x_{\tau_k},a_{\tau_k}|\bs_{k},x_{k},a_{k})\nonumber\\
    &\qquad\qquad\quad+\sum_{\tau_k=0}^{\tau_{k,f}-1}q(\bs_{\tau_{k,f}},x_{\tau_{k,f}},a_{\tau_{k,f}})T^{\tau_{k}}_{\pi^*}(\bs_{\tau_k},x_{\tau_k},a_{\tau_k}|\bs_{k},x_{k},a_{k})\Bigg]=\\
    & q(\bs_{t_f},x_{t_f},a_{t_f}) + \sum_{k=0}^{k_f-1}\Bigg[q(\bs_{\tau_{k,f}},x_{\tau_{k,f}},a_{\tau_{k,f}})T^{\tau_{k,f}}(\bs_{\tau_k},x_{\tau_k},a_{\tau_k}|\bs_{k},x_{k},a_{k},\pi^*_{\bs})\mu^*(\pi^*_{\bs}|\bs,x)\nonumber\\
    &\qquad\qquad\quad+\sum_{\tau_k=0}^{\tau_{k,f}-1}q(\bs_{\tau_{k,f}},x_{\tau_{k,f}},a_{\tau_{k,f}})T^{\tau_{k}}(\bs_{\tau_k},x_{\tau_k},a_{\tau_k}|\bs_{k},x_{k},\pi^*_{\bs})\mu^*(\pi^*_{\bs}|\bs,x)\Bigg],\label{eq:unrolled-with-k2}
\end{align}
where $\mu^*: \Sigma \times \mc X \rightarrow \mc O_{\bs}$, $\mc O_{\bs} = \{\pi^*_{\bs_1}, \pi^*_{\bs_2},...\}$ is a map to the set of optimal policies conditioned on $\bs$, and $\pi^*_{\bs}: \mc X \rightarrow \mc A$ is the optimal state-action map conditioned on $\bs$ being constant. We omit the timestamps on the arguments of $\pi^*_{\bs}$, since it is a stationary policy.

Note that the term inside the first summation in \eqref{eq:unrolled-with-k2} is itself a first-exit control problem, which can be defined as:
\begin{align*}
    &v^*_{\bs}(\bs_k,x_k|\pi^*_{\bs}) := q(\bs_{\tau_{k,f}},x_{\tau_{k,f}},a_{\tau_{k,f}})T^{\tau_{k,f}}(\bs_{\tau_k},x_{\tau_k},a_{\tau_k}|\bs_{k},x_{k},a_{k},\pi^*_{\bs})\mu^*(\pi^*_{\bs}|\bs,x)\nonumber\\
    &\qquad\qquad\quad+\!\!\!\sum_{\tau_k=0}^{\tau_{k,f}-1}\!\!\!q(\bs_{\tau_k},x_{\tau_k},a_{\tau_k})T^{\tau_{k}}(\bs_{\tau_k},x_{\tau_k}|\bs_{k},x_{k},\pi^*_{\bs})\mu^*(\pi^*_{\bs}|\bs,x).
    %&\mathbf{v}^*_{\bs,k} = (I-T_{\pi \bs,NN})^{-1}\mathbf{q}_N + (I-T_{\pi \bs,NN})^{-1}T_{\pi \bs,NT}\mathbf{q}_T,\\
    %&\mathbf{v}^*_{\bs,k} = M_{\pi,\bs}(\mathbf{q}_N +T_{\pi,NT}\mathbf{q}_T),
\end{align*}

%where $\mathbf{v}^*_{\bs,k}$ is the vector form of the optimal value function, $M_{\pi,\bs} = (I-T_{\pi \bs,NN})^{-1}$ is the fundamental matrix for the inter-sub-goal dynamics, $\mathbf{q}_N$ is the cost vector restricted to the non-terminal states, and $\mathbf{q}_T$ is the cost vector restricted to the terminal states. $NN$ stands for non-terminal to non-terminal transitions, and $NT$ stands for non-terminal to terminal transitions. Since $\bs$ can only transition at the affordance sub-goal state-actions, the terminal (terminal) state-actions are defined as $\mc T = \{(x,a)\leftarrow h(\alpha), \forall \alpha \in \mc A_{\sigma}\}$, thus, the interior first-exit control problem is defined with respect to a set of aggregated terminal state-actions. 

Substituting $v^*_{\bs}(\bs_k,x_k|\pi^*_{\bs})$ for the inner summation we obtain: 
\begin{align}
    v^*(\bs,x) = q(\bs_{t_f},x_{t_f},a_{t_f}) + \sum_{k=0}^{k_f-1} v^*_{\bs}(\bs_k,x_k|\pi^*_{\bs})P(\bs_k,x_k|\bs_{k-1},x_{k-1},\pi^*_{\bs_k})\label{eq:reduced}\mu^*(\pi^*_{\bs}|\bs,x),
\end{align}

where $P(\bs_k,x_k|\bs_{k-1},x_{k-1},\pi^*_{\bs_k}) = T^{\tau_{k,f}}_{\pi^*}(\bs_{\tau_k},x_{\tau_k}|\bs_{k},x_{k},\pi^*_{\bs_k})$. %Also, we know that in matrix form $T^{\tau_k}_{\pi^*} = (I-T_{\pi^*})^{-1}H$, where $H$ is a matrix of zeros except for the terminal state columns which are taken from $T_{\pi^*}$.
Given an optimal policy, we have no \textit{a priori} knowledge of which terminal states are traversed under the dynamics.  However, if we define a new \textit{set} of first-exit problems specified by \textit{individual} terminal states in $\mc T$, we can substitute these individual first-exit problems into the overall problem.

In \eqref{eq:reduced} we are compressing the inner loop into a policy evaluation with the (deterministic) probability of jumping from $(\bs_{k-1}, x_{k-1})$ to $(\bs_{k}, x_{k})$.  
% The cost function $q$ decomposes into $q_{\sigma \alpha} + q_{\sigma} + q_{xa}$ (This isn't represented in the equation, it needs to be fixed!) over the full space, and becomes $q_{\sigma \alpha} + q_{\sigma} + q_{\pi}$ in the compressed space.  
Since $q_{xa}(x,a)$ is not dependent on $\bs$, we can reuse the associated policy for any $\bs$.  Therefore, we can introduce a set of policies $\hat{\Pi} = \{\hat{\pi}_1,\hat{\pi}_2,...\}$, derived from a set of value functions $\mc V = \{\hat{v}_1, \hat{v}_2, ...\}$, the induced Markov chains $\mc M = \{T_{\pi_1}, T_{\pi_2}, ...\}$, and its corresponding ``jump" $J(x_j|x_i,\hat{\pi})$. Recall that $J$ is defined as:
\begin{align*}
    J(x_j|x_i,\pi_g) = \mathbf{e}_i^T(I-A_{\bar g})^{-1}B_g\mathbf{e}_j = X_{\bar{g}}(i,j),
\end{align*} 

where $U_{\dg} = \sum_a T(x'|x,a)\pi(a|x)$ is the controlled Markov chain under policy $\pi$, where $\bar g$ indicates that we remove the row and column corresponding to the terminal state $x_g$. Here, $M_{\bar{g}}=(I-A_{\bar g})^{-1}$ is the fundamental matrix, and $H$ is a rank $1$ matrix of zeros with a column at index $g$ which is the column that was removed from $U_{\dg}$.

Writing the composition of the option kernel with the policy and task transition, we define $\hat{P}$,

\begin{align*}
    \hat{P}(\bs',x',a'|\bs,x_i,\pi_g) = P_{\boldsymbol{\sigma}}(\bs'|\bs,\alpha)F(\alpha|x',a')\pi(a'|x')J(x'|x_i,\hat{\pi}),
\end{align*}

which will substitute into the original problem for $P(\bs_k,x_k,a_k|\bs_{k-1},x_{k-1},a_{k-1},\pi^*_{\bs_k})$ along with $v^*_{\bs}(\bs_k,x_k|\pi^*_{\bs})$ substituted with $\hat{v}$. That is, if we can show that $v^*_{\bs}$ is contained within $\mc V$, and $P_\pi$ can be produced by the corresponding policy in $\hat{P}$, then we can write an optimization which substitutes the $(\hat{v},\hat{P})$ pair for the original functions $(v_{\bs}^*,P)$. 
% T(\bs)(\bs'|\bs,\alpha)h(\alpha|x',a')

Note that $P(\bs_k,x_k|\bs_{k-1},x_{k-1},\pi^*_{\bs_k})$ is defined in terms of $k$, the period index. Therefore, it represents the abstract time when $\bs$ transitions to $\bs'$. If we are to substitute $\hat{P}$ for $P$, clearly it must be the case a that policy called by $J$ cannot produce any actions that transition $\bs$ en route to another goal. To ensure this doesn't happen, the cost function which produces $\hat{o_\dg}$ must penalize the potential alternative goal state-actions with an arbitrarily high cost.

We define a new high-level policy $\hat{\mu}: \Sigma \times \mc X \rightarrow \hat{\Pi}$. Since we are working with deterministic dynamics and policies, there is only a single trajectory followed by the agent under the optimal policy, implying that only one of the terminal states is traversed.  This means that for a given $P(\bs_k,x_k,a_k|\bs_{k-1},x_{k-1},a_{k-1},\pi^*_{\bs_k})$, there exists a way to condition $J(\bs',x'|\bs,x,\hat{\pi}_\dg)$ to achieve the same terminal state.  That is, 

$$P(\bs_k',x_k',a_k'|\bs_{k-1},x_{k-1},a_{k-1},\pi^*_{\bs_k}) - \max_{\hat{\pi}\in \hat{\Pi}}\hat{P}(\bs',x',a'|\bs,x,a,\hat{o_\dg})=0.$$

% However, the above expression can only be true in all cases so long as $\hat{o_\dg}$ does not produce any trajectories from any initial state that induces a different goal variable that advances the state $\bs$. 

We call the policy that reaches the single traversed terminal state-action of the full-kernel policy the jump-optimal policy:
% How about jump-optimal? (I don't think I should call this "optimal" yet because this is only with respect to the transition kernel, not the value function.)
$$\hat{\pi}_\dg^+ = \argmax_{\hat{\pi}\in \hat{\Pi}}P_{\boldsymbol{\sigma}}(\bs'|\bs,\alpha)F(\alpha|x',a')J(\bs',x',a'|\bs,x,a,\hat{\pi_g}).$$

Because the jump-optimal policy, $\hat{\pi}_\dg^+$, is a deterministic shortest-path to the same terminal state as $\pi^*_{\bs_k}$, the corresponding ensemble value function, $\hat{v}_{\dg}$, must be the same as $\hat{v}_{\bs}^*$ evaluated at a given state $x$.  Therefore, the ``jump-optimal" policy is also the value-optimal policy to select from state $x$, which must be equivalent in value to the policy $o_{\bs}^*$ followed from $x$: $\hat{\mu}^*(\hat{\pi}_\dg^+|\bs,x)=\hat{\mu}^*(\hat{\pi}_\dg^*|\bs,x)=\mu^*(o_{\bs}^*|\bs,x)$. We can then write the following equivalence: 

$$P(\bs_k',x_k'|\bs_{k-1},x_{k-1},\pi^*_{\bs_k})\mu^*(\pi^*_{\bs}|\bs,x) = J(\bs',x'|\bs,x,\hat{\pi}_\dg^*)\hat{\mu}^*(\hat{\pi}_\dg^*|\bs,x).$$

Substituting the option kernel for $P$ and the optimal ensemble value function $\hat{v}^*$ for $v_{\bs}$, the optimization in \eqref{eq:reduced} becomes:

\begin{align*}
    &v^*(\bs,x) = q(\bs_{t_f},x_{t_f},a_{t_f}) + \sum_{k=0}^{k_f-1}\left[ \hat{v}_{\bs}^*(\bs_k,x_k|\pi^*_{\bs}) J(\bs_k,x_k|\bs_{k-1},x_{k-1},\hat{\pi}_{\dg})\hat{\mu}^*(\hat{\pi}_\dg|\bs,x)\right]. \hspace{10mm}
\end{align*}

Substituting the optimal policy for a minimization over policies, we have:

\begin{align*}
    % &v^*(\bs,x) = q(\bs_{t_f},x_{t_f},a_{t_f}) + \sum_{k=0}^{k_f-1}\left[ (\hat{v}_{\bs}^*(\bs_k,x_k|\pi^*_{\bs}) - \hat{v}_{\hat{\pi}_\dg}^*+\hat{v}_{\hat{\pi}_\dg}^*) J(\bs_k,x_k|\bs_{k-1},x_{k-1},\hat{\pi}_{\dg})\mu^*(\hat{\pi}_\dg|\bs,x)\right]\\
    % &v^*(\bs,x) = q(\bs_{t_f},x_{t_f},a_{t_f}) + \sum_{k=0}^{k_f-1}\left[\hat{v}_{o_\dg}^*(\bs,x)J(\bs_k,x_k|\bs_{k-1},x_{k-1},\hat{\pi}_{\dg})\mu^*(\hat{o_\dg}|\bs,x)\right]\\
    &v^*(\bs,x) = q(\bs_{t_f},x_{t_f},a_{t_f}) + \min_{\hat{\pi}_\dg \in \hat{\Pi}}\sum_{k=0}^{k_f-1}\left[ \hat{v}_{\dg}(\bs_k,x_k|\hat{\pi}_{\dg})J(\bs_k,x_k|\bs_{k-1},x_{k-1},\hat{\pi}_{\dg})\right],\\
    &v^*(\bs,x) = \min_{\hat{\pi}_\dg \in \hat{\Pi}}\left[\hat{q}(\bs_k,x_k,\hat{\pi}_{\dg}) + v^*(\bs', x')J(\bs',x'|\bs,x,\hat{\pi}_{\dg})\right],\\
    &\hat{\mu}^*(\hat{\pi}_\dg|\bs,x) = \argmin_{\hat{\pi}_\dg \in \hat{\Pi}}\left[\hat{q}(\bs_k,x_k,\hat{\pi}_{\dg}) + v^*(\bs', x')J(\bs',x'|\bs,x,\hat{\pi}_{\dg})\right],
\end{align*}
% \begin{align}
%     % &v^*(\bs,x) = q(\bs_{t_f},x_{t_f},a_{t_f}) + \sum_{k=0}^{k_f-1}\left[ (\hat{v}_{\bs}^*(\bs_k,x_k|\pi^*_{\bs}) - \hat{v}_{\hat{\pi}_\dg}^*+\hat{v}_{\hat{\pi}_\dg}^*) J(\bs_k,x_k|\bs_{k-1},x_{k-1},\hat{\pi}_{\dg})\mu^*(\hat{\pi}_\dg|\bs,x)\right]\\
%     % &v^*(\bs,x) = q(\bs_{t_f},x_{t_f},a_{t_f}) + \sum_{k=0}^{k_f-1}\left[\hat{v}_{o_\dg}^*(\bs,x)J(\bs_k,x_k|\bs_{k-1},x_{k-1},\hat{\pi}_{\dg})\mu^*(\hat{o_\dg}|\bs,x)\right]\\
%     &v^*(\bs,x) = q(\bs_{t_f},x_{t_f},a_{t_f}) + \min_{\hat{\pi}_\dg \in \hat{\Pi}}\sum_{k=0}^{k_f-1}\left[ \hat{v}_{\dg}(\bs_k,x_k|\hat{\pi}_{\dg})J(\bs_k,x_k|\bs_{k-1},x_{k-1},\hat{\pi}_{\dg})\right],\\
%     &v^*(\bs,x) = \min_{\hat{\pi}_\dg \in \hat{\Pi}}\left[\hat{q}(\bs_k,x_k,\hat{\pi}_{\dg}) + \expec_{\bs',x'\sim J(\cdot|\bs,x,\hat{\pi}_{\dg})}v^*(\bs', x')\right],\\
%     &\hat{\mu}^*(\hat{\pi}_\dg|\bs,x) = \argmin_{\hat{\pi}_\dg \in \hat{\Pi}}\Bigg[\hat{q}(\bs_k,x_k,\hat{\pi}_{\dg}) + \expec_{\bs',x'\sim J(\cdot|\bs,x,\hat{\pi}_{\dg})}v^*(\bs', x')\Bigg],
% \end{align}
% where
% \begin{align}\label{eq:phi_star}
% \end{align}
where $\hat{q}(\bs_k,x_k,\hat{\pi}_{\dg}) = \hat{v}_{\dg}(\bs_k,x_k|\hat{\pi}_{\dg})$.

We have thus obtained a new Bellman equation with a different transition kernel, the ``goal kernel". This new goal kernel Bellman equation is guaranteed to be equivalent to the value function of the Bellman equation with the full kernel when the cost function is defined as the value function of the ensemble policy.$\hfill\square$

\end{document}